\documentclass{article}

\usepackage{arxiv}

\usepackage[T1]{fontenc}
\AtBeginDocument{%
  \providecommand\BibTeX{{%
    \normalfont B\kern-0.5em{\scshape i\kern-0.25em b}\kern-0.8em\TeX}}}

\usepackage{enumitem}
\usepackage{svg}
\usepackage{algorithm}
\usepackage{algpseudocode}
\usepackage{tabto} 
\usepackage{dsfont}
\usepackage{bm}
\usepackage{hyperref}
\usepackage[titletoc]{appendix}
\usepackage{mathtools}
\usepackage{eqparbox}
\usepackage{optidef}
\usepackage{appendix}
\usepackage{etoolbox,xspace}
\usepackage{multicol}
\usepackage{amsfonts}   
\usepackage{amssymb,amsmath,amsthm}
\usepackage[rightcaption]{sidecap}
\newcommand{\OMDNEG}{$\OMD_\negent$}

\newcommand{\deriv}[2]{	\frac{\partial #1}{\partial #2}}

\DeclarePairedDelimiter\floor{\lfloor}{\rfloor}
\newcolumntype{C}[1]{>{\raggedright\arraybackslash}p{#1}}
\newcommand{\card}[1]{\left|#1\right|}
\newcommand{\given}{\;\vert\;}
\newcommand{\norm}[1]{\left\lVert#1\right\rVert}
\newcommand{\BigO}[1]{\ensuremath{\operatorname{\mathcal O}\left(#1\right)}}
\newcommand{\BigOmega}[1]{\ensuremath{\operatorname{\Omega}\left(#1\right)}}

\newcommand{\argmin}{\mathop{\arg\min}}

\newcommand{\new}[1]{\textbf{\textcolor{blue}{#1}}}

\newcommand{\Regret}{\ensuremath{\mathrm{Regret}}}
\newcommand{\OMD}{\ensuremath{\mathrm{OMD}}}
\newcommand{\OGD}{\ensuremath{\mathrm{OGD}}}
\newcommand{\nrm}{\ensuremath{\text{-}\mathrm{norm}}}
\newcommand{\negent}{\ensuremath{\mathrm{NE}}}
\newcommand{\updatecost}{\ensuremath{\mathrm{UC}}}
\newcommand{\supp}{\mathop{\ensuremath{\mathrm{supp}}}}
\renewcommand{\vec}[1]{\bm{\mathbf{#1}}}

\renewcommand{\vec}[1]{\pmb{#1}}
\newcommand{\statesSetFrac}{\mathcal{X}}

\newcommand{\reals}{\mathbb{R}}

\newcommand{\catalog}{\mathcal{N}}
\newcommand{\x}{\vec{x}}
\newcommand{\y}{\vec{y}}
\newcommand{\z}{\vec{z}}

\newcommand{\policy}{\mathcal{A}}

\newcommand{\Rounding}{\textsc{Online Rounding}\xspace}

\renewcommand{\new}[1]{{\textcolor{black}{#1}}}
\newcommand{\nnew}[1]{{#1}}

\usepackage{cancel}

\usepackage{amsmath,thmtools}
\declaretheorem[name=Theorem]{theorem}
\declaretheorem[name=Lemma]{lemma}
\declaretheorem[name=Definition]{definition}

\declaretheorem[name=Proposition]{proposition}
\declaretheorem[name=Corollary]{corollary}

\usepackage[font=footnotesize,labelfont=bf, textfont=footnotesize]{caption}
\usepackage[font=footnotesize,labelfont=bf, textfont=footnotesize]{subcaption}

\title{No-Regret Caching via Online Mirror Descent}

\author{Tareq Si Salem\\\texttt{tareq.si-salem@inria.fr}\\Inria, Universit\'e C\^ote d'Azur\\Sophia Antipolis\\France\\
\And  Giovanni Neglia\\\texttt{giovanni.neglia@inria.fr}\\Inria, Universit\'e C\^ote d'Azur\\Sophia Antipolis\\ France
\And Stratis Ioannidis \\ \texttt{ioannidis@ece.neu.edu} \\ Northeastern University \\ Boston \\ USA
}
\date{}
\begin{document}

\maketitle

\begin{abstract}
We study an online caching problem in which requests can be served by a local cache to avoid retrieval costs from a remote server. The cache can update its state after a batch of requests and store an arbitrarily small fraction of each file. We study no-regret algorithms based on Online Mirror Descent (OMD) strategies. \new{We show that  bounds for the regret crucially depend on the diversity of the request process, provided by the diversity ratio $R/h$, where $R$ is the size of the batch, and $h$ is the maximum multiplicity of a request in a given batch.} \new{We characterize the optimality of OMD caching policies w.r.t.
regret under different diversity regimes.} We also prove that, when the cache must store the entire file, rather than a fraction, OMD strategies can be coupled with a randomized rounding scheme that preserves regret guarantees, \new{even when update costs cannot be neglected.} \new{We provide a formal characterization of the rounding problem through optimal transport theory, and moreover we propose a computationally efficient randomized rounding scheme.}
\end{abstract}
%    <concept>
%        <concept_id>10003752.10003809.10010047.10010049</concept_id>
%        <concept_desc>Theory of computation~Caching and paging algorithms</concept_desc>
%        <concept_significance>500</concept_significance>
%        </concept>
%  </ccs2012>

% \ccsdesc[500]{Theory of computation~Caching and paging algorithms}

% \maketitle

\section{Introduction}
Caches are deployed at many different levels in computer systems: from CPU hardware caches to operating system memory caches, from application caches at clients to CDN caches deployed as physical servers in the network or as cloud services like Amazon's ElastiCache~\cite{amazonElastiCache}. They aim to provide  faster service to the user and/or to reduce the computation/communication load on other system elements, like hard disks, file servers, etc.

The ubiquity of caches has motivated extensive research on the performance of existing caching policies, as well as on the design of new policies with provable guarantees. To that end, most prior work has assumed that caches serve requests generated according to a stochastic process, ranging from the simple, memory-less independent reference model~\cite{coffman1973operating}  to more complex models trying to capture temporal locality effects and time-varying popularities (e.g., the shot-noise model~\cite{traverso13}).
An alternative modeling approach is to consider an \emph{adversarial} setting. Assuming that the sequence of requests is generated by an adversary, an online caching policy can be compared to the optimal offline policy that  views the sequence of requests in advance. 
Caching was indeed one of the first problems studied by Sleator and Tarjan in the context of the  competitive analysis of online algorithms~\cite{sleator85}.  In competitive analysis, the metric of interest is the \emph{competitive ratio}, i.e., the worst-case ratio between the costs incurred by the online algorithm  and the  optimal offline \emph{dynamic} algorithm. 
This line of work led to the study of metrical task systems~\cite{borodin92}, a popular research area in the algorithms community~\cite{koutsoupias09}. 

Recently, Paschos et al.~\cite{PaschosLearningToCache,paschos2020cachebook}  proposed studying caching  as an online convex optimization (OCO) problem~\cite{Hazanoco2016}. OCO considers again an adversarial setting, but the metric of interest is the \emph{regret}, i.e., the difference between the costs incurred over a time horizon $T$ by the algorithm and by the optimal offline \emph{static} solution. Online algorithms whose regret grows sublinearly with  $T$ are called \emph{no-regret} algorithms, as their time-average regret becomes negligible for large $T$.
% \new{; although in practice, classifying sublinear regret policies as good policies must be  made ``within reason'', since the constants appearing in the regret are of importance when targeting performance guarantees for a finite horizon $T$, e.g., applying naively learning from expert advice~\cite{Hazanoco2016} over the exponentially sized universe of cache configurations (each configuration is an expert) yields directly a sublinear regret policy, but with large regret constants, which makes it futile for finite horizons.}
Paschos et al.~proposed a no-regret caching policy based on the classic online gradient descent method ($\OGD$),  
under the assumption that (1) the cache can store arbitrarily small fractions of each file (the so-called fractional setting), and (2) the cache state is updated after each request.

In this paper, we extend and generalize the analysis of Paschos et al.~in three different directions:
\begin{enumerate}
    \item We assume the cache can update its state after  processing a batch of $R \ge 1$ requests. This is of interest both in high-demand settings, as well as in cases when updates are infrequent, because they are costly w.r.t.~either computation or communication.

    %This is a  it may happen when updates are particularly costly in terms of computation or communication; 
    \item \nnew{We consider a family of caching policies based on online mirror descent ($\OMD$); $\OGD$, employed by Paschos et al., is a special instance of this family. }
    \item We also depart from the fractional setting, extending our analysis to the case when the cache can only store entire files (the integral setting).
\end{enumerate}

\new{Batching is a generalization from the point of view of the practical application to caching: online algorithms applied to caching have considered until now a single request $R = 1$~\cite{PaschosLearningToCache, NEURIPS2021_2387337b, FundamentalLimits, mukhopadhyay2021online}, whereas in this work,  we consider a more general operation, and we recover the basic one for $R=1$.} \new{In particular, OCO learning algorithms applied to caching suffer from a time complexity that is dependent on the catalog size~\cite{Hazanoco2016}, which can be extremely large.  Therefore, despite their theoretical guarantees, their computational overhead is difficult to justify if requests are processed individually, especially  when cache updates are costly and can then occur only sporadically.
%it is difficult to justify their applicability when requests are processed individually, because they would generate either in high-demand settings, as well as in cases when updates are infrequent,
%because they are costly w.r.t. either computation or communication. 
However, this difficulty can be overcome through batching, where a batch includes the requests arriving between two consecutive cache updates.
%corresponds to the arriving requests
%while the policy is an intermediate state. 
%By this construction, 
Batching amortizes the computational cost  of the different policies, reducing the cost \emph{per request} by the batch size~$R$. Moreover, the batch size $R$   could simply be a characteristic of the  caching system instead of being a design choice.}

Our contributions are summarized as follows.
\nnew{First, applying the analysis of OMD by Bubeck~\cite{bubeck2015convexbook} to the caching setting, we show that the $\BigO{\sqrt T}$ regret of $\OGD$ observed by Paschos et al.~in the fractional setting extends to general $\OMD$ caching policies. We also show that constants in regret bounds  depend on the diversity of the request process.} In particular, the regret depends on the \emph{diversity ratio} $R/h$, where $R$ is the size of the batch, and $h$ is the maximum multiplicity of a request in a given batch.
Second, we characterize the optimality of OMD caching policies w.r.t.  regret under different diversity regimes. %and different cache sizes (normalized with respect to the catalog size). 
We observe that, for a large region of possible values of the diversity ratio, the optimum is either $\OGD$ or $\OMD$ with a neg-entropy mirror map ($\OMD_{\negent}$). In particular, $\OGD$ is optimal in the \emph{low diversity} regime, while $\OMD_{\negent}$ is optimal in the \emph{high diversity} regime. 
Third, $\OMD$ algorithms include a gradient update followed by a projection to guarantee that the new solution is in the feasible set (e.g.,  it does not violate the cache capacity constraints). The projection is often the most computationally expensive step of the algorithm. We show that efficient polynomial algorithms exist both for $\OGD$ (slightly improving the algorithm in~\cite{PaschosLearningToCache}) and for $\OMD_{\negent}$.
Finally, $\OMD$ algorithms work in a continuous space, and are therefore well-suited for the fractional setting originally studied by Paschos et al.  Still, we show that, if coupled with opportune rounding techniques, they can also be used when the cache can only store a file in its entirety, while preserving their regret guarantees. \new{{To the best of our knowledge, this is the first paper to provide a formal characterization of the randomized
rounding problem in caching, wherein the objective is to maintain the regret guarantees for the expected service
cost, while minimizing the update costs. This characterization casts the rounding problem as an optimal
transport problem in Sec.~\ref{s:rounding_schemes}. Moreover, we further prove that an opportune modification of Madow’s sampling~\cite{Madow1944Mar,NEURIPS2021_2387337b,geocaching2015, ioannidis2016adaptive} enables
to guarantee sublinear expected update costs.}}

The remainder of this paper is organized as follows. After an overview of the related work in Sec.~\ref{sec:related}, we introduce our model assumptions in Sec.~\ref{s:system} and provide  technical background on gradient algorithms in Sec.~\ref{sec:grad}. Section~\ref{sec:OMDanalysis} presents our main results on the regret of $\OMD$ caching policies and their computational complexity. A discussion about extending the model to include cache update costs, in Sec.~\ref{sec:movement}, is required to introduce the integral setting in Sec.~\ref{sec:rounding}.
Finally, numerical results are presented in Sec.~\ref{sec:experiments}.

\section{Related work}
\label{sec:related}
% Classical caching and what is done up to now.

The caching problem has been extensively studied in the literature under different assumptions on the request process.  When the requests occur according to a given stochastic process, 
the analysis leads usually to complex formulas even in simple settings. For example, even the hit ratio of a single cache managed by the LRU eviction policy under the  independent reference model %(at any time a request is for file $i$ with probability $p_i$ independently from the past), and the metric of interest is the miss ratio
is hard to precisely characterize~\cite{king72,flajolet92}.
 The characteristic time approximation (often referred to as Che's approximation)  significantly simplifies this analysis by assuming that a file, in absence of additional requests for it, stays in the cache for a random time %$\tau$
 sampled independently from requests for other files. %(but the distribution of $\tau$ depends on the request process). 
 %To the best of our knowledge, it was 
 Proposed by Fagin~\cite{fagin77} and rediscovered and popularized by~Che et al.~\cite{che02}, the approximation has been justified formally by several works ~\cite{Jele99,fricker2012,jiang18}
%The characteristic time approximation 
and has allowed the study of a large number of existing~\cite{garetto16} and new~\cite{gast2017ttl,leonardi18jsac} caching policies. It also applies to   networked settings~\cite{choungmo14,berger2014exact,alouf2016performance,chu18}
%In particular, new caching policies have been designed to optimize 
and to  more general utilities beyond the hit ratio~\cite{dehghan19,neglia18ton}, all under  stochastic requests. 

Online caching policies based on gradient methods have also been studied in the stochastic request setting, leading to Robbins-Monro/stochastic approximation algorithms (see, e.g., \cite{ioannidis2010distributed,ioannidis2016adaptive}). Though related to OCO, guarantees are very different than the regret metric we study here.
 Many works have also explored the offline, network-wide static allocation of files, presuming demand is known~\cite{borst2010distributed,shanmugam13,poularakis2016exploiting}.
We differ from the work above, as we consider adversarial requests.

%%%%%%%%%%% Adversarial, compare to dynamic optimum 
Caching under adversarial requests has been studied since Sleator and Tarjan's seminal paper~\cite{sleator85} through the competitive ratio metric. An algorithm is said to be $\alpha$-competitive when its competitive ratio is bounded by $\alpha$ over all possible input sequences. The problem has been generalized by Manasse et al.~\cite{manasse88}~under the name \emph{$k$-server problem}, and further generalized  by Borodin et al.~under the name \emph{metrical task systems (MTS)}~\cite{borodin92}. The literature on both the $k$-server and MTS problems is vast.
A recent trend is to apply continuous optimization techniques to solve these combinatorial problems. Bansal et al.~\cite{bansal12}  study the $k$-server problem on a weighted star metric space. In the same spirit, Bubeck et al.~\cite{bubeck_kserver_2018} use the framework of continuous online mirror descent to provide an $o(k)$-competitive algorithm for the $k$-server problem on  hierarchically separated trees. 
% \new{The competitive ratio approach is well understood for paging, and it is considered practically too pessimistic to provide a good classification of policies---e.g., LRU and any marking policy have the highest competitive ratio ~\cite{borodin2005online,PaschosLearningToCache}.} 
In this paper, we focus on regret rather than competitive ratio as \new{the} main performance metric.
Andrew et al.~\cite{taleoftwometrics} give a formal comparison between competitive ratio and regret and 
%competitive analysis and  OCO/no-regret algorithms.  
prove that there is an intrinsic incompatibility between the two: no algorithm can have both sub-linear regret and a constant competitive ratio. At the same time, they propose an algorithm with sub-linear regret and slowly increasing competitive ratio.

Online convex optimization (OCO)  was first proposed by Zinkevich~\cite{Zinkevich2003}, who showed that projected gradient descent attains sublinear regret bounds in the online setting. OCO generalizes previous online problems like the experts problem ~\cite{littlestone94}, and has become widely influential in the learning community~\cite{Hazanoco2016, ShalevOnlineLearning}.  
To the best of our knowledge, Paschos et al.~\cite{PaschosLearningToCache,paschos2020cachebook} were the first to apply the OCO framework to caching. Beside\new{s} proposing OGD for the single cache, they extended it to a simple networked scenario, where users have access to a set of parallel caches that store pseudo-random linear combinations of the files.  They proposed  no-regret algorithms in both settings. Bhattacharjee et al.~ \cite{FundamentalLimits} extended this work proving tighter lower bounds for the regret and proposing new caching policies for the networked setting that do not require file coding;  Mukhopadhyay and Sinha~\cite{mukhopadhyay2021online} accounted for switching costs due to file retrievals.  \nnew{Our work drops assumption A2 and A6 stated by  Bhattacharjee et al.~\cite{FundamentalLimits} under both fractional and integral caching settings, because we account for the update cost associated to changing the cache state, and moreover, we permit in our caching model to have  multiple requests be processed in a single timeslot $R\geq 1$. In particular, in  Sec.~\ref{sec:grad}, only assumptions A3--A5 are needed for the proposed algorithms $\OGD$ and $\OMD_{\negent}$, and in Sec.~\ref{sec:rounding}, we also require assumption A1, i.e., the cache can fetch files that are not necessarily requested in the previous timeslot.} \new{Paria and Sinha~\cite{NEURIPS2021_2387337b}  studied integral caching over bipartite network topologies. They employ a randomized rounding scheme (Madow's sampling~\cite{Madow1944Mar}) which is also the starting point for our rounding scheme (Online Rounding in Alg.~\ref{algo:online_rounding}), however, they only  provide  update cost guarantees under a strong stochastic regularity assumption over the request process. In this work, an opportune modification of the Madow’s sampling scheme, motivated by an optimal transport~\cite{peyre2019computational} formulation of the randomized rounding problem,  guarantees sublinear update cost
%enables a sublinear update cost guarantee 
even under adversarial requests.  
Li et al.~\cite{Li2021}, building on our proposed randomized rounding scheme, studied integral caching networks under arbitrary topology and adversarial requests.} We depart from these works in considering OMD algorithms, a more general request process, and allowing for integral cache states obtained through randomized rounding.

This work is an extension of our previous work~\cite{sisalemicc21}. In particular, (1) we analy\new{z}e and derive regret bounds for \nnew{a family of} OMD algorithms ($q$-norm mirror maps), and (2) we extend our analysis to the integral caching setting.

\section{System description}
\label{s:system}
\begin{table}[t!]

\begin{footnotesize}
\begin{tabular}{|C{.145\textwidth}C{.33\textwidth}|C{.12\textwidth}C{.3\textwidth}|}
     \hline
    & {\textbf{Notational Conventions}} & 	$\mathcal{R}_{R,h}$ & Set of possible adversarial requests \\
    \cline{1-2}
    $[n]$& Set of integers $\{1,2, \dots, n\}$  & $\vec{r}_t$ & Batch of request at timeslot $t$ \\
    \cline{1-2}
      & {\textbf{Caching}} & $f_{\vec{r}_t}$ & Cost received at timeslot $t$ \\
    \cline{1-2}
    $\mathcal{N}$ & Catalog set with size $\card{\catalog} = N$  & $\updatecost_{\vec{r}_t}$ & Update cost of the cache at timeslot $t$ \\ 
    $k$ & Cache capacity & $\vec{w}$ / $\vec{w}'$  & Service / update costs in $\mathbb{R}_+^N$\\
	\cline{3-4}
	$\mathcal{X}$ & Set of fractional cache states & & {\textbf{Online Learning}} \\
	\cline{3-4}
	$\mathcal{X}_\delta = \mathcal{X} \cap [\delta, 1]^N$ & The $\delta$-interior of $\mathcal{X}$ &  $T$ & The time horizon \\ 
	$\mathcal{Z} =  \mathcal{X} \cap \{0,1\}^N$ & Set of integral cache states & $\eta$ & Learning rate \\
	$\vec{x}_t$ & Fractional cache state at timeslot $t$ &   $\updatecost_{\vec{r}_t} (\vec{x}_{t}, \vec{x}_{t+1} ) $ & Update cost at timeslot $t$ \\
	$\vec \zeta_t$ & Integral cache state at timeslot $t$ & Regret$_T(\mathcal{A})$ & Regret of policy $\mathcal{A}$ over $T$\\
	$\vec z_t$ & Random integral cache state at timeslot $t$ & $\mathrm{E\mbox{-}Regret}_T(\mathcal{A}, \Xi) $ & Extended regret of policy $\mathcal{A}$ over $T$ \\ 
	$\vec{x}_*$ & Optimal cache allocation in hindsight & 	$\Phi(\vec{x})$ & Mirror map \\
	$R$ & Number of files' requests in a batch 	& $D_\Phi(\vec{x},\vec{y})$ & Bregman divergence associated to  $\Phi$\\ 
	$h$ & Maximum multiplicity of a requested file & $\Pi^{\Phi}_\mathcal{B}(\vec{y})$ & The projection onto $\mathcal{B}$ under $D_\Phi$\\
  \hline

\end{tabular}
\end{footnotesize}
\caption{{Notation Summary}\vspace{-2em}}
\end{table}
\noindent\textbf{Remote Service and Local Cache.} We consider a system in which requests for files are served either remotely or by an intermediate cache of finite capacity; a cache miss incurs a file-dependent remote retrieval cost. Formally, we consider a sequence of requests for files of equal size from a catalog $\mathcal{N}=\{1,2,\dots, N\}$.  These requests can be served by a remote server at cost $w_i \in \mathbb{R}_+$  per request for file $i\in\mathcal{N}$. This cost could be, e.g., an actual monetary cost for using the network infrastructure, or a quality of service cost incurred  due to fetching latency. Costs may  vary across files, as each file may be stored at a different remote location. We denote by %The cost of serving these files is summarized by a cost weight 
$\vec{w}=[w_i]_{i \in \mathcal{N}}\in \mathbb{R}_+^N$ the vector of costs %where $w_i \in \mathbb{R}_+$ is the cost of serving  file $i \in \mathcal{N}$. We assume that weights 
and assume that $\vec{w}$ is known.

 A local cache of finite capacity is placed in between the source of requests and the remote server(s). The local cache's role is to reduce the costs incurred by satisfying requests locally. We denote by $k \in \{1,2,\dots,N\}$ the capacity of the cache. The cache is allowed to store fractions of files (this assumption will be removed in Sec.~\ref{sec:rounding}).  We assume that time is slotted, and  denote by $x_{t,i}\in[0,1]$  the fraction of file $i \in \mathcal{N}$ stored in the cache at  timeslot $t \in \{1,2,\dots,T\}$.  The cache state is then given by vector $\vec{x}_t=[x_{t,i}]_{i\in \mathcal{N}}\in \mathcal{X}$, where $\mathcal{X}$ is the capped simplex  determined by the capacity constraint, i.e., $\mathcal{X} = \left\{ \vec{x} \in [0,1]^N : \sum^N_{i=1} x_i = k \right\}$.

\noindent\textbf{Requests.} We assume that a batch of multiple requests may arrive within a single timeslot.  The number of requests (i.e., the batch size) at each timeslot is given by $R \in \mathbb{N}$. A file may be requested multiple times (e.g., by different users, whose aggregated requests form the stream reaching the cache) within a single timeslot. We denote by $r_{t,i} \in \mathbb{N}$  the \emph{multiplicity} of file $i \in \mathcal{N}$, i.e., the number of requests for $i$,  at time $t$, and by $\vec{r}_t=[r_{t,i}]_{i\in\mathcal{N}} \in \mathbb{N}^N$ the vector of such requests, representing the entire batch.  We also assume that  the maximum multiplicity of a file in a batch is bounded  by $h \in \mathbb{N}$. As a result, $\vec{r}_t$ belongs to set $\mathcal{R}_{R,h} = \left\{\vec{r} \in \{0,\dots, h\}^N : \sum^N_{i=1} r_i = R \right\}.$

Intuitively, the ratio $\frac{R}{h}$ defines the diversity of request batches in a timeslot. For example, when $\frac{R}{h} = 1$, all $R$ requests are concentrated on a single file. When $\frac{R}{h} = N$, requests are spread evenly across the catalog $\mathcal{N}$. In general, $\frac{R}{h}$ is a lower bound for the number of distinct files requested in the batch. For that reason, we refer to $\frac{R}{h}$ as the \emph{diversity ratio}.\footnote{This definition of diversity is consistent with other notions of diversity, such as, e.g., the entropy; indeed the diversity ratio provides a lower bound on the entropy of the normalized batch vector $\frac{\vec{r}_t}{R}$, %\in \Delta_N$,
as $E\left(\frac{\vec{r}_t}{R}\right) \geq \log\left(\frac{R}{h}\right)$ \cite[Lemma 3]{lin2013diversity},
where $E(\vec p) = -\sum_i p_i \log(p_i)$ is the entropy function.} %Nevertheless, we adopt $\frac{R}{h}$ as a measure of diversity as, as we will see, it better characterizes system performance (see Thms.~\ref{xxx}-\ref{yyy}). 
We note that our request model
generalizes the setting  by Paschos et al. ~\cite{PaschosLearningToCache}, which can be seen as the case $R = h = 1$, i.e., the batch contains only one request per timeslot. 
We make no additional assumptions on the request arrival process; put differently, we operate in the adversarial online setting, where a potential adversary may select an arbitrary request sequence $\{\vec{r}_t\}_{t=1}^T$ in $\mathcal{R}_{R,h}$ to increase system costs. 

\noindent\textbf{Service Cost Objective.} When a request batch  $\vec{r}_t$ arrives, the cache incurs the following cost:
\begin{align}
\label{e:cost}
\textstyle f_{\vec{r}_t}(\vec{x}_t) = \sum^N_{i=1} w_i r_{t,i}(1 - x_{t,i}).\end{align}
In other words, for each file $i\in\mathcal{N}$, the system pays a cost proportional to the file fraction $(1-x_{t,i})$ missing from the local cache, weighted by the file cost $w_i$ and by the number of times $r_{t,i}$ file~$i$ is requested in the current batch $\vec{r}_t$.

The cost objective \eqref{e:cost} captures several possible real-life settings. First, it can be interpreted as a QoS cost paid by each user for the additional delay to retrieve part of the file from the server.  Second, assuming that the $R$ requests arrive and are served individually (e.g., because they are spread-out within a timeslot),  Eq.~\eqref{e:cost} can represent the load on  the servers or on the network to provide the missing part of the requested files.
Our model also applies when all requests for the same file are aggregated and served simultaneously  by a single fetch operation. In this case, $r_{t,i}$ in Eq.~\eqref{e:cost} should be interpreted as the indicator variable denoting if file $i$ was requested; correspondingly, $R$ then indicates the total number of \emph{distinct} files requested, and $h=1$.

\noindent\textbf{Online Caching Algorithms and Regret.} Cache files are determined online as follows. %: that is,
The cache has selected a state $\vec{x}_t \in \mathcal{X}$ at the beginning of a timeslot.\footnote{\new{We neglect the cost associated with the initial population of the cache since it is a fixed one-time cost.}}  The request batch $\vec{r}_t$ arrives, and the linear cost $f_{\vec{r}_t}(\vec{x}_t)$ is incurred; the state is subsequently updated to $\vec{x}_{t+1}$. Formally, the  cache state is determined by an online policy $\mathcal{A}$, i.e.,  a sequence of mappings  $\{\mathcal{A}_t\}^{T-1}_{t=1}$, where for every $t \geq 1$, $\mathcal{A}_t : (\mathcal{R}_{R,h} \times \mathcal{X})^{t} \to \mathcal{X}$ maps the sequence of past request batches and decisions $\{(\vec{r}_s, \vec{x}_s)\}^{t}_{s=1}$ to the next state $\vec{x}_{t+1}\in \mathcal{X}$. We assume that the policy starts from a feasible state $\vec{x}_1\in \mathcal{X}$. 

 We measure the performance of an online algorithm $\mathcal{A}$ in terms of regret, i.e., the difference between the total cost experienced by a policy $\mathcal{A}$ over a time horizon $T$ and that of the best static state $\vec{x}_*$ in hindsight. Formally,
\begin{align}
\textstyle \text{Regret}_T(\mathcal{A}) = \underset{\{\vec{r}_1, \vec{r}_2, \dots, \vec{r}_t\} \in \mathcal{R}^T_{R,h}}{\sup} \left\{ \sum^T_{t=1} f_{\vec{r}_t}(\vec{x}_t)- \sum^T_{t=1} f_{\vec{r}_t}(\vec{x}_*) \right\}, 
\label{eq:regret}
\end{align}
where
$\vec{x}_*=\argmin_{\vec{x} \in \mathcal{X}}\sum^T_{t=1} f_{\vec{r}_t}(\vec{x})$ 
 is the optimal static cache state (in hindsight). 
Note that, by taking the supremum in Eq.~\eqref{eq:regret}, we indeed measure regret in the adversarial setting, i.e., against an adversary that potentially picks requests in $\mathcal{R}_{R,h}$ trying to jeopardize  cache performance.

\noindent\textbf{Update Costs.} An online algorithm $\mathcal{A}$ updating the cache state at timeslot $t$ may require moving a portion of a file from a remote server to the cache to implement this update. The update cost of the online algorithm is  not explicitly modeled in our cost  and  regret (Eqs.~\eqref{e:cost} and~\eqref{eq:regret}, respectively).
% This happens for two reasons. 
% First, the gradient-based algorithms we consider (see Sec.~\ref{sec:grad}) adapt the state by fetching fractions of files scaled by a gain $\eta=\BigO{{1}/{\sqrt{T}}}$; hence, when the time horizon is large,    file retrievals  satisfying requests would dominate this term.
% These small costs are  be fully eliminated in practice. 
We postpone the discussion of such cost in Sec.~\ref{sec:movement}. For the moment we observe that updates come ``for free'' for files requested in the current timeslot. 
% For example, increasing the fraction $x_{t,i}$ for some file~$i$ such that $r_{t,i}>0$, can be performed by recovering the additional part
% %a small ($\BigO{\eta}$) fraction 
% out of the  ($1-x_{t,i}$) missing fraction that needs to be retrieved to serve the file; updates can thus ``free-ride'' on regular traffic, at no additional cost.
% %This is because updates come ``for free'' for files requested in the current timeslot. 
% %For example, increasing a coordinate $x_{t,i}$ for some file $i$ such that $r_{t,i}>0$, can be performed by recovering a small ($\BigO{\eta}$) fraction out of the overall ($1-x_i$) missing part of the file at the time of its retrieval; updates can thus ``free-ride'' on regular traffic, at no additional cost. 
% % As we discuss in Sec.~\ref{sec:movement}, 
\new{The main algorithms studied in this paper ($\OGD$ and $\OMD_{\negent}$) implement cache updates by fetching parts of files that have been requested in the previous timeslot. As a result, to implement these updates we can piggyback the traffic created to serve the user, and the cost of this traffic is already  accounted for in our service cost model~\eqref{e:cost}. As a result, the update cost is zero (see also Proposition~\ref{prop:free_updates}).} \new{We note that this property does not hold} \new{for randomized integral caching  policies in Sec.~\ref{sec:rounding}, which may require to store files that have not been requested.}

% We prove in Proposition~\ref{prop:free_updates} that the main algorithms studied in this paper ($\OGD$ and $\OMD_{\negent}$) are in this regime, as  \emph{they only increase current state coordinates corresponding to files requested in the previous timeslot}; as such, their update costs can be considered to be zero. 
% \old{There can be cases however where update costs cannot be ignored; we revisit this issue in Secs.~\ref{sec:movement} and~\ref{sec:rounding}, where we account for update costs explicitly.} 
% \new{ The same argument, unfortunately, cannot be transferred to the setting where  the fractional states are interpreted as probabilities of storing a given file; we revisit this issue in Secs.~\ref{sec:movement} and~\ref{sec:rounding}, where we account for update costs explicitly. In what follows, we consider deterministic cache states.}

\section{Fractional Caching and  Gradient-based Algorithms}\label{sec:grad}
% Any policy that updates the cache state by deterministic evictions and insertions of files in their entirety is too weak against an adversary and has worst case regret (see Proposition~\ref{proposition:deterministic_worst_case_regret}); therefore, classical policies, such as LRU, LFU, FIFO, etc., are all rendered obsolete in this setting. This observation, suggests that policies with low regret can only emerge from policies employing  fractional caching, or  randomization. 

 Inspired by offline minimization, it is natural to design a policy that, upon seeing $\vec{r}_t$,  selects as $\vec{x}_{t+1}$ the state that would have minimized (on hindsight) the aggregate cost up to time $t$ (i.e., $\sum_{t'=1}^{t}f_{\vec{r}_{t'}}(\vec{x})$). Unfortunately, such \new a policy has poor regret: 
 \begin{proposition}
\label{proposition:aggregate_cost_policy}
The aggregate cost minimization policy is a policy $\policy$ that selects for every timeslot $t \in [T-1]$ the state $\x_{t+1} = \argmin_{\x \in \statesSetFrac} \sum_{t'=1}^{t}f_{\vec{r}_{t'}}(\vec{x})$. This policy has linear (worst-case) regret, i.e., $\Regret(\policy) = \BigOmega{T}$.
\end{proposition}
% We provide a proof in Appendix~\ref{proof:aggregate_cost_policy}.
\new{The proof follows the same argument of Shalev-Shwartz~\cite[Example~2.2]{ShalevOnlineLearning}.} A more conservative approach, that indeed leads to sublinear regret, is to take gradual steps, moving in the direction of a better decision according to the latest cost; we present algorithms of this nature in this section.

% \subsection{Gradient-based Algorithms}

\subsection{Online Gradient Descent (OGD)}
\label{section:OGD}
In OGD, introduced by Paschos et al. \cite{PaschosLearningToCache} for  online caching, the cache  is  initialized with a feasible state $\vec{x}_1\in \mathcal{X}$ and updated as follows.  Upon receiving a request batch $\vec{r}_t$, the cost $f_{\vec{r}_t}(\vec{x}_t)$ is incurred and the next state becomes:
\begin{align} \vec{x}_{t+1} = \Pi_\mathcal{X} \left(\vec{x}_t - \eta \nabla f_{\vec{r}_t}(\vec{x}_t)\right), \quad\text{for all}~t \in [T-1],\label{eq:ogd}\end{align}
where $\Pi_\mathcal{X}(\,\cdot\,)$ is the Euclidean projection onto $\mathcal{X}$, that ensures feasibility, and $\eta\in \mathbb{R}_+$ is called the learning rate. %OGD 
%The caching decision $\vec{x}_t$ is updated by an $\eta$ movement. We take a step in the direction of the gradient $\nabla f_{\vec{r}_t} (\vec{x}_t)$ at time $t$. Since this update step could lead to an invalid caching configuration, we are required to find the closest point $\vec{x}_{t+1}$ in the decision set $\mathcal{X}$. We use the Euclidean projection to obtain a feasible point, where $\Pi_\mathcal{X}(.)$ is the Euclidean projection onto $\mathcal{X}$. 
Note that the state $\vec{x}_{t+1}$ obtained according \new{to}  Eq.~\eqref{eq:ogd}  is indeed a function of  $\{(\vec{r}_{t}, \vec{x}_{t})\} \subset \{(\vec{r}_{s}, \vec{x}_{s})\}^{t}_{s=1} $ for every $t \geq 1$; hence, OGD is indeed an online caching policy as defined in Sec.~\ref{s:system}. Paschos et al. \cite{PaschosLearningToCache} show that OGD attains sub-linear regret when $R=h=1$; more specifically: %the following holds: %The cache performs as well as the best fixed caching decision as $T\to \infty$, even with the presence of an adversary.
\begin{theorem}%Paschos et al.~\cite{PaschosLearningToCache}]%
\!\!\!{\protect\textrm{(\cite[Theorem 2]{PaschosLearningToCache})}}
\label{theorem:pascos_regret}
When $R = h = 1$,   the regret of OGD is bounded as follows:
\begin{equation}
    \Regret_T(\OGD) \leq \norm{\vec w}_\infty \sqrt{ \min(2k ,2(N-k)) T}.
\end{equation}
\end{theorem}
% There are different ways we could project back to the decision set besides the Euclidean projection and perhaps more efficiently, for instance, one could take advantage of the geometry of $\mathcal{X}$. Furthermore, the OGD algorithm is unaware of the sparsity of cost gradients that could be used as an advantage to take better decisions, this is why focus our attention on the class of Online Mirror Descent (OMD) strategies described in section \ref{section:OMD}.

In other words, OGD attains an $\BigO{\sqrt{T}}$ regret when $R=h=1$. In this paper, we study a broader class of gradient descent algorithms that include OGD as a special case.  As we will see below (see Thm.~\ref{theorem:optimalq_q1}), the regret attained by OGD is not necessarily the tightest possible when $R\neq 1 \neq h$; broadening the  class  of algorithms we consider  allows us to improve upon this bound. 
%In the next section, we present a generalization of the OGD Algorithm. We show that we can obtain stronger regret guarantees, with this larger space of algorithms.
\subsection{Online Mirror Descent (OMD)}
\label{section:OMD}
%The class of OMD strategies contains the OGD algorithm as a particular instance, and OMD provides a generalization and the flexibility to adapt a strategy depending on the geometry of the constraint set and the boundedness of the loss gradients,
%  \begin{minipage}{.5\textwidth}
\begin{footnotesize}
    \begin{algorithm}[t]
    % \begin{multicols}{1}
    % \end{multicols}
    % \begin{footnotesize}
    % \end{footnotesize}
	\begin{algorithmic}[1]\small
% 		\Procedure {OnlineMirrorDescent}{$\vec{x}_1 \gets \underset{\vec{x} \in \mathcal{X} \cap \mathcal{D}}{\arg\min} \,\Phi(\vec{x}) , \eta$}
\Require $\vec{x}_1 = \underset{\vec{x} \in \mathcal{X} \cap \mathcal{D}}{\arg\min} \,\Phi(\vec{x})$ , $\eta \in \reals_+$
		\For{$t \gets 1,2,\dots,T$}\Comment{{ Incur a cost $f_{\vec{r}_t}(\vec{x}_t)$, and receive a gradient  $  \nabla f_{\vec{r}_t}(\vec{x})$}}
		\State
		$\hat{\vec{x}}_t\gets \nabla \Phi (\vec{x}_t)$\Comment{{ Map primal point to dual point}}
		\State
		$\hat{\vec{y}}_{t+1}   \gets \hat{\vec{x}}_t - \eta \nabla f_{\vec{r}_t}(\vec{x}_t)$\Comment{{ Take gradient step in the dual space}}
		\State
		$\vec{y}_{t+1}   \gets \left(\nabla \Phi\right)^{-1}(\hat{\vec{y}}_{t+1})$\Comment{{ Map dual point to a primal point}}
		\State
		$\vec{x}_{t+1}\gets \Pi_{\mathcal{X} \cap \mathcal{D}}^\Phi(\vec{y}_{t+1})$\Comment{{ Project new point onto feasible region $\mathcal{X}$}}
		\EndFor
% 		\EndProcedure
	\end{algorithmic}
% 	\end{multicols}
	\caption{Online mirror descent ($\text{OMD}_\Phi$)}
	\label{algo:online_mirror_descent}
\end{algorithm}
	\end{footnotesize}
% \end{minipage}

OMD \cite[Sec.~5.3]{Hazanoco2016} is the online version of the mirror descent (MD) algorithm \cite{beck2003mirror} for convex optimization of a fixed, known function. The main premise behind mirror descent is that variables and gradients live in two distinct spaces: the \emph{primal space}, for variables, and the \emph{dual space}, for gradients. The two are linked via a function known as a \emph{mirror map}. Contrary to standard gradient descent, updates using the gradient occur on the dual space; the mirror map is used to invert this update to a change on the primal variables. For several constrained optimization problems of interest, mirror descent leads to faster convergence compared to gradient descent \cite[Sec. 4.3]{bubeck2015convexbook}. OMD arises by observing that MD is agnostic to whether the gradients are obtained from a \emph{fixed} function, or a sequence revealed adversarially.

\noindent\textbf{OMD for Caching.} Applied to our caching problem, OMD takes the form summarized in Algorithm~\ref{algo:online_mirror_descent}. %performs a gradient descent update, similarly to OGD. However, the decision variable is mapped to a \emph{dual space}, before performing the gradient descent step. The resulting dual point is mapped to the \emph{primal space}. Since this point can lie outside of $\mathcal{X}$, we also require a projection step. 
In our case, both the primal and dual spaces are $\mathbb{R}^N$. To disambiguate between the two, we denote primal points by $\vec{x}, \vec{y} \in \mathbb{R}^N$   and dual points by  $\hat{\vec{x}},\hat{\vec{y}} \in \mathbb{R}^N$, respectively.   Formally, OMD is parameterized by (1) a fixed learning rate $\eta\in \mathbb{R}_+$, and (2) a  differentiable map $\Phi : \mathcal{D} \to \mathbb{R}$, strictly convex over $\mathcal D$ and $\rho$-strongly convex over $\mathcal{X} \cap \mathcal{D}$, where $\mathcal{X}$ is included in the closure of $\mathcal{D}$; that is %$\mathcal{X} \subseteq \mathrm{closure}(\mathcal{D})$. 
\begin{align}\mathcal{X} \subseteq \mathrm{closure}(\mathcal{D}).\label{eq:closure} \end{align}
Function $\Phi$ is called the \emph{mirror map}, that links the primal to the dual space.  

Given $\eta$ and $\Phi$, an OMD iteration proceeds as follows. After observing  the request batch $\vec{r}_t$ and incurring the cost $f_{\vec{r}_t}(\vec{x}_t)$, the current state $\vec{x}_t$ is first mapped from the primal  to the dual space via: 
\begin{align}\label{eq:mapstep}\hat{\vec{x}}_{t} = \nabla \Phi(\vec{x}_t).\end{align} Then, a regular gradient descent step is performed \emph{in the dual space} to obtain an updated dual point: \begin{align}\hat{\vec{y}}_{t+1}   = \hat{\vec{x}}_t - \eta \nabla f_{\vec{r}_t}(\vec{x}_t). \end{align} This updated dual point is then mapped back to the primal space using the inverse  of mapping $\nabla \Phi$, i.e.:
\begin{align}\label{eq:invstep}\vec{y}_{t+1} = \left(\nabla \Phi\right)^{-1}\!(\hat{\vec{y}}_{t+1}).\end{align}
The resulting primal point $\vec{y}_{t+1}$ may lie outside the constraint set $\mathcal{X}$. To obtain the final feasible point $\vec{x}_{t+1}\in \mathcal{X}$, a projection is made using the Bregman divergence associated with the mirror map~$\Phi$; that is, instead of the orthogonal projection used in OGD, the final cache state becomes:
\begin{align}\label{eq:bregprojstep}
\vec{x}_{t+1}= \Pi_{\mathcal{X} \cap \mathcal{D}}^\Phi(\vec{y}_{t+1}),
\end{align}
where $\Pi_{\mathcal{X} \cap \mathcal{D}}^\Phi(\,\cdot\,)$ is the Bregman projection, which we define formally below, in Definition~ \ref{def:Bregman_projection}. %instead of the Euclidean distance. 

Together, steps \eqref{eq:mapstep}--\eqref{eq:bregprojstep} define OMD. Note that, as it was the case for OGD, $\vec{x}_{t+1}$ is a function of $\{(\vec{r}_{t}, \vec{x}_{t})\} \subset \{(\vec{r}_{s}, \vec{x}_{s})\}^{t}_{s=1} $, hence OMD is indeed an online algorithm.
Two additional technical assumptions on $\Phi$ and $\mathcal{D}$ must hold for steps \eqref{eq:invstep} and \eqref{eq:bregprojstep} to be well-defined.\footnote{All hold for the algorithms we consider in Sec.~\ref{sec:OMDanalysis}.}
First, the gradient of~$\Phi$ must diverge at the boundary of $\mathcal{D}$; this, along with strict convexity, ensures the existence and uniqueness of the Bregman projection in~\eqref{eq:bregprojstep}. Second, the image of $\mathcal{D}$ under the gradient of $\Phi$ should take all  possible values, that is $\nabla\Phi(\mathcal{D}) = \mathbb{R}^N$; this, along again with strict convexity, ensures that $\nabla\Phi$ is one-to-one and onto, so its inverse exists and Eq.~\eqref{eq:invstep} is well-defined.

%We can see that OMD for every $t\geq 2$, 
Setting $\Phi(\vec{x}) = \frac{1}{2} \norm{\vec{x}}^2_2$ and $\mathcal{D} = \mathbb{R}^N$ yields the identity mapping $\nabla \Phi(\vec{x}) = \vec{x},$ for all $\vec x \in \mathcal{D}$. Furthermore, the Bregman divergence associated with this map is just the Euclidean distance $D_\Phi (\vec{x}, \vec{y}) = \frac{1}{2}\norm{\vec{x} -\vec{y}}^2_2 $. Thus, this Euclidean version of OMD is equivalent to OGD, and OMD can be seen as a generalization of the OGD to other mirror maps.

%\noindent\textbf{Bregman Projection.} 
To conclude our description of OMD, we define the Bregman projection \cite{kiwiel1997proximal}.
\begin{definition}
\label{def:Bregman_projection}
    The Bregman projection denoted by $\Pi^{\Phi}_{\mathcal{X} \cap \mathcal{D}}:\mathbb{R}^N\to\mathcal{X} \cap \mathcal{D}$, is defined as
    \begin{align}
        \Pi^{\Phi}_{\mathcal{X} \cap \mathcal{D}}(\vec{y}) &= \underset{\vec{x} \in {\mathcal{X} \cap \mathcal{D}}}{\arg\min}\,D_\Phi (\vec{x},\vec{y}), & \text{where}& &  D_\Phi(\vec{x},\vec{y}) = \Phi(\vec{x}) - \Phi(\vec{y}) - \nabla {\Phi(\vec{y})}^T (\vec{x} - \vec{y})
    \end{align}
 is the Bregman divergence associated with the mirror map $\Phi$. 
\end{definition}

%The mapping $\vec{x} \to D_\Phi(\vec{x},\vec{y})$ is strictly convex, and locally increasing at the boundary of $\mathcal{D}$, this implies the existence and the uniqueness of this projection. 

\subsection{Analysis of Online Mirror Descent Algorithms}\label{sec:OMDanalysis}

We present our main results regarding the application of  OMD under several different mirror maps to the online caching problems. 
%We present our main results regarding the application of different OMD algorithms to the online caching problems. 
%To that end, we discuss the performance of OMD under several different mirror maps. %
We will be concerned with both (1) the regret attained, and (2)~computational complexity issues, particularly pertaining to the associated Bregman projection. 
Our key observation is that \emph{the regret of different algorithms is significantly influenced by  demand diversity, as captured by the diversity ratio $\frac{R}{h}$}. In particular, our analysis allows us to characterize regimes of the diversity ratio in which OGD outperforms other mirror maps, and vice versa.

\subsection{$q$-Norm Mirror Maps}
\label{subsection:q-norm-mirror-maps}
A natural generalization of the OGD algorithm to a broader class of OMD algorithms is via $q$-norm mirror maps, whereby: 
\begin{align}
     \Phi(\vec{x}) = \frac{1}{2}\norm{\vec{x}}^2_q, \quad\text{where }q \in (1,2],~\text{and }   \quad \mathcal{D}&=\mathbb{R}^N.
\label{eq:qnorm_mirrormap}
\end{align}
It is easy to verify that $\Phi$ and $\mathcal{D}$, defined as above, satisfy all technical requirements set in Sec.~\ref{section:OMD} on a mirror map and its domain.
We define $\OMD_{q\nrm}$ to be the OMD Algorithm $\ref{algo:online_mirror_descent}$ with $\Phi$ and $q$ given by Eq.~\eqref{eq:qnorm_mirrormap}.
Note that this map generalizes OGD, which corresponds to the special case $q=2$. 
In what follows, we denote by $\|\cdot\|_p$ the dual norm of $\|\cdot\|_q$. Then, $p \in [2, \infty)$ is such that $\frac{1}{p} + \frac{1}{q} = 1$. Note that sometimes $\OMD_{q\nrm}$
% OMD strategies generated by this class of $q$-norm mirror maps 
is referred to as  a \emph{$p$-norm algorithm}~\cite{ShalevOnlineLearning}. %, where $p= \frac{1}{1 - 1/q} \in [2, \infty)$~

\subsubsection{Regret Analysis} We begin by providing a regret bound for $\OMD_{q\nrm}$   algorithms: %in the context of online caching:
\begin{theorem}
\label{theorem:qnorm_regret}
% $\eta = \sqrt{\frac{(q-1) k^2 \left(k^{-\frac{2}{p}}  - N^{-\frac{2}{p}}\right)}{\norm{\vec w}^2_\infty h^2 \left(\frac{R}{h}\right)^{\frac{2}{p}} T}}$%
For $\eta = \textstyle\sqrt{\frac{(q-1) k^2 \left(k^{-\frac{2}{p}}  - N^{-\frac{2}{p}}\right)}{\norm{\vec w}^2_\infty h^2 \left(\frac{R}{h}\right)^{\frac{2}{p}} T}}$, the regret of $\OMD_{q\nrm}$  over $\mathcal{X}$ satisfies:
\begin{equation}
    \textstyle \Regret_T(\OMD_{q\nrm}) \leq    \textstyle \norm{\vec w}_\infty h k \left(\frac{R}{h}\right)^{\frac{1}{p}}\sqrt{\frac{1}{q-1}  \left(k^{-\frac{2}{p}} - N^{-\frac{2}{p}}\right) T}.
    \label{eq:qnorm_regret}
\end{equation}
\end{theorem}
\noindent The proof can be found in Appendix \ref{proof:qnorm_regret}. We use an upper bound on the regret of general OMD from %Bubeck 
\cite[Theorem 4.2]{bubeck2015convexbook} and relate it to our setting; in doing so, we  bound the diameter of~$\mathcal{X}$ w.r.t. Bregman divergence under $\Phi$ as well as the dual-norm $\norm{\,\cdot\,}_p$ of the gradients $\nabla f_{\vec{r}_t}(\vec{x}_t)$. %The mirror map is strongly convex w.r.t the norm $\norm{.}_q$, and the Bregman divergence of $\vec{x}_1$ and $\vec{x}_*$ is bounded under $\Phi$. 

\begin{SCfigure}[50][t]
\centering
   \includegraphics[width = 100pt]{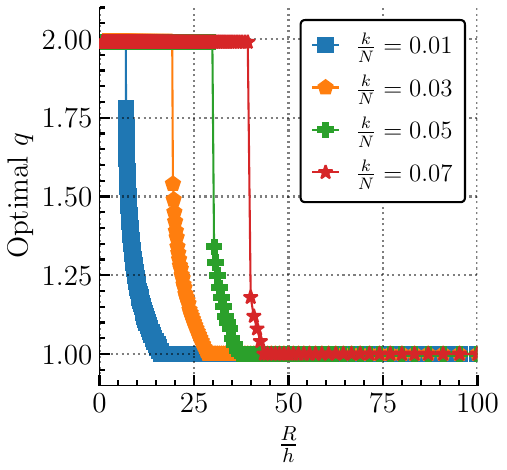}
    \caption{\protect\rule{0ex}{-5ex}Numerical characterization  of  $q^*\in [1,2]$  as a function of the diversity ratio $R/h$, for different cache capacities $k$ expressed as fractions of the catalog size ($N=100$). Given $R/h$, the optimal $q^*$ is determined as the value in $[1,2]$ that minimizes the upper-bound in Eq.~\eqref{eq:qnorm_regret}. Higher values of $R/h$ represent more diverse requests. Under small diversity, OGD is optimal; as diversity increases, mirror maps for which $q<2$ attain a more favorable upper bound than OGD.}  \label{fig:diversity_regimes}  
    \vspace{-1.5em}
\end{SCfigure}

Comparing Theorem~\ref{theorem:qnorm_regret} to Theorem~\ref{theorem:pascos_regret}, we see that both attain an $\BigO{\sqrt{T}}$ regret. A natural question to ask when comparing the two bounds is whether there are cases where $\OMD_{q\nrm}$ with $q\neq 2$ outperforms OGD (i.e.,  $\OMD_{2\nrm}$).  The constants in the r.h.s.~of Eq.~\eqref{eq:qnorm_regret} depend on the diversity ratio $\frac{R}{h}$; this, in turn, affects  which is the optimal $q$, i.e., the one that minimizes the bound in Eq.~\eqref{eq:qnorm_regret}. 
%
%In particular, given a diversity ratio $\frac{R}{h}$, we can use the r.h.s.~of Eq.~\eqref{eq:qnorm_regret} to determine the optimal $q$, i.e, the $q^*\in [1,2]$ that minimizes the upper bound on the regret. 
Let $q^*=\mathop{\arg\inf}_{q\in (1,2]} \mathtt{ub}(q)$ be the optimal $q$, where $\mathtt{ub}:(1,2] \to \reals_+$ is the upper bound in Eq.~\eqref{eq:qnorm_regret}. Note that $q^* \in  [1,2]$. Figure \ref{fig:diversity_regimes} shows $q^*$ as a function of the diversity ratio, for different values of cache capacity $k$.  We observe that OGD ($q=2$) is optimal for lower diversity regimes and larger caches; when diversity $\frac{R}{h}$ increases or cache capacity $k$ decreases, values $q<2$ become optimal. The transition from $q^*=2$ to $q^*=1$ is sharp, and becomes sharper as $k$ increases. 
\sidecaptionvpos{figure}{c}
%  \begin{minipage}[c]{0.3\textwidth}
      
%   \end{minipage}\hfill
%   \begin{minipage}[c]{0.3\textwidth}
  
%   \end{minipage}

% \begin{figure}[!t]
%     \centering
%     \includegraphics[width = 130pt]{graphical_res/Aug/optimal_q.pdf}
%     \caption{Numerical characterization of  $q^*\in [1,2]$  as a function of the diversity ratio $R/h$, for different cache capacities $k$ expressed as fractions of the catalog size ($N=100$). Given $R/h$, the optimal $q^*$ is determined as the value in $[1,2]$ that minimizes the upper-bound in Eq.~\eqref{eq:qnorm_regret}. Higher values of $R/h$ represent more diverse requests. Under small diversity, OGD is optimal; as diversity increases, mirror maps for which $q<2$ attain a more favorable upper bound than OGD.}
%     \label{fig:diversity_regimes}
% \end{figure}
\subsubsection{Optimality Regimes.}\label{sec:optregimes}
Motivated by these observations, we turn our attention to formally characterizing the two regimes under which optimality transitions from $q^*=2$ to $q^*=1$.
We first determine the upper bound on the regret for these two regimes. Indeed,
by setting $q=2$ in Theorem~\ref{theorem:qnorm_regret}, we obtain the following bound, generalizing Theorem~\ref{theorem:pascos_regret} to the case $R/h > 1$:
\begin{corollary}
\label{corollary:regret_qnorm_q2}
For $\eta  = \textstyle \sqrt{\frac{k \left(1 - \frac{k}{N}\right)}{ \norm{\vec w}^{2}_\infty  h R T}}$ the regret of OGD, satisfies:
\begin{equation}
\textstyle  \Regret_T(\OGD) \leq \norm{\vec{w}}_\infty\sqrt{ hR k \left(1 - {k}/{N}\right) T}.
\label{eq:regret_qnorm_q2}
\end{equation}
\end{corollary}
\noindent This a direct consequence of Theorem~\ref{theorem:qnorm_regret} by replacing $q= 2$ in Eq.~\eqref{eq:qnorm_regret}. 
We note that, in this result, we tighten the bound of Paschos et al.~\cite{PaschosLearningToCache}: for $R=h=1$, the bound in Eq.~\eqref{eq:regret_qnorm_q2} is smaller than  the one in Theorem~\ref{theorem:pascos_regret} by  at least a $\sqrt{2}$ factor.

%The fact that $\text{OMD}_{q\text{-norm}}$ interpolates between additive and multiplicative algorithms for $q=2$ and $q\to1$ respectively, motivates us to provide the other bound for the case of $\text{OMD}_{1\text{-norm}}$. We will further demonstrate why the two corner settings of $\text{OMD}_{q\text{-norm}}$ are of importance in our analysis in Sec. \ref{sec:diversity_and_regret}.

We also characterize the limiting behavior of $\OMD_{q\nrm}$ as $q$ converges to $1$. 
\begin{corollary}
\label{corollary:regret_qnorm_q1}
As $q$ converges to $1$, the upper bound on $\OMD_{q\nrm}$ regret given by Eq.~\eqref{eq:qnorm_regret} converges to:
\begin{equation}
\textstyle  \norm{\vec w}_\infty h k \sqrt{ 2\log\left({N}/{k}\right) T}.
\label{eq:regret_qnorm_q1}
\end{equation}
\end{corollary}
%\noindent The regret bound in Eq. \eqref{eq:qnorm_regret} is not defined for $q=1$. Taking the limit $q$ converges to $1$, this gives the desired result. 
The proof can be found in Appendix \ref{proof:regret_qnorm_q1}.  This  limit is precisely the bound on the regret attained under the neg-entropy mirror map (see Theorem~\ref{theorem:regret_negentropy} below). Armed with Corollaries~\ref{corollary:regret_qnorm_q2} and~\ref{corollary:regret_qnorm_q1}, we can formally characterize the regimes in which either of the two strategies become dominant:
%\subsection{Diversity and regret tightness}
%\label{sec:diversity_and_regret}
%We provide analytical regimes of the optimal $q^*$. Theorem \ref{theorem:optimalq_q2} specifies the regime where $q=2$ should be selected to obtain the tightest regret in the class of $q$-norm mirror maps OMD, and in Theorem \ref{theorem:optimalq_q1}, we give the regime where $q\to1$ should be picked instead. 
%In particular, the following result holds:
\begin{theorem}
\label{theorem:optimalq_q2}
The regret bound for  $\OMD_{q\nrm}$ in Eq.~\eqref{eq:qnorm_regret} is minimized for $q = 2$, when $\frac{R}{h} \leq k$.
% OGD, i.e., $\OMD_{q\nrm}$ for $q=2$, is optimal in the class of $\OMD_{q\nrm}$ algorithms when $\frac{R}{h} \leq k$.
\end{theorem}
In other words, when the diversity ratio is smaller than the cache size, it is preferable to update the cache via OGD. The proof, in Appendix \ref{proof:optimalq_q2}, establishes that the upper bound in Eq.~\eqref{eq:qnorm_regret} is monotonically decreasing w.r.t $q$ in the specified interval $\frac{R}{h} \leq k$. 
%This is achieved by showing that the first derivative of Eq.~\eqref{eq:qnorm_regret} is upper-bounded by a negative factor.
Our next result characterizes then the neg-entropy ($q$ converges to $1$) mirror map outperforms OGD:
\begin{theorem}
\label{theorem:optimalq_q1}
The limit, as $q$ converges to $1$, of the  $\OMD_{q\nrm}$ regret bound in Eq.~\eqref{eq:regret_qnorm_q1} is smaller than the corresponding bound for 
%tighter than , yields a tighter regret than  
OGD ($\OMD_{q\nrm}$ with $q=2$) when $\frac{R}{h} > 2 \sqrt{N k}$.
\end{theorem}
The proof is provided in Appendix \ref{proof:optimalq_q1}. 
%G: REMOVED TO GAIN SPACE. ALSO WHAT WE SAY IS QUITE TRIVIAL
%The result is obtained by relaxing the regret bound in Eq.~\eqref{eq:regret_qnorm_q1} using a sharp upper bound, which is subsequently compared to Eq.~\eqref{eq:regret_qnorm_q2}. 
We stress that Theorem~\ref{theorem:optimalq_q1} implies the sub-optimality of OGD in the regime $\frac{R}{h} > 2 \sqrt{N k}$. The experiments in Fig.~\ref{fig:diversity_regimes} suggest the bound in Theorem~\ref{theorem:optimalq_q1} is quite tight: for example for $k=7$ the bounds suggest $q=1$ should be optimal when $R/h$ exceeds $2 \sqrt{100 \times 7} \approx 52.9$, while experiments show that it is optimal when $R/h$ exceeds $45$. On the contrary, we observe that the bound in  Theorem~\ref{theorem:optimalq_q2} seems to be loose and the transitions we observe in Fig.~\ref{fig:diversity_regimes} are sharper than what one would predict from the bounds.

\subsubsection{Dual-Primal Update and Bregman Projection}\label{sec:bregmanomdq}

Having characterized the regret of $\OMD_{q\nrm}$ algorithms, we turn our attention to implementation issues. 
The map to the dual space and back in
%is obtained by replacing 
Eq.~\eqref{eq:mapstep} and Eq.~\eqref{eq:invstep}  (Lines 2 and 4 in Algorithm~\ref{algo:online_mirror_descent}), have the following expression~\cite{Gentile_pnorm}, respectively: 
\begin{align}
\label{eq:q-norm-update}
& \hat{x}_{t,i} = \textstyle \left(\nabla \Phi(\vec{x}_t)\right)_i = \text{sign} (x_{t,i}){|x_{t,i}|^{q-1}}/{\norm{\vec{x}_t}^{q-2}_q}, \quad\text{for all }~i \in \mathcal{N}, \\
%\end{align}
%and 
%\begin{align}
& {y}_{t+1,i} = \textstyle \left( \left(\nabla \Phi\right)^{-1}(\hat{\vec{y}}_{t+1}) \right)_i = \text{sign} (\hat{y}_{t+1,i}) {|\hat{y}_{t+1,i}|^{p-1}}/{\norm{\hat{\vec{y}}_{t+1}}^{p-2}_p}, \quad\text{for all }~i \in \mathcal{N}.\label{eq:q-norm-update2}\end{align}
%G: GAIN SPACE
%The first update rule is obtained by taking the gradient of the mirror map, while the inverse mapping is the gradient of Legendre conjugate of $\Phi$, that is $\frac{1}{2}\norm{\vec{x}}^2_p$ (see Gentile and Littlestone \cite{Gentile_pnorm}).

Finally, for all $q\in (1,2]$ the Bregman projection in Eq.~\eqref{eq:bregprojstep} (Line 5 in Algorithm \ref{algo:online_mirror_descent}) involves solving a convex optimization problem, in general. For the OGD Algorithm however ($q=2$) the projection is the usual Euclidean projection. \new{The following theorem holds:}
\begin{theorem}
\label{theorem:ogd_time_complexity}
\new{The Euclidean projection  requires $\BigO{N^2}$ operations per iteration, for general values of ${R}$ and ${h}$, and only $\BigO{N}$ operations, when $\frac{R}{h} = 1$.}
\end{theorem}
For general values of ${R}$ and ${h}$ the Euclidean projection is performed using  the projection algorithm by Wang and Lu \cite{wang2015projection} in $\BigO{N^2}$ time.  Specifically when $\frac{R}{h} = 1$, only a single coefficient is updated through the gradient step (Lines 2--4 in Algorithm~\ref{algo:online_mirror_descent}) per iteration, and Paschos et al. \cite{PaschosLearningToCache} provide an algorithm that performs the projection in $\BigO{N}$ time.\footnote{To be precise, the projection algorithm as presented in~\cite{PaschosLearningToCache} requires at each iteration a preliminary step with complexity $\BigO{N \log(N)}$ to sort a vector of size $N$, followed by $\BigO{N}$ steps. However, it is possible to replace sorting by $\BigO{\log(N)}$  binary search and insertion operations reducing the complexity to $\BigO{N}$ per iteration.}

% and can be performed in $\BigO{N^2}$ steps, using the projection algorithm by Wang and Lu \cite{wang2015projection}. Specifically when $\frac{R}{h} = 1$, only a single coefficient is updated through the gradient step (Lines 2--4 in Algorithm~\ref{algo:online_mirror_descent}) per iteration, and Paschos et al. \cite{PaschosLearningToCache} provide an algorithm that performs the projection in $\BigO{N}$ time.

%For $q$-norm mirror map OMD as $q\to1$ we provide an alternative mirror map that attains the same regret bound and overcomes the projection difficulty in \ref{sec:neg_entropy}, and in the next section we show why we emphasize on the corner cases $q=2$ and $q\to1$.

\subsection{Neg-Entropy Mirror Map}
\label{sec:neg_entropy}

To conclude this section, we turn our attention to the neg-entropy mirror map that, as discussed earlier, attains the same regret performance as $\OMD_{q\nrm}$ as $q$ converges to $1$.  Beyond its improved performance in terms of regret in the high diversity ratio regime, the neg-entropy mirror map comes with an additional computational advantage: the Bregman projection admits a highly efficient implementation.

Formally,  OMD under the neg-entropy mirror map uses: 
\begin{align}\label{eq:neg-entropy}
\textstyle \Phi(\vec{x}) &= \sum^N_{i=1} x_i \log\left(x_i\right), \text{  and } \mathcal{D} = \mathbb{R}_{>0}^N.\vspace{-10em}
\end{align}
Note that, as per the requirements  in Sec.~\ref{section:OMD},  $\mathcal{X}\subseteq\mathrm{closure}(\mathcal{D})$. Also, $\nabla\Phi$ indeed diverges at the boundary of $\mathcal{D}$, and $\nabla \Phi(\mathcal{D})=\mathbb{R}^N$, as
\begin{align}\label{eq:negrad}
\textstyle   \deriv{\Phi(\vec{x})}{x_i}= 1+\log (x_i), \quad \text{for all }i\in \mathcal{N}.
\end{align}
 We refer to the resulting algorithm as $\OMD_{\negent}$.

\subsubsection{Regret Analysis}

We first characterize the regret of $\OMD_\negent$:
\begin{theorem}
\label{theorem:regret_negentropy}
For $\eta = \sqrt{\frac{2 \log(N/k)}{\norm{\vec w}^{2}_\infty h^2 T}} $, the regret of $\OMD_{\negent}$ satisfies:
\begin{align}
\Regret_T(\OMD_{\negent}) \leq \norm{\vec w}_\infty hk \sqrt{ 2 \log (N/k)}.
\label{eq:theorem:regret_negentropy}
\end{align}
\end{theorem}
\noindent The proof, in Appendix \ref{proof:regret_negentropy}, is similar to the proof of Theorem \ref{theorem:qnorm_regret}. Using again the general bound of the regret of OMD algorithms in Bubeck \cite[Theorem 4.2]{bubeck2015convexbook}, we bound the diameter of $\mathcal{X}$ w.r.t.~to the Bregman divergence as well as the dual norm $\norm{\,\cdot\,}_\infty$ of gradients $\nabla f_{\vec{r}_t}(\vec{x}_t)$. 
Crucially, we observe that $\OMD_{\negent}$ indeed attains the same regret bound as the one in Corollary~\ref{corollary:regret_qnorm_q1}, namely, the bound on $\OMD_{q\nrm}$ when $q$ converges to $1$. This immediately implies the advantage of $\OMD_\negent$ over OGD in high diversity ratio regimes, as described in Sec.~\ref{sec:optregimes} and Theorem~\ref{theorem:optimalq_q1}.

%as $q$ converges to $1$ \cite{Gentile_pnorm}.  Though the mirror map is not defined for $q=1$, as the resulting $\Phi$ is \emph{not} differentiable, the limiting algorithm as $q$ converges to $1$ is equivalent to the so-called neg-entropy map over the simplex (capped simplex with $k=1$), and have the same multiplicative update rule with a scaling factor over a general setting of the capped simplex. We prove this result in Theorem~\ref{theorem:equivalent_1norm_negentropy}, and we elaborate in great detail in Sec.~\ref{sec:neg_entropy}.

 %OMD with a neg-entropy map amounts to a multiplicative update rule (see Eq.~\eqref{eq:multi}). Intermediate values of $q\in(1,2]$ thus establish a continuum \cite{Gentile_pnorm} between a multiplicative-update algorithm  (OMD with a neg-entropy map) and an additive-update algorithm (OGD).

% \ref{theorem:equivalent_1norm_negentropy}

\subsubsection{Dual-Primal Update  and Bregman Projection}
As $\nabla \Phi(\vec{x})$ is given by Eq.~\eqref{eq:negrad}, the inverse mapping is given by $ \left( \left( \nabla \Phi\right)^{-1}(\hat{\vec{y}}_{t+1}) \right)_i = \mathrm{exp}(\hat{y}_{t,i} - 1)$. Hence, the map to the dual space and back in Eq.~\eqref{eq:mapstep}--Eq.~\eqref{eq:invstep}  (Lines 2--4 in Algorithm~\ref{algo:online_mirror_descent}) can be concisely written as:
\begin{align}
\textstyle 	y_{t+1,i} & 
	\textstyle= \exp\left(\hat{x}_{t,i} - \eta \deriv{f_{\vec{r}_t}(\vec{x}_t)}{x_i} - 1\right)
	          = \exp\left( \log({x}_{t,i}) - \eta \deriv{ f_{\vec{r}_t}(\vec{x}_t)}{x_i} \right)
	           = x_{t,i}\,e^{-\eta \deriv{f_{\vec{r}_t}(\vec{x}_t)}{x_i}}, \, \text{for all}~i \in \mathcal{N}. 
	           %\nonumber
	           \label{eq:multi}
\end{align}
In other words, OMD under the neg-entropy mirror map adapts the cache state via \emph{a multiplicative rule} (namely, the one implied by the above equation), as opposed to the additive rule of OGD (see Eq.~\eqref{eq:ogd}). In Theorem~\ref{theorem:equivalent_1norm_negentropy} we prove that $\OMD_{q\nrm}$ when $q$ converges to $1$ also adapts the cache state via a multiplicative update rule; moreover, it is equivalent to $\OMD_{\negent}$ over the simplex. This justifies why the regret bounds for the two algorithms in Eq.~\eqref{eq:regret_qnorm_q1} and Eq.~\eqref{eq:theorem:regret_negentropy} are identical.

% [a]
\begin{algorithm}[t]
	\caption{Neg-Entropy Bregman projection onto the capped simplex}
	\begin{footnotesize}
	 \begin{multicols}{2}

	\begin{algorithmic}[1]
		\Require $N$; $k$; $\norm{\vec{y}}_1$; $P$; Partially sorted $y_N \geq \cdots \geq y_{N-k+1} \geq y_{i}, \forall i \leq N-k$
		\Statex $\triangleright${ $\vec{y}$ is the intermediate cache state, and $P$ is  a scaling factor initialized to 1}
% 		\Statex  $\triangleright$
		\State $y_{N+1} \gets +\infty$
		\For{ b $\in \{N, \dots , N-k+1\}$ }
		\State $m_{b} \gets \left( {k+b-N}\right) / \left({\norm{\vec{y}}_1 - \sum_{i=b+1}^{N} {y_i P}}\right)$
		\If {$y_{b} m_{b}P  < 1 \leq y_{b+1} m_{b} P$}
		\Statex$\triangleright${ Appropriate $b$ is found}
		\For {i $\geq$ b+1 }
		\State $y_i \gets {1} / { (m_{b} P)}$
		\EndFor
		\State $P \gets m_b P$ 
		\State \Return $\vec{y} P$\Comment {$\vec{y} P$ is the result of the projection }
		\EndIf
		\EndFor
	\end{algorithmic}
   \end{multicols}
	\end{footnotesize}
	\label{alg:bregman_divergence_projection}
\end{algorithm}
% \begin{table}[t]
% \centering
% \begin{footnotesize}
% \[
% \begin{array}{@{}l*{3}{c}@{}}
% \toprule
% \text{Diversity ratio} & \multicolumn{2}{c@{}}{\text{Policy}}\\
%     \cmidrule(l){2-3}
%     & \OGD & \OMD_\negent \\
% \midrule
% R/h=1 & \BigO{N}&  \BigO{k} \\
% R/h > 1 & \BigO{N^2} & \BigO{ N + k \log(k) }\\
% \bottomrule
% \end{array}
% \]
%  \vspace{-2em}\caption{Per-iteration time complexity of $\OGD$ and $\OMD_\negent$\vspace{-3em}} \label{table:timecomplexity}
%   \end{footnotesize}

% \end{table}

 Finally, the projection algorithm onto the capped simplex can be implemented in $\BigO{N + k \log(k)}$ time for arbitrary $R$ and $h$ values using a waterfilling-like algorithm. The full procedure is presented in Algorithm~\ref{alg:bregman_divergence_projection}.  The algorithm receives as input the top-$k$ elements of $\vec{y}$, sorted in descending order. It then identifies via a linear search 
 which elements exceed an appropriate threshold and set them to one. The other elements
 %of which the highest-value value elements are set to one, and the remaining elements 
 are scaled by a constant factor to satisfy the capacity constraint.
%  by a different constant factor; which elements are deemed high and low value is determined via a linear search.
 The following theorem holds:
\begin{theorem}
\label{theorem:bregman_divergence}
Algorithm \ref{alg:bregman_divergence_projection} returns the projection $\Pi^{\Phi}_{\mathcal{X} \cap \mathcal{D}}(\vec{y})$ onto the capped simplex $\mathcal{X}$ under the neg-entropy $\Phi$. It requires
% $\BigO{k}$ steps per iteration.  This results in an overall time complexity of 
$\BigO{N + k \log(k)}$ operations per iteration, for general values of ${R}$ and ${h}$, and only $\BigO{k}$ operations, when $\frac{R}{h} = 1$.
\end{theorem}
\noindent The proof is given in Appendix \ref{proof:bregman_divergence}. To prove this theorem, we characterize the KKT conditions of the minimization problem. Then we show that these conditions can be checked in $\BigO{k}$ time. Finally, we show how maintaining $\vec{y}$ in a partially sorted list across iterations leads to the reported complexity results.
% When $\frac{R}{h} = 1$, Alg.~1 leads to only a single state coordinate update. The projection step takes $\BigO{k}$,  and an extra $\BigO{\log(k)}$ to maintain top-$k$ components of $\vec{x}_t$ sorted online. This gives an overall time complexity of $\BigO{k}$ per iteration of OMD; we note that, in contrast, OGD requires $\BigO{N}$ per iteration in the $\frac{R}{h} = 1$ setting (see Sec.~\ref{sec:bregmanomdq}).
% When $\frac{R}{h} \neq  1$, a partial sort is to maintain top-$k$ components of $\vec{y}_t$ sorted; this can be done using partial sorting in $\BigO{N + k \log(k)}$  \cite{paredes2006optimal}. This gives an $\BigO{N + k \log(k)}$ time complexity per iteration; in contrast, OGD requires $\BigO{N^2}$ steps per iteration in this setting (see Sec.~\ref{sec:bregmanomdq}).
%
Theorem~\ref{theorem:bregman_divergence} implies that $\OMD_{\negent}$ \emph{has significant computational savings when compared to OGD} (cf.~Theorem~\ref{theorem:ogd_time_complexity}), both when $\frac{R}{h} = 1$ and for general values of $R$ and $h$. 
% The per-iteration time complexity of $\OGD$ and $\OMD_\negent$ is summarised in Table~\ref{table:timecomplexity}.

%When $\frac{R}{h} = 1$, the overall time complexity is $\BigO{k}$ per iteration of $\OMD_{\negent}$;  in contrast, OGD requires $\BigO{N}$ steps per iteration in the same setting. Similarly, for general values of $R$ and $h$, $\OMD_{\negent}$ per-iteration complexity is $\BigO{N + k \log(k)}$, while OGD requires $\BigO{N^2}$ steps per iteration. The per-iteration time complexity of $\OGD$ and $\OMD_\negent$ is summarised in Table~\ref{table:timecomplexity}.

\section{Update Cost}
\label{sec:movement}
The model presented in Sec.~\ref{s:system} can be extended by adding the  cost to update the cache state after the batch of $R$ requests has been served. This cost may quantify the additional load on the server or on the network. This update cost is often called  \emph{movement cost} \cite{bubeck2015convexbook} or \emph{switching cost} \cite{taleoftwometrics}.
As the state changes from $\vec x_t$ to $\vec x_{t+1}$, the cache evicts part of the file $i$ if $x_{t+1,i} < x_{t,i}$ and stores additional bytes of it if $x_{t+1,i} > x_{t,i}$.
We make the following assumptions:
\begin{enumerate}
    \item Evictions do not engender update costs, as the cache can perform them autonomously;
    \item Insertions of (part of) files which have been requested do not engender update costs, as these files have already been retrieved by the cache in their entirety to satisfy the requests.
    \item Insertions of (part of) files which have not been requested incur a cost proportional to the fraction of file retrieved. 
\end{enumerate}

We can then define the update cost at time slot $t$ as
\begin{align}
\label{eq:movement_fractional}
\textstyle \updatecost_{\vec{r}_t} (\vec{x}_{t}, \vec{x}_{t+1} ) = \sum_{i \notin \supp(\vec{r}_t)} w_i' \max\left\{0, x_{t+1,i} - x_{t,i}\right\},
\end{align}
where $\supp(\vec{r}_t) = \left\{i \in \mathcal{N}: r_{t,i}  \neq 0\right\}$ denotes the support of $\vec{r}_t$, i.e., the set of files that have been requested during the $t$-th timeslot, and $w'_i \in \reals_+$ is the cost to retrieve the whole file $i$, and can in general be different from the cost $w_i$ appearing in~\eqref{e:cost}.

If the update cost is introduced in the model, the \emph{extended} regret can be defined as follows:
% Call it extended regret in text.
\begin{align}
   \textstyle &\text{E-Regret}_T({\mathcal{A}}) = \textstyle\underset{\{\vec{r}_1, \vec{r}_2, \dots, \vec{r}_t\} \in \mathcal{R}^T_{R,h}}{\sup} \left\{ \sum^T_{t=1} f_{\vec{r}_t}(\vec{x}_t) + \updatecost_{\vec{r}_t} (\vec x_{t}, \vec x_{t+1})-\sum^T_{t=1} f_{\vec{r}_t}(\vec{x_*}) \right\} \label{eq:regret_extended}\\
    &\le \textstyle \underset{\{\vec{r}_1, \vec{r}_2, \dots, \vec{r}_t\} \in \mathcal{R}^T_{R,h}}{\sup} \left\{  \sum^T_{t=1} f_{\vec{r}_t}(\vec{x}_t) - \sum^T_{t=1} f_{\vec{r}_t}(\vec{x}_*) \right\} +  \underset{\{\vec{r}_1, \vec{r}_2, \dots, \vec{r}_t\} \in \mathcal{R}^T_{R,h}}{\sup}\left\{ \sum^T_{t=1}  \updatecost_{\vec{r}_t} (\vec x_{t}, \vec x_{t+1})\right\}. \label{eq:rounding_regret_upper_bound}
\end{align}
Equation~\eqref{eq:rounding_regret_upper_bound} shows that the regret of an arbitrary online algorithm can be bounded by considering the regret we have derived so far (Eq.~\eqref{eq:regret}), ignoring update costs, and subsequently accounting for an additional term corresponding to the update. Note that the optimal static allocation does not incur any update cost.  Equation~\eqref{eq:rounding_regret_upper_bound} implies that any policy with $\BigO{\sqrt{T}}$ regret and  $\BigO{\sqrt{T}}$ update cost in expectation has also $\BigO{\sqrt{T}}$ extended  regret.

One of the reasons why we did not introduce directly the update cost is that, in the fractional setting, OMD update cost is zero both for the Euclidean (OGD) and the neg-entropy ($\OMD_{\negent}$) mirror maps. Formally, we have:
\begin{proposition}
\label{prop:free_updates}
For any request batch $\vec r_t$ received at time slot $t \in [T]$, the update of fractional cache state from $\x_t \in \mathcal{X}$ to $\x_{t+1} \in \mathcal{X}$ obtained by $\OMD_\negent$ or $\OGD$ has no cost, i.e., $\updatecost_{\vec r_t} (\vec x_t, \vec x_{t+1}) = 0$.
\end{proposition}
The proof is provided in Appendix~\ref{proof:free_updates}. In fact, the gradient step increases the fraction  $x_{t,i}$ only for files $i$  that have been requested, and the projection step reduces the fraction for all other files in order to satisfy the capacity constraint. It follows that $x_{t+1,i} - x_{t,i} > 0$ if and only if $i \in \supp(\vec r_t)$, and thus $\updatecost_{\vec{r}_t} (\vec{x}_{t}, \vec{x}_{t+1} )=0$. Hence, the $\BigO{\sqrt{T}}$ regret guarantees we proved in the previous sections for OGD and $\OMD_{\negent}$ extend to the more general definition in~\eqref{eq:regret_extended}.
In the next section, we show that update costs \emph{cannot be neglected} when caches are forced to store files in their entirety.

\section{Integral Caching}
\label{sec:rounding}
In the previous sections, we assumed that the cache can store arbitrarily small chunks of a file, and this allowed us to design no-regret policies that employ fractional caching. However, this assumption can be too strong in some applications. For example, when the catalog is composed of small-sized files, the discreteness of chunks sizes cannot be neglected;
% we are limited to only extract relatively big chunks and rounding such allocations can cause a significant performance drop;
moreover, the metadata needed for each chunk can cause memory and computational overheads.  These observations motivate us to study the case when the cache can only store the entire file. We refer to this setting as the \emph{integral} caching. Formally, we restrict the cache states to belong to the set  $\textstyle \mathcal{Z} = \left\{\vec \zeta \in \{0,1\}^N: \sum_{i \in \mathcal{N}} \zeta_i = k\right\}$. Note that the set $\mathcal{Z}$ is a restriction of the set of fractional caching states $\mathcal{X}$ to its corners, i.e., $ \mathcal{Z} = \mathcal{X} \cap \{0,1\}^N$; thus, we maintain the same definition of the requests and the service cost objective in Sec.~\ref{s:system}.  In this setting, we allow policies to be randomized. This extension turns out to be necessary in order to have a sublinear regret policy; formally, we have:
\begin{proposition}
\label{proposition:deterministic_worst_case_regret}
Any deterministic policy restricted to select integral cache states in $\mathcal{Z}$ has the following lower bound on its regret: $\Regret_T(\policy) \geq  k \left(1 - {k}/{N}\right)T$.
\end{proposition}
To prove the proposition, we show that an adversary can exploit the deterministic nature of the policy by continuously requesting the files that are not stored in the cache. We provide the proof in Appendix~\ref{proof:deterministic_worst_case_regret}. 

We thus turn our attention to randomized policies. In particular, we focus on a special class of randomized policies, constructed by (1) a fractional online caching policy $\policy$, i.e., of the type we have studied so far (see Sec.~\ref{s:system}),   combined with (2)  a randomized rounding scheme $\Xi$, that maps fractional caching states to integral ones.  In particular, for every $t \geq 1$ the randomized rounding scheme $\Xi: \mathcal{X}^t \times \mathcal{Z}^{t-1} \times [0,1] \to \mathcal{Z}$ maps the previous fractional cache states $\{\x_{s}\}^{t-1}_{s=1} \in \mathcal{X}^{t-1}$, the current fractional cache state $\x_t \in \mathcal{X}$, the previous random cache states $\{\z_{s}\}^{t-1}_{s=1} \in \mathcal{Z}^{t-1}$, and a source of randomness\footnote{\new{In this section, we assume the adversary is oblivious~\cite[Sec.~5.5]{Hazanoco2016}, i.e., he selects the request process
adversarially ahead of time, independently of the decisions of the online learner.}} $\xi_t \in [0,1]$ to a new random cache state $\z_t \in \mathcal{Z}$ where 
\begin{align}\label{eq:expectzt}\mathbb{E} [\z_t] = \x_t.\end{align}
Note that the rounding function takes into account not only the current fractional state $\x_t$, which determines its expectation, but also the past fractional and integral states ($\{(\z_{s},\x_{s})\}^{t-1}_{s=1}$); this is in fact instrumental in attaining a sublinear extended regret (see Theorems~\ref{theorem:rounding_independent} and~\ref{theorem:rounding_coupled} below).

We extend the definitions of the regret and the extended regret as follows:
\begin{align}
\textstyle\mathrm
{Regret}_T(\policy, \Xi) = \textstyle \underset{\{\vec{r}_1, \vec{r}_2, \dots, \vec{r}_t\} \in \mathcal{R}^T_{R,h}}{\sup} \left\{ \mathbb{E} \left[ \sum^T_{t=1} f_{\vec{r}_t}(\vec{z}_t)\right]- \sum^T_{t=1} f_{\vec{r}_t}(\vec{z}_*) \right\}, 
\label{eq:integral_regret}
\end{align}
and 
\begin{align}
  \textstyle  \mathrm{E\mbox{-}Regret}_T(\policy, \Xi) &=\textstyle \underset{\{\vec{r}_1, \vec{r}_2, \dots, \vec{r}_t\} \in \mathcal{R}^T_{R,h}}{\sup} \left\{ \mathbb{E} \left[\sum^T_{t=1} f_{\vec{r}_t}(\vec{z}_t) + \updatecost_{\vec{r}_t} (\vec z_{t}, \vec z_{t+1})\right]-\sum^T_{t=1} f_{\vec{r}_t}(\vec{z}_*) \right\},
    \label{eq:integral_extendedregret}
\end{align}
where the expectation is taken over the random choices of the rounding scheme $\Xi$, and  \begin{align}\textstyle \vec{z}_*=\argmin_{\vec{z} \in \mathcal{Z}}\sum^T_{t=1} f_{\vec{r}_t}(\vec{z})\end{align}
 is the optimal static integral cache state (in hindsight).  By restricting our focus to such randomized policies, we obtain a regret that is equal to the fractional caching policy's regret. Formally, we have:
  \begin{proposition}
\label{proposition:regret_transfer}
% Any random integral caching states $\z_t \in \mathcal{Z}$, for $t\in[T]$, generated from fractional caching states of policy $\policy$ via a randomized rounding $\Xi$ that satisfies \eqref{eq:expectzt}, 
Any randomized caching policy constructed by an online policy $\policy$ combined with  a randomized rounding scheme $\Xi$ has the same regret as $\policy$, i.e., $\Regret_T(\policy, \Xi) =\Regret_T(\policy)$, given by \eqref{eq:integral_regret} and \eqref{eq:regret}, respectively.
\end{proposition}
 The result follows from the linearity of the cost functions and the expectation operator; moreover, the static optimum can always be selected to be integral from the integrality of the capacity constraint and linearity of the objective function. The proof is provided in Appendix.~\ref{proof:regret_transfer}.

Proposition~\ref{proposition:regret_transfer} thus implies that regret guarantees for a fractional policy $\mathcal{A}$ readily transfer to the integral regime, when coupled with rounding $\Xi$. Unfortunately, when considering the extended regret  (Eq.~\eqref{eq:integral_extendedregret}) instead, na\"ive  rounding policies can \new{arbitrarily} evict and fetch objects to the cache causing large update costs (see Theorem~\ref{theorem:rounding_independent}). Thus, unless rounding is carefully designed, we may fail to have sublinear regret guarantees when accounting for update costs. In the next section, we show how a randomized rounding scheme~$\Xi$ can be  selected to avoid incurring large update costs.

\subsection{Rounding Schemes and Extended Regret} 
\label{s:rounding_schemes}
% \begin{align}\mathcal{Z} = {\left\{ \vec{\zeta} \in \{0,1\}^N: \sum^N_{i=1} \zeta_i = k\right\}}.\end{align}
%be the set of possible integral cache configurations.
% We are going to show that, by coupling OMD methods with opportune rounding techniques, the service-cost regret guarantees derived for the fractional setting transfer (in expectation) to the integral one. %
% A straightforward approach is to use OMD to update a fractional state $\vec{x}_t\in \mathcal{X}$ and then randomly sample an integral state $\vec z_t \in \mathcal{Z} $ after each round of requests so that $\mathbb{E}[\vec z_t] = \vec{x}_t$, for all $t\geq 1$. 
%\begin{align}\mathbb{E}[\vec z_t] = \vec{x}_t,\quad \text{for all}~t\geq 1.\end{align}

% S: describe formally randomized algos.

% In this section, we 

\subsubsection{Online Independent Rounding.}  If we consider a fractional caching state $\x_t \in \mathcal{X}$, then a random integral caching state $\vec z_t \in \mathcal{Z}$ with the marginal $\mathbb{E}[\z_t] = \x_t$ exists and can be sampled in polynomial time (see, e.g., \cite{Madow1944Mar,geocaching2015, ioannidis2016adaptive}).  Thus, a rounding scheme $\Xi$ can be constructed with such \new{a} strategy that takes as input the current fractional cache state $\x_t$ 
ignoring \new{the} previous fractional cache states $\{\x_{s}\}^{t-1}_{s=1} \in \mathcal{X}^{t-1}$, and previous random cache states $\{\z_{s}\}^{t-1}_{s=1} \in \mathcal{Z}^{t-1}$.
%and ignores the previous cache state $\x_{t-1}$ and random cache state $\z_{t-1}$. 
We provide pseudocode for this procedure in Algorithm~\ref{algo:online_rounding}.\footnote{Algorithm~\ref{algo:online_rounding} provides a linear-time variant of the algorithms proposed in~\cite{geocaching2015, ioannidis2016adaptive}. The algorithm samples an integral caching state without constructing a distribution and its support. \new{This sampling scheme is also known as Madow's sampling~\cite{Madow1944Mar}.}} Because at any time $t$ the random cache states are sampled independently from previous random cache states, we refer to this rounding as \emph{online independent rounding}. Unfortunately, when considering the extended regret~\eqref{eq:integral_extendedregret}, any caching policy coupled with this rounding scheme loses its $\BigO{\sqrt{T}}$ regret guarantee. Formally, we have the following:

\begin{theorem}
\label{theorem:rounding_independent}
% Consider a policy that adds $\eta$ fraction to the requested file and 
Any randomized caching policy constructed by an online policy $\policy$ combined with  online independent rounding as a randomized rounding scheme $\Xi$ has linear (worst-case) extended regret, i.e., $\mathrm{E\mbox{-}Regret}_T(\policy, \Xi)  = \Omega (T)$.

% Online independent rounding incurs $\Omega(T)$ update cost in the worst case.
\end{theorem}
The proof is provided in Appendix~\ref{proof:rounding_independent}. Online independent rounding causes frequent cache updates, as it samples a new state from $\z_t$ ignoring the previous state $\vec\zeta_{t-1}$ sampled from $\z_{t-1}$. Intuitively, imposing dependence (coupling) between the two consecutive random states may significantly reduce the expected update cost.

\begin{algorithm}[!t]
\begin{multicols}{2}

\begin{footnotesize}
	\begin{algorithmic}[1]
% 	\State Sample $\xi$ u.a.r in the interval $[0,1]$ 
    % \State Assume that files' representations in $\mathcal{N}$ are assigned in increasing movement cost, such that $w'_i \geq w'_{i-1}$ for all $i \in \mathcal{N} \setminus \{1\}$. 
	\Procedure {\Rounding}{$\x \in \mathcal{X}$, $\xi \in [0,1]$}
    \State $\mathcal I_0 = \emptyset$
    \For {$i = 1,2,\dots, N$} 
    \State $\mathcal I_{i} \gets \begin{cases} \mathcal I_{i-1}\cup \{ i\} &\text{if}\quad \sum^i_{j=1} x_{j} \geq \xi + |\mathcal I_{i-1}|,\\ \mathcal I_{i-1} &\text{otherwise.} \end{cases}$
    %\State $I_{i} \gets  I_{i-1} \cup C_i$
    \EndFor
    \State \Return $\vec \z \gets \sum_{i \in I_N} \vec e_i $
	\EndProcedure
\Statex $\triangleright$ In  \emph{online independent rounding},  \Rounding  is called with arguments ($\x_t$, $\xi_t$), where $\x_t$ are provided by algorithm $\mathcal{A}$ and $\{\xi_t\}_{t=1}^{T-1}$ are i.i.d., sampled u.a.r. from $[0,1]$.
\Statex $\triangleright$ In \emph{online coupled rounding}, a $\xi$ is sampled once u.a.r. from $[0,1]$; then,  \Rounding  is called with arguments ($\x_t$, $\xi$), i.e., using the same $\xi$ for all $\x_t$  provided by algorithm $\mathcal{A}$. 
\Statex $\triangleright$ Both return an integral r.v.~$\z_t$ s.t. $\mathbb{E}[\z_t]=\vec x_t$, with the expectation being over $\{\xi_t\}_{t=1}^{T-1}$ and $\xi$, respectively.
	\end{algorithmic}

	\end{footnotesize}
	\end{multicols}
		\caption{\Rounding }
			\label{algo:online_rounding}
\end{algorithm}

\subsubsection{Online Coupled Rounding.}  To address this issue, our proposed \emph{online coupled rounding} scheme is  described also in Algorithm~\ref{algo:online_rounding}, using however the same randomization source across all timeslots. In particular, the coupling across states comes from the use of the same uniform random variable~$\xi$.
% the dependency on the previous random states (coupling) is possible when fixing the value of the random variable  sampled only once. 
%In that sense, %Algorithm~\ref{algo:online_rounding} is an adaptation of the random placement schemes proposed by~\cite{geocaching2015, ioannidis2016adaptive} with the main difference is for any time slot $t \in {T}$ the random variables $\vec z_t$ are implicitly constructed and coupled with $\vec z_{t-1}$. 
A consequence of this coupling is that 
%the computation of the next integral state is computationally efficient and requires only small movement costs. 
the next integral state can be computed efficiently and leads to small movement costs.
Note that Algorithm~\ref{algo:online_rounding} does not necessarily find an optimal coupling, %but it is sufficiently conservative: overall, 
still it yields a sublinear update cost, and thus preserves the sublinearity of the regret. This is formally expressed in the following Theorem: 
\begin{theorem}
\label{theorem:algorithm_rounding_coupled}
Consider a randomized caching policy constructed by an OMD policy $\mathcal{A}$  with sublinear regret (i.e., configured with \new{a} learning rate $\eta = \Theta{\left( {1}/{\sqrt{T}}\right)}$) combined with online coupled rounding $\Xi$ in Algorithm~\ref{algo:online_rounding} (fixed $\xi_t = \xi$ for $t \in [T]$). The expected movement cost of the random integral cache states is $\mathbb E \left[\updatecost_{\vec r_t}\left(\vec z_{t}, \vec z_{t+1} \right)\right] = \BigO{\sqrt{T}}$. Moreover, the extended regret is sublinear $\mathrm{E\mbox{-}Regret}_T(\policy, \Xi)  = \BigO {\sqrt{T}}$.
\end{theorem}
We provide the proof in Appendix~\ref{proof:algorithm_rounding_coupled}. In summary, any OMD policy combined with online coupled rounding yields $\BigO {\sqrt{T}}$ extended regret in the integral caching setting. The computational complexity of online coupled rounding is \BigO{N}  (see also  Fig.~\ref{fig:elementary_movement}).

\subsubsection{Online Optimally-Coupled Rounding.}
It is possible in general to reduce the update cost of online coupled rounding.
In particular, minimizing the expected update cost over all joint distributions of the random variables $\vec z_t$ and $\vec z_{t+1}$ leads to an optimal transport problem~\cite{peyre2019computational}.
%If we optimize over all the joint distributions of the random variables $\vec z_t$ and $\vec z_{t+1}$ seeking to minimize the expected update cost, then such objective is attained by solving an optimal transport problem~\cite{peyre2019computational}. 
For completeness, we describe this rounding scheme here, though (1) it does not reduce the extended regret guarantee attained by online coupled rounding (up to multiplicative constants), and (2) it has an increased computational cost.

Formally, at each time $t$ the random variables $\vec z_t$ with marginal $\vec x_t$ can be constructed by sampling from a distribution $\vec p_t$ with \BigO{N} support $\left \{\vec{\zeta}^1_t, \vec{\zeta}^2_t, \dots, \vec{\zeta}^{|\vec p_t|}_{t}\right\}$, where $p_{t,i} = \mathbb{P} (\vec z_t = \vec\zeta^i_t)$ for $i \in [\card{\vec p_t}]$. The decomposition can be performed in \BigO{k N \log(N)} steps~\cite{ioannidis2016adaptive}. We denote the joint probability $\mathbb{P} \left (\vec z_{t+1} = \vec\zeta^j_{t+1} , \vec z_t = \vec\zeta^i_{t}\right)$ by the flow $f_{i,j}$ for all $(i,j) \in [|\vec p_t|] \times [|\vec p_{t+1}|]$. The optimal transport problem can be described by the following linear program:
\begin{align*}
\vec f = &\argmin_{{[f_{i,j}]_{(i,j)\in  [|\vec p_t|] \times [|\vec p_{t+1}|]}}}\textstyle\mathbb{E}\left[ \updatecost\left(\z_t,\z_{t+1}\right)\right] = \sum^{|\vec p_t|}_{i=1} \sum^{|\vec p_{t+1}|}_{j=1}  \updatecost_{\vec r_t} \left(\vec \zeta^i_{t},\vec \zeta^j_{t+1}\right) f_{i,j}\\
&\text{s.t.} \quad\textstyle
\sum^{|\vec p_{t+1}|}_{j=1}f_{i,j} = p_{t,i}, \quad\sum^{|\vec p_{t}|}_{i = 1} f_{i,j} = p_{t+1, j},\quad f_{i,j} \in [0,1],  \forall (i,j)\in  [|\vec p_t|] \times [|\vec p_{t+1}|].
\end{align*}
% where $\mathcal{C} (\vec F)$ is the expected update cost, given as
% \begin{align}
%     \mathcal{C} (\vec F) \triangleq \mathbb{E}\left[ \updatecost\left(\z_t,\z_{t+1}\right)\right] = \sum^{|\vec p_t|}_{i=1} \sum^{|\vec p_{t+1}|}_{j=1}  \updatecost_{\vec r_t} (\vec \zeta^i_{t},\vec \zeta^j_{t+1} ) f_{i,j}.
% \end{align}

We  solve the above linear program to obtain a minimum-cost flow $\vec f$. If the random state at time $t$ is $\vec \zeta^i_t$, then we select the new random state to be $\vec  \zeta^j_{t+1}$ with (conditional) probability $ \mathbb{P} \left( \vec z_{t+1} =\vec \zeta^j_{t+1} \given \vec z_t =\vec \zeta^i_t \right) = \frac{f^*_{i,j}}{p^t_i}$.
Such coupling ensures that the expected update cost is minimized. When we combine this rounding scheme with a no-regret fractional policy we obtain sublinear extended regret~\eqref{eq:integral_extendedregret}:
\begin{corollary}
\label{corollary:opt_coupling_sublinear_updatecost}
Consider an OMD policy $\mathcal{A}$ configured with learning rate $\eta = \Theta{\left( \frac{1}{\sqrt{T}}\right)}$ combined with online optimally-coupled rounding $\Xi$. The obtained randomized integral caching policy has sublinear extended regret, i.e., $\mathrm{E\mbox{-}Regret}(\policy, \Xi) = \BigO{\sqrt{T}}$.
\end{corollary}
The corollary follows from Theorem~\ref{theorem:algorithm_rounding_coupled}, because online coupled rounding constructs a feasible transportation flow (see Fig.~\ref{fig:conditioned_implicit} for an illustration) that  gives sublinear update costs, and the optimal flow can only have lower update costs.
% The proof is provided in Appendix~\ref{proof:opt_coupling_sublinear_updatecost}.
%
The na\"ive implementation of the optimal transport problem has $\BigO{N^3}$ time complexity, but several efficient approximations exist in the literature~\cite{peyre2019computational} at the expense of \new{losing} the established guarantee.

\section{Numerical Experiments}
\label{sec:experiments}

%We evaluate the performance of the different caching policies under synthetic traces and a real trace.  

\subsection{Experimental setup}

\subsubsection{Datasets.}
Throughout all experiments, we assume equal costs per file, i.e., $w_i = w'_i = 1, \forall i \in \mathcal{N}$. The learning rate $\eta^*$ denotes the learning rate value specified in Corollary~\ref{corollary:regret_qnorm_q2} and in Theorem~\ref{theorem:regret_negentropy} for $\OGD$ and $\OMD_\negent$, respectively. \new{Note that all the parameters assumed known to the algorithm can be learned through the following meta-algorithm:
%forming an expert problem to select the best policy, i.e., 
one can execute in parallel multiple $\OMD_{\negent}$ and $\OGD$ algorithms configured for different $R/h$ values in $\{\Delta,2\Delta,  \dots, N\}$ and frame an expert problem to learn the best expert (policy)}. \nnew{The meta problem is a standard prediction with expert advice problem, and can be tackled with well understood and computationally efficient learning algorithms~\cite{Hazanoco2016,ShalevOnlineLearning, mcmahan2017survey}. } In what follows, we distinguish the number of batches  in the trace~($B$) and the time horizon ($T$). 

We generate the following synthetic datasets, summarized in Table~\ref{tab:traces}.

\noindent\textbf{Fixed Popularity.}
Requests are i.i.d.~and sampled from a catalog of $N =10^5$ files according to a \emph{Zipf} distribution with exponent $\alpha = 0.8$. 
%This distribution represents the popularity of the files.
Each batch counts a single request ($R=1$). We set set the time horizon as $T = 10^5$.  The cache capacity is $k = 100$.
The total number of requests is  the product of the requests in each batch ($R$) and the number of batches ($B$), both values are reported in Table~\ref{tab:traces}.

\noindent\textbf{Batched Fixed Popularity}.
Request are generated as above from a Zipf distribution with exponent~$\alpha$, but are now grouped in batches of $R=5 \times 10^3$ requests.
We take different exponents $\alpha \in \{0.1, 0.2, 0.7\}$ for  traces \emph{Batched Fixed Popularity} (1), (2), and (3), respectively, in Table \ref{tab:traces}.
The parameter $\alpha$ controls the diversity of the files in the request batches. If $\alpha = 0$, then each file is requested with equal probability, %of each file to be requested,
corresponding to $\frac{R}{h} \to N$ (high diversity). As we increase $\alpha$, the requests become more concentrated; this corresponds to $\frac{R}{h} \to 1$ (low diversity). Table~\ref{tab:traces} shows the value of $h$ observed in each trace. In all cases, we select catalog size $N = 10^4$, cache size $k \in \{25,125,250\}$,  and time horizon $T = 10^4$. 

\noindent\textbf{Transient Popularity}. We also generate two non-stationary request traces. In these traces, we reset the popularity distribution periodically.

In the first scenario (\emph{Partial Popularity Change} traces), we still have batches of $R=5 \times 10^3$ requests sampled from a catalog of $N=10^4$ files according to a Zipf distribution with parameter $\alpha \in \{0.1, 0.3,0.4\}$ for traces (1), (2), and (3), respectively. But now the popularities of a subset of files is modified every $10^3$ time slots. In particular the 5\% most popular files become the 5\% least popular ones and vice versa. 
We want to model a situation where the cache knows the timescale over which the request process changes and which files are affected (but not how their popularity changes). Correspondingly, the time horizon is also set to $T=10^3$ and, at the end of each time horizon, the cache redistributes uniformly the cache space currently allocated by those files. The cache size is $k=50$.

%the file provider periodically refreshes its library, and both catalog changes and the periods of stationarity are known (even though the popularity values are not).
%In every period of $10^3$ timeslots, the  we change the popularity of 10\% of the catalog  as follows: the popularities of most 5\% popular files are randomly assigned to the 
%least 5\% popular files, and vice versa.
 
% From the algorithmic perspective,
%we set the targeted time horizon to be equal to the (known) period of stationarity; i.e., $T=10^3$. 
%That is, when implementing OGD and OMD, we set the time horizon $T$ (used to determine $\eta$) to be equal to this period; at the end of each period $T$, we reset the caching decision of each algorithm, by spreading the mass uniformly among the files affected by this popularity change. %We set the period of stationarity to $T=10^3$. 
% This happens 5 times, resulting in a total of 5T batches in the entire trace. We  again generate different diversity regimes, with exponents $\alpha \in \{0.1, 0.4,0.7\}$, batch size $R=5 \times 10^3$, cache size $k =5$, catalog size $N = 500$. This gives traces \emph{Partial Popularity Change} (1), (2) and (3), corresponding to the different exponents, respectively.
 
In the second scenario (\emph{Global Popularity Change} trace) each batch counts only a single request ($R=1$) sampled from a catalog of $N=10^4$ files according to a {Zipf} distribution with exponent $\alpha=0.8$. Every $5\times 10^4$ time slots (or requests in this case) the popularity of each files change: file $i \in \{1, \dots, N\}$ assumes the popularity of file $j=(1+ (i + N/4) \! \mod N$). The cache size is $k=200$.  We also generate the \emph{Downscaled Global Popularity Change} trace as  a downscaled version of \emph{Global Popularity Change} trace, where the catalog size is reduced to $N = 25$, the cache size to $k=4$, and the number of requests to $ 9 \times 10^3$. The learning rate is set to $\eta = 0.01$.

%each file we perform a circular shift by a step $50$ to the components of the popularity distribution vector. Hence,  the popularity profile of the requests changes significantly from one timeslot to the next. This gives the \emph{Global Popularity Change} trace, with a total duration (total number of batches) of $15 \times 10^4$.

\noindent\textbf{Akamai Trace.}
We consider also a real file request trace collected from Akamai Content Delivery Network (CDN)~\cite{akamaiGiovanni}. The trace spans 1 week, and we extract from it about $8.5 \times 10^7$ requests for the $N=10^4$ most popular files. We group requests in batches of size $R = 5 \times 10^3$, and we consider a time horizon $T = 100$ time slots corresponding roughly to 1 hour. %($5 \times 10^5$ individual file requests). 
The cache size is $k=25$.
% 

% \begin{minipage}{.3\linewidth}
\begin{table}[t]
\centering
\begin{footnotesize}

  \begin{tabular}{|l|l|l|l|l|l|}
    \hline
    Trace  & $B$ &$T$ &$N$& $R$ & $h$ \\
    \hline
    Fixed Popularity & $1.5  \times 10^5$ & $1.5  \times 10^5$& $10^4$ &1 & 1\\
    Batched Fixed Popularity  (1) & $10^4$ & $10^4$&$  10^4$& $5\times10^3$  & 2\\
    Batched Fixed Popularity  (2) & $10^4$ & $10^4$&$ 10^4$& $5\times10^3$  & 5\\
    Batched Fixed Popularity (3) & $10^4$ & $10^4$&$ 10^4$& $5\times10^3$  & 87\\
    Partial Popularity Change  (1)  & $5 \times 10^3$ & $10^3$&$10^4$& $5\times10^3$ & 2\\
    Partial Popularity Change  (2)  &  $5 \times 10^3$ & $10^3$&$10^4$& $5\times10^3$ & 6\\
    Partial Popularity Change  (3) &  $5 \times 10^3$  & $10^3$&$10^4$& $5\times10^3$ & 10\\
    Global Popularity Change & $1.5 \times 10^5$ & $1.5 \times 10^5$& $10^4$&1 & 1\\
    Downscaled Global Popularity Change & $9 \times 10^3$ & $9 \times 10^3$& $25$&1 & 1\\
    Akamai CDN & $1.7\times 10^4$ & $10^2$ & $10^3$ & $5 \times 10^4$ & $380$ \\
    \hline
\end{tabular}
\end{footnotesize}
\vspace{0em}\caption{Trace Summary\vspace{-2em}}  \label{tab:traces}
\end{table}
% \end{minipage}
% \begin{minipage}{.4\textwidth}
\begin{table}[t]
\centering

  \begin{footnotesize}
  \begin{tabular}{|l|l|l|}
    \hline
    Performance metric & Definition & Range \\
    \hline
    Normalized Average Cost  & $ \mathrm{NAC} (\mathcal{A}) =  \frac{1}{R t} \sum^t_{s = 0 }f_{\vec{r}_s}(\vec{x}_s)$ & $[0,1]$\\
    Normalized Moving Average Cost  &  $\mathrm{NMAC} (\mathcal{A}) =  \frac{1}{R \min(\tau,t)} \sum^t_{s = t-\min(\tau,t) }f_{\vec{r}_s}(\vec{x}_s)$& $[0,1]$ \\
    Time Average Regret& $\mathrm{TAR} (\mathcal{A}) = \frac{1}{t}\left(\sum^t_{s=1} f_{\vec{r}_s}(\vec{x}_s)- \sum^t_{s=1} f_{\vec{r}_s}(\vec{x_*}) \right)  $ & $[0,R]$ \\
    Cumulative Update Cost & $\mathrm{CUC} (\mathcal{A}) = \sum^t_{s=1} \updatecost_{\vec{r}_s} (\vec{x}_{s}, \vec{x}_{s+1} )$ & $ [0, \infty)$ \\
    \hline
\end{tabular}
\end{footnotesize}
\vspace{-1em}\caption{Performance Metrics. All are better if lower. \vspace{-3em}}
  \label{tab:performance_metrics}
\end{table}
% \end{minipage}

% \vspace{-2em}
\subsubsection{Online Algorithms.} Starting with the gradient based algorithms, we implemented $\OMD_\negent$ with the projection defined in Algorithm \ref{alg:bregman_divergence_projection}. We implemented two different projection algorithms for $\OGD$: the one by Paschos et al. \cite{PaschosLearningToCache} for the setting $\frac{R}{h} = 1$, and the one by Wang and Lu  \cite{wang2015projection} for the general setting $\frac{R}{h} > 1$. 

In addition, we implemented four caching eviction policies: LRU, LFU, W-LFU, and FTPL. LRU and LFU evict the least recently used and least frequently used file, respectively. While LFU estimates file popularities considering all requests seen in the past, W-LFU~\cite{karakostas2002exploitation} is an LFU variant that only considers requests during a recent time window $W$, which we set equal to $T \times R$ in our experiments. The policies LRU, LFU, and W-LFU are allowed to process individual requests.  FTPL is a no-regret policy proposed by Mukhopadhyay and Sinha~\cite{mukhopadhyay2021online}, which, roughly speaking, behaves as a LFU policy whose request counters are perturbed by some Gaussian noise. 
Finally, we define \emph{Best Static} to be the optimal static allocation $\vec{x}^*
$, i.e., the configuration storing the $k$ most popular files as we consider $w_i = 1, \forall i \in \mathcal{N}$. We also define \emph{Best Dynamic} to be the caching policy that stores the $k$ most popular files at any time for the synthetic traces (for which the instantaneous popularity is well defined). The optimality of such policy is formally studied in~\cite{panigrahy21}.

\subsubsection{Online Rounding.}
We also implemented the three rounding schemes described in~Sec.~\ref{sec:rounding}: (a) the online independent rounding in Algorithm~\ref{algo:online_rounding}, (b) the online coupled rounding in Algorithm~\ref{algo:online_rounding}, and (c) the online optimally-coupled rounding.
%in Sec~\ref{sec:rounding}.
The rounding schemes are combined with  $\OGD$ configured with learning rate $\eta = 0.01$ under the \emph{Downscaled Global Popularity Change} trace.

\subsubsection{Performance Metrics.} We measure performance  w.r.t.~four metrics  defined in Table~\ref{tab:performance_metrics}. The Normalized Average Cost $\mathrm{NAC}(\mathcal{A}) \in [0,1]$ corresponds to the time-average cost over the first $t$ time slots, normalized by the batch size $R$.  The Normalized Moving Average Cost  $\mathrm{NMAC}(\mathcal{A}) \in [0,1]$ is computed similarly, using a moving average instead over a time window $\tau > 0$; we use $\tau = 500$ in our experiments.  We also consider the Time Average Regret $\mathrm{TAR} (\mathcal{A}) \in [0, R]$, which is precisely the time average regret over the first $t$ time slots. Finally, when studying rounding algorithms, we also measure and report the Cumulative Update Cost $\mathrm{CUC}(\mathcal{A}) \in [0, \infty)$.

\subsection{Results}
\subsubsection{Stationary Requests}
\begin{SCfigure}[50][t]
\centering
  \begin{minipage}{.7\textwidth}
    \subcaptionbox{$\mathrm{NAC}$ of OGD}{
    \includegraphics[width=.32\textwidth]{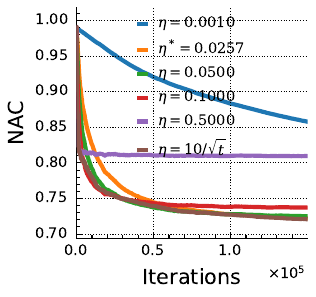}\!\!
  }%
  \subcaptionbox{$\mathrm{NAC}$ of $\OMD_\negent$}{
     \centering
    \includegraphics[width=.32\textwidth]{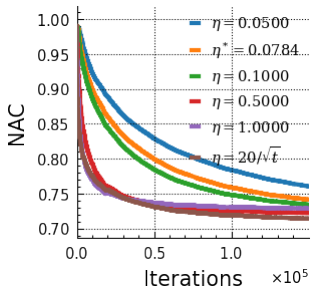}\!\!
  }%
  \subcaptionbox{Time-Average Regret}{
    \includegraphics[width=.32\textwidth]{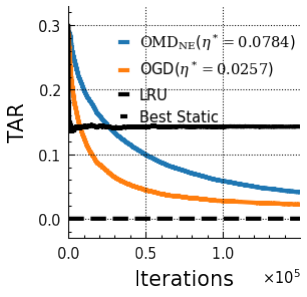}
  }
  \end{minipage}
  \hspace{-1.5em}
  \caption{NAC of the different caching policies over the \emph{Fixed Popularity} trace. Subfigures (a) and (b) show the performance of OGD and $\OMD_\negent$ respectively under different learning rates. For small learning rates the algorithms both converge slower but more precisely, while for larger learning rates they converge faster, but to a higher cost. Subfigure (c) shows the time average regret of the two gradient algorithms. When the regret is sub-linear, the time average regret converges to $0$ as $T \to \infty$.  }
  \label{fig:performance_stationary}
\end{SCfigure}

Figures \ref{fig:performance_stationary} (a) and \ref{fig:performance_stationary} (b) show the performance w.r.t.~$\mathrm{NAC}$ of OGD and $\OMD_\negent$,  respectively, under different learning rates $\eta$ on the \emph{Fixed Popularity} trace. We observe that  both algorithms  converge slower under small learning rates, but reach a final lower cost, while  larger learning rates lead to faster convergence, albeit to higher final cost. This may motivate the adoption of 
%motivates choosing  
a diminishing learning rate, 
%as also indicated by our theory (see Theorems~\ref{theorem:optimalq_q1} and \ref{theorem:optimalq_q2}): 
that combines the best of the two options, starting large to enable fast convergence, and enabling eventual fine-tuning \new{(as it is also advocated by the theory of stochastic approximation~\cite{robbins1951stochastic})}. We show curves corresponding to a diminishing learning rate both for $\OGD$ and $\OMD_\negent$, and indeed they achieve the smallest costs. The  learning rate %denoted by 
$\eta^*$ %is the learning rate that 
gives the tightest worst-case regrets for $\OGD$ and $\OMD_\negent$, as stated in Theorems~\ref{corollary:regret_qnorm_q2} and \ref{theorem:regret_negentropy}. While this learning rate is selected to protect against any (adversarial) request sequence, it is not too pessimistic: Figures \ref{fig:performance_stationary} (a) and \ref{fig:performance_stationary}~(b) show it performs well when  compared to other learning rates. 
%(\fix{see Theorems~\ref{theorem:optimalq_q1} and \ref{theorem:optimalq_q2}}):%

Figure \ref{fig:performance_stationary} (c) shows the time-average regret $\mathrm{TAR}$ of OGD and $\OMD_\negent$  over the \emph{Fixed Popularity} trace. As both algorithms have sub-linear regret, their time average regret goes to $0$ for $T \to \infty$.
Note how instead LRU exhibits a constant time average regret.
%We observe that both converge to the static optimum cost; this confirms also the sub-linearity of the regret, which implies the time average regret goes to 0 for $T\to \infty$. We also observe that the two algorithms provide a cost reduction of 20\% compared to LRU.
% \sidecaptionvpos{figure}{c}
% \begin{SCfigure}[50][t]
% \centering
%   \subcaptionbox{$\mathrm{NAC}$ of OGD}{
%     \includegraphics[width=.2\textwidth]{graphical_res/Aug/fixed_ogd.pdf}\!\!
%   }%
%   \subcaptionbox{$\mathrm{NAC}$ of $\OMD_\negent$}{
%      \centering
%     \includegraphics[width=.2\textwidth]{graphical_res/Aug/fixed_omd.pdf}\!\!
%   }%
%   \subcaptionbox{Time-Average Regret}{
%     \includegraphics[width=.25\textwidth]{graphical_res/Aug/fixed_average_regret.pdf}
%   }
%   \caption{NAC of the different caching policies over the \emph{Fixed Popularity} trace. Subfigures (a) and (b) show the performance of OGD and $\OMD_\negent$ respectively under different learning rates. For small learning rates the algorithms both converge slower but more precisely, while for larger learning rates they converge faster, but to a higher cost. Subfigure (c) shows the time average regret of the two gradient algorithms. Since the regret is sub-linear, the time average regret converges to $0$ as $T \to \infty$.  }
%   \label{fig:performance_stationary}
% \end{SCfigure}

\sidecaptionvpos{figure}{c}
\begin{SCfigure}[50][t]
  \centering
  \begin{minipage}{.7\textwidth}
  \subcaptionbox{ $k = 25, \alpha =0.1$}{
  \includegraphics[width=.3\textwidth]{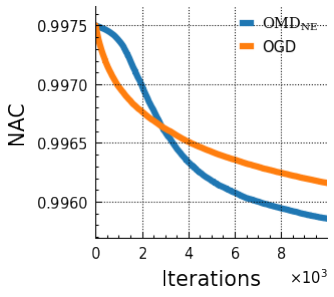}
  }
  \subcaptionbox{$k = 125, \alpha =0.1$}{
  \includegraphics[width=.3\textwidth]{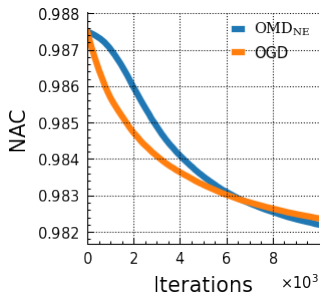}
  }
  \subcaptionbox{$k = 250, \alpha =0.1$}{
  \includegraphics[width=.3\textwidth]{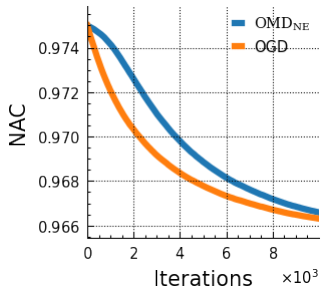}
  }\\
  \subcaptionbox{$k = 25, \alpha =0.2$}{
  \includegraphics[width=.3\textwidth]{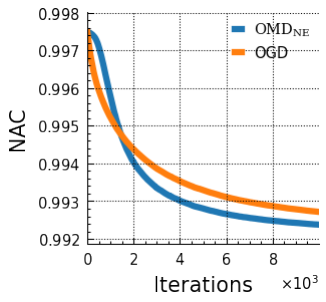}
  }
  \subcaptionbox{$k = 125, \alpha =0.2$}{
  \includegraphics[width=.3\textwidth]{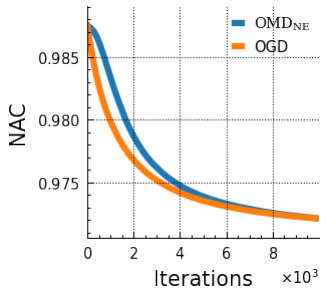}
  }
  \subcaptionbox{$k = 250, \alpha =0.2$}{
  \includegraphics[width=.3\textwidth]{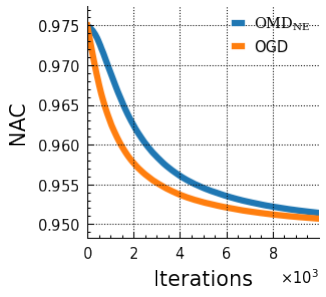}
  }\\
    \subcaptionbox{$k = 25, \alpha =0.7$}{
  \includegraphics[width=.3\textwidth]{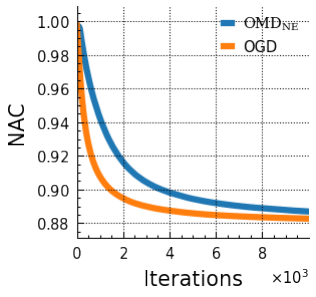}
  }
  \subcaptionbox{$k = 125, \alpha =0.7$}{
  \includegraphics[width=.3\textwidth]{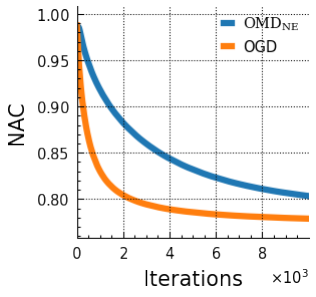}
  }
  \subcaptionbox{$k = 250, \alpha =0.7$}{
  \includegraphics[width=.3\textwidth]{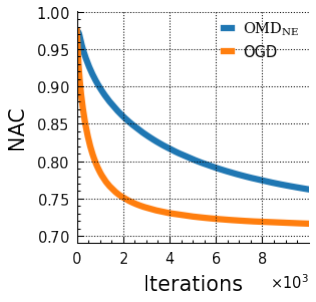}
  }
  \end{minipage}\hspace{-1.5em}
  \caption{NAC of $\OMD_\negent$ and OGD evaluated under different cache sizes and diversity regimes. We use  traces \emph{Batched Fixed Popularity} (1), (2), and (3) corresponding to different diversity regimes. Figures from left to right correspond to  different cache sizes $ k \in \{25,125,250\}$, while figures from top to bottom correspond to  different exponent values $\alpha \in \{0.1, 0.2, 0.7\}$. $\OMD_\negent$ outperforms OGD in the more diverse regimes and for small cache sizes, while OGD outperforms for large cache sizes and concentrated requests.}
  \label{fig:diversity_regimes_stationary}
\end{SCfigure}

\subsubsection{Effect of Diversity.}
Figure \ref{fig:diversity_regimes_stationary} shows the $\mathrm{NAC}$ performance of $\OMD_\negent$ and OGD on the traces \emph{Batched Fixed Popularity} (1), (2), and (3) under different cache capacities $k$ and  exponent values $\alpha$. We observe that  $\OMD_\negent$ outperforms OGD in the more diverse regimes ($\alpha \in \{0.1, 0.2\}$). This is more apparent for smaller cache sizes $k$. In contrast, OGD outperforms $\OMD_\negent$ when requests are less diverse  ($\alpha =0.7$); again, this is more apparent for larger cache size $k$. These observations agree with Theorems~\ref{theorem:optimalq_q1} and \ref{theorem:optimalq_q2} in Sec. \ref{sec:optregimes}: high diversity and small cache sizes indeed favor $\OMD_\negent$.

\subsubsection{Robustness to Transient Requests.}
\begin{figure}[!t]
  \centering
  \subcaptionbox{$\alpha = 0.1$}{
  \includegraphics[width=.22\textwidth]{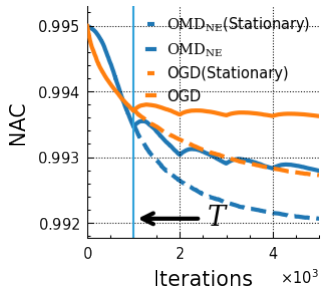}
  }
  \subcaptionbox{$\alpha = 0.3$}{
  \includegraphics[width=.22\textwidth]{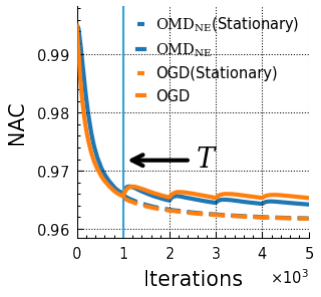}
  }
  \subcaptionbox{$\alpha = 0.4$}{
  \includegraphics[width=.22\textwidth]{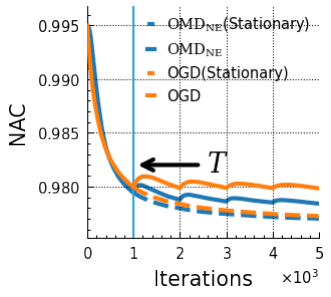}
  }
   %NAC of OGD%
  \subcaptionbox{NAC of OGD}{
    \includegraphics[width=.22\textwidth]{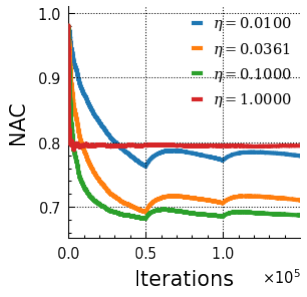}
  }
  %NAC of $\OMD_\negent$%
  \subcaptionbox{NAC of $\OMD_\negent$}{
    \includegraphics[width=.21\textwidth]{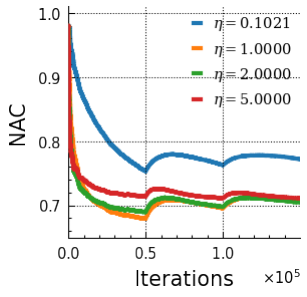}
  }
  %NAC of $\delta-\OMD_\negent$%
  \subcaptionbox{NAC of $\delta$-$\OMD_\negent$}{
    \includegraphics[width=.21\textwidth]{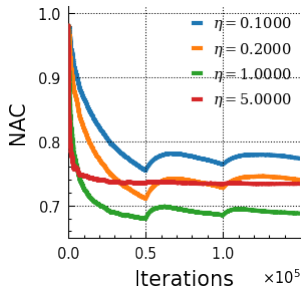}
  }
  \subcaptionbox{NAC of FTPL}{
    \includegraphics[width=.21\textwidth]{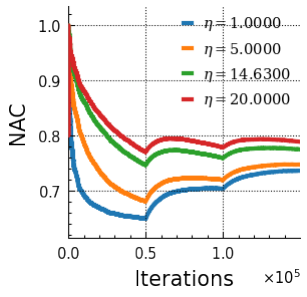}
  }
    \subcaptionbox{NAC of the different policies.
    %, LRU, Best static and Best dynamic.
    }{
    \includegraphics[width=.29\textwidth]{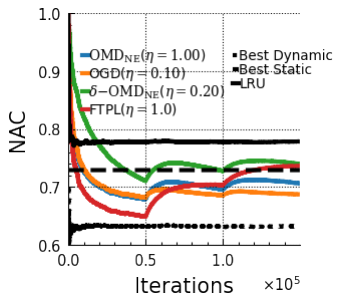}
  }

  \caption{Subfigures (a)--(c) show NAC of OGD and $\OMD_\negent$ evaluated under different diversity regimes when $10\%$ of the files change popularity over time. We use traces \emph{Partial Popularity Change} (1), (2), and (3) corresponding to the different diversity regimes. The diversity regimes are selected, such that, in the stationary setting (dashed line): (a) $\OMD_\negent$ outperforms OGD, (b) $\OMD_\negent$ has similar performance to OGD and (c) $\OMD_\negent$ performs slightly worse than OGD. %We observe that 
  $\OMD_\negent$ is consistently more robust to partial popularity changes than OGD. Subfigures (d)--(h) show the NAC of the different caching policies evaluated on the \emph{Global Popularity Change} trace, where file popularity changes every $5 \times 10^4$ iterations. %The popularity of the requests changes sharply at fixed periods.  
  While OGD adapts  seamlessly to  popularity changes (d), multiplicative updates can make $\OMD_\negent$  less reactive~(e), unless  $\OMD_\negent$ is forced to operate over the $\delta$-interior of $\mathcal{X}$~ (f) ($\delta=10^{-4}$).
  %In (e), $\OMD_\negent$, operating over  the constraint set $\mathcal{X}$, adapts slowly, because multiplicative updates make it ``stuck'' in a corner configuration; in contrast, in (d), OGD adapts more seamlessly to  popularity changes. In (f), we see that modifying $\OMD_\negent$ to operate over the $\delta$-interior of $\mathcal{X}$ resolves this issue,  the cache states are bounded away from zero. 
  Finally, our mirror descent policies outperform competitors~(h). 
  }
  \label{fig:performance_partial_popularity_change}
\end{figure}

Figure \ref{fig:performance_partial_popularity_change} shows the normalized average cost of $\OMD_\negent$ and OGD over the \emph{Partial Popularity Change} traces,  evaluated under different diversity regimes. Dashed lines indicate the projected performance in the stationary setting (if request popularities stay fixed). 
% The diversity regimes are selected to provide different performance: in (a) $\OMD_\negent$ outperforms OGD, in (b) $\OMD_\negent$ has similar performance to OGD, and in (c) $\OMD_\negent$ performs worse than OGD.
 %
 Across the different diversity regimes, we find the $\OMD_\negent$ is more robust to popularity changes. In (a), (b) and (c) $\OMD_\negent$ outperforms OGD in the non-stationary popularity setting: we observe a wider performance gap as compared to the stationary setting.
%  In (c), the algorithms exhibit similar performance.
 
 Figure \ref{fig:performance_partial_popularity_change} (d) and (e) show the normalized average cost over the  \emph{Global Popularity Change} trace for the policies OGD and $\OMD_\negent$, respectively. We observe in Figure \ref{fig:performance_partial_popularity_change}~(b) the NAC of $\OMD_\negent$ performance degrades after each popularity change. This is a limitation due to the multiplicative nature of $\OMD_\negent$. When the algorithm learns that a file, say it $i$, is not important, it can set $x_{t,i}$ arbitrarily close to $0$. If, suddenly, this content becomes popular, then $\OMD_\negent$ adapts slowly, due to its multiplicative nature---remember Eq.~\eqref{eq:multi}. This is shown in Figure~\ref{fig:performance_partial_popularity_change} (e). We can overcome this limitation by requiring all state variables to be larger than some small $\delta>0$; $\OMD_\negent$ is then limited to  $\mathcal{X}_\delta$, the $\delta$ interior of the capped simplex $\mathcal{X}$. More precisely, the $\delta$ interior of the capped simplex is defined as  $\mathcal{X}_\delta \triangleq  \mathcal{X} \cap [\delta, 1]^N $. In Figure~\ref{fig:performance_partial_popularity_change}~(f), we use $\delta = 10^{-4}$. This parameter indeed prevents the algorithm from driving the fractional allocations 
 %make caching decisions 
 arbitrary close to 0, improving its adaptability. In Figure~\ref{fig:performance_partial_popularity_change}, we show the performance of FTPL~\cite{FundamentalLimits}; we observe that this policy fails to adapt to popularity changes. Both our mirror descent algorithms outperform competitors (Fig.~\ref{fig:performance_partial_popularity_change}(h)). 

% \begin{figure}[!t]
%   \centering
%   %NAC of OGD%
%   \subcaptionbox{NAC of OGD}{
%     \includegraphics[width=.16\textwidth]{graphical_res/Aug/dynamic_ogd.pdf}
%   }
%   %NAC of $\OMD_\negent$%
%   \subcaptionbox{NAC of $\OMD_\negent$}{
%     \includegraphics[width=.16\textwidth]{graphical_res/Aug/dynamic_omd.pdf}
%   }
%   %NAC of $\delta-\OMD_\negent$%
%   \subcaptionbox{NAC of $\delta$-$\OMD_\negent$}{
%     \includegraphics[width=.16\textwidth]{graphical_res/Aug/dynamic_somd.pdf}
%   }
%   \subcaptionbox{NAC of FTPL}{
%     \includegraphics[width=.16\textwidth]{graphical_res/Aug/ftpl_changing.pdf}
%   }
%     \subcaptionbox{NAC of the different policies, LRU, Best static and Best dynamic.}{
%     \includegraphics[width=.23\textwidth]{graphical_res/Aug/compare_changing.pdf}
%   }
%   \caption{NAC of the different caching policies evaluated on the \emph{Global Popularity Change} trace. The popularity of the requests changes sharply at fixed periods.  In (b), $\OMD_\negent$, operating over  the constraint set $\mathcal{X}$, adapts slowly, because multiplicative updates make it ``stuck'' in a corner configuration; in contrast, in (a), OGD adapts more seamlessly to  popularity changes. In (c), we see that modifying $\OMD_\negent$ to operate over the $\delta$-interior of $\mathcal{X}$ resolves this issue,  the cache states are bounded away from zero. Finally, both policies outperform competitors, as shown in (d).}
%   \label{fig:performance_changing}
% \end{figure}

\subsubsection{Akamai Trace}
\begin{SCfigure}[50][t]
    \centering
    \includegraphics[width=.35\linewidth]{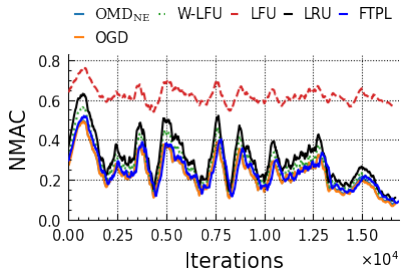}
    \caption{NMAC of the different caching policies evaluated on the \emph{Akamai Trace}. $\OMD_\negent$ and OGD provide consistently the best performance compared to W-LFU, LFU, LRU, and FTPL. OGD performs slightly better than OMD in some parts of the trace.}
    \label{fig:cdn_trace}
\end{SCfigure}

 Figure~\ref{fig:cdn_trace} shows that  the two gradient algorithms, $\OMD_\negent$ and OGD, perform similarly  over the Akamai Trace w.r.t. NMAC; OGD is slightly better in parts of the trace. Overall, these algorithms consistently outperform  LFU, W-LFU, LRU, and FTPL. Note that these caching policies process requests individually, while  $\OMD_\negent$ and OGD adapt slower, \emph{freezing} their state for the entire batch size ($R=5000$). Nevertheless, $\OMD_\negent$ and OGD still perform better. \new{Despite the observation that $\OMD_\negent$ falls behind $\OGD$ in some parts of the trace, note that in many scenarios $\OMD_\negent$ remains  an attractive choice because it provides
similar performance to $\OGD$ at a much lower computation cost (cf. Theorems~\ref{theorem:ogd_time_complexity} and~\ref{theorem:bregman_divergence}). }
\sidecaptionvpos{figure}{c}
\subsubsection{Randomized Rounding}
\begin{SCfigure}[50][t]
  \centering
\begin{minipage}{.6\textwidth}
  \subcaptionbox{Normalized average cost}{
  \includegraphics[width=.43\textwidth]{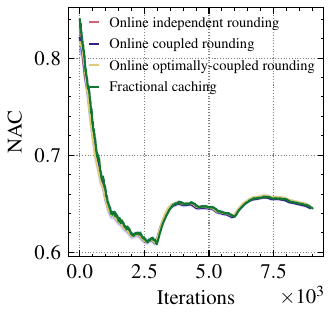}
  }
    \subcaptionbox{Cumulative update cost}{
  \includegraphics[width=.43\textwidth]{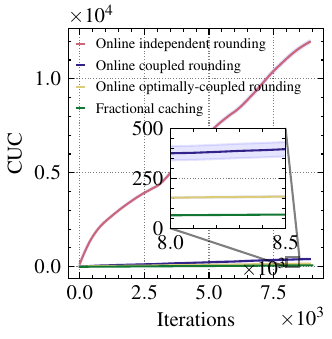}
  }
\end{minipage}
\hspace{-2em}
  \caption{Costs associated with the rounded integral caching over the \emph{Downscaled Global Popularity Change} trace. The normalized average costs shown in (a) are the same for the online independent rounding, the online coupled rounding and the online optimally-coupled rounding. The cumulative update cost of the online coupled rounding and online optimally-coupled rounding in (b) is of the same order as in the fractional setting, while the online independent rounding in (b) gives a much higher update cost. The reported values are averaged over 20 experiments, and the blue shaded area represents 10\% scaling of the standard deviation. }
  \label{fig:rounding_costs}
\end{SCfigure}

Figure \ref{fig:rounding_costs} shows the cumulative update cost for the online independent rounding, the online coupled rounding, and the online optimally-coupled rounding algorithms over the \emph{Downscaled Global Popularity Change} trace. All the rounding algorithms exhibit the same service cost in expectation. The update cost of online coupled rounding and the online optimally-coupled rounding is small, in the order of  the learning rate $\eta$; moreover, we observe that online optimally-coupled rounding yields lower update costs than the online coupled rounding. In contrast, online independent rounding incurs a significantly larger update cost.

\begin{figure}[!t]
  \centering
  \subcaptionbox{Fractional cache states}{
  \includegraphics[width=.23\textwidth]{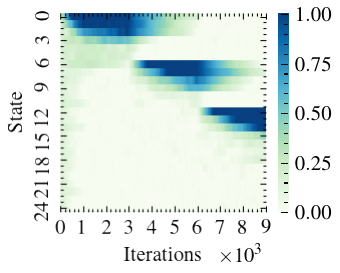}
  }
     \subcaptionbox{Online independent rounding}{
  \includegraphics[width=.23\textwidth]{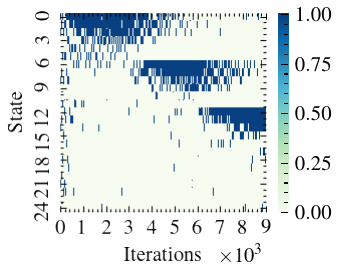}
  }
  \subcaptionbox{Online coupled rounding}{
  \includegraphics[width=.23\textwidth]{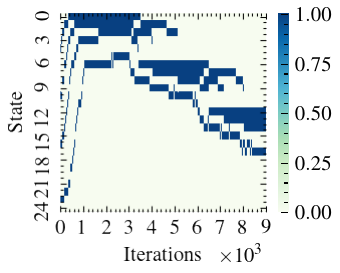}
  }
 \subcaptionbox{Online optimally-coupled rounding}{
  \includegraphics[width=.23\textwidth]{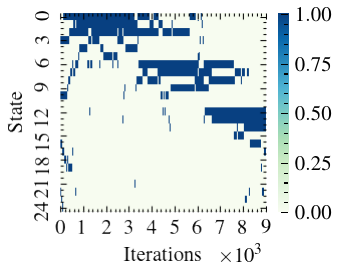}
  }
  \caption{Online rounding of fractional caching states. Visually, we see that the online independent rounding has more frequent updates than the online coupled rounding. This leads to large update costs. The online coupled rounding and the online optimally-coupled rounding  prevents the cache to perform unnecessary updates.  }
  \label{fig:rounding_caching_decisions}
\end{figure}

Figure \ref{fig:rounding_caching_decisions} shows the fractional and (rounded) integral cache states  under \emph{Downsampled Global Popularity Change} trace.   Online independent rounding indeed leads to more frequent updates than  online coupled rounding, while the latter maintains a more stable cache configuration by coupling the consecutive states and avoiding unnecessary updates.

\subsubsection{Computational Cost}
\begin{SCfigure}[40][t]
    \centering
    \begin{minipage}{.6\textwidth}
      \subcaptionbox{$R/h = 1$}{
    \includegraphics[width=.47\textwidth]{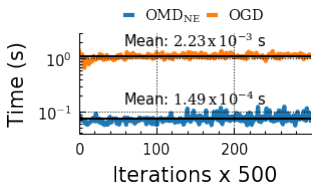}
  }
  \subcaptionbox{$R/h > 1$}{
    \includegraphics[width=.47\textwidth]{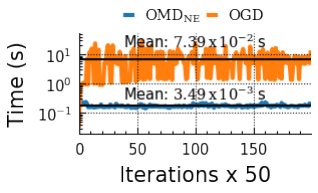}
  }  
    \end{minipage}
   
    \caption{Runtime per iteration of $\OMD_\negent$ and $\OGD$. Subfigure (a) shows the runtime per 500 iterations over the \emph{Fixed Popularity} trace.  Subfigure (b) shows the runtime per 50 iterations over the \emph{Batched Fixed Popularity (2)} trace.   }
    \label{fig:time_complexity}
\end{SCfigure}

Figure~\ref{fig:time_complexity} shows the time taken by both policies OMD and \OMDNEG{} to perform 500 iterations over the \emph{Fixed Popularity} trace (Fig.~\ref{fig:time_complexity} (a)), and the time taken to perform 50 iterations over  the \emph{Batched Fixed Popularity (2)} trace (Fig.~\ref{fig:time_complexity} (b)). We observe that \OMDNEG{} is at least 15 times faster in computing cache states on average.
% ---80 times higher than coupled rounding.

\sidecaptionvpos{figure}{c}

\sidecaptionvpos{figure}{c}

\section{Conclusions}

We study no-regret caching algorithms based on OMD with $q$-norm  and neg-entropy mirror maps.  We find that batch diversity impacts regret performance; a key finding is that OGD is optimal in low-diversity regimes, while $\OMD_\negent$ is optimal under high diversity. With an appropriately designed  rounding scheme, our $\BigO{\sqrt{T}}$ bound on the regret for general OMD algorithms extends to  integral caches as well, despite the need to account for update costs in this setting. 

Our numerical experiments indicate that the gap between the regimes in which OGD and $\OMD_\negent$ are optimal, w.r.t. the diversity ratio, is narrow; this suggests that our characterization of the two regimes can be further improved. Also $\OMD_{q\nrm}$ algorithms for arbitrary values of $q\in(1,2)$ deserves more investigation to 1) devise strongly polynomial, efficient algorithms for their Bregman projection, 2) characterize their update costs, and 3) compare their performance with $\OMD_\negent$.

%mirror descent algorithm using  comparison of $\OMD_\negent$ with other Other open research problems include extending our analysis of studying which 

%proving the optimality of $\OMD_\negent$ within the $q$-norm mirror map class, as opposed to merely its domination over OGD, is also open. Finally, devising strongly polynomial, efficient algorithms for the Bregman projection when $q\in(1,2)$, and characterizing update costs in this setting, also remain open problems.

\noindent \textbf{Acknowledgements.} This research was supported in part by the French Government through the ``Plan de Relance'' and ``Programme
d’investissements d'avenir'' and by Inria under the exploratory action MAMMALS. The authors gratefully acknowledge support from the National Science Foundation (grants 2107062 and 2112471).

\bibliographystyle{unsrt}
\bibliography{Bibliography}

\begin{appendices}

% \appendix
\vspace{-.75em}\section{Fractional Caching and Gradient-based algorithms}

\subsection{Online Mirror Descent}
\begin{theorem}
\!\!\!{\protect\textrm{(\cite[Theorem 4.2]{bubeck2015convexbook})}}
\label{theorem:bubeckomdbound}
Let (1) the map $\Phi: \mathcal{D} \to \reals$ be a mirror map (see Sec.~\ref{section:OMD}) $\rho$-strongly convex w.r.t a norm $\norm{\,\cdot\,}$ over $\mathcal{S} \cap \mathcal{D}$ ($\mathcal{S}$ is a convex set), (2) the cost functions $f_{t}: \mathcal{S} \to \reals$ be convex with bounded gradients (i.e., $\norm{\nabla f_t(\x)}_* \leq L, \forall \x \in \mathcal{S}$) for every $t \in [T]$, where $\norm{\,\cdot\,}_*$ is the dual norm of $\norm{\,\cdot\,}$,
    (3) and the Bregman divergence $D_{\Phi}(\vec{x}, \vec{x}_1)$ be bounded by $D^2$ for $\x \in \mathcal{S}$ where $\x_1 = \argmin_{\x \in \mathcal{S} \cap \mathcal{D}} \Phi(\x) $.
Then Algorithm~\ref{algo:online_mirror_descent} satisfies 
\begin{align}
\textstyle\sum^T_{t=1} f_t(\x_t) - \sum^T_{t=1} f_t(\x)   &\leq 
		{D^2}/{\eta} + {\eta L^2 }/({2 \rho} T), & & \text{for $\x \in S$.} 
\label{eq:omd_general_regret}
\end{align}
% where $\x$ is a fixed point in $\mathcal{S}$.
% Moreover, when we select the learning rate $\eta = \frac{D}{L} \sqrt{\frac{2\rho}{T}}$ yielding the tightest bound we have
% \begin{align}
%     	\sum^T_{t=1} f_t(\x_t) - \sum^T_{t=1} f_t(\x)   \leq LD \sqrt{\frac{2 T}{\rho}}.
%     		\label{eq:omd_general_simple_regret}
% \end{align}
\end{theorem}
We remark that Theorem~\ref{theorem:bubeckomdbound} is expressed differently in~\cite{bubeck2015convexbook}, where $f_t = f, \forall t \in [T]$ (fixed cost function). Nonetheless, as observed in \cite[Sec.~4.6]{bubeck2015convexbook} the bound obtained in Eq.~\eqref{eq:omd_general_regret} holds as long as the dual norms of the gradients are bounded by $L$.

\vspace{-.75em}\subsection{Proof of Theorem~\ref{theorem:qnorm_regret}}
\label{proof:qnorm_regret}

% Bubeck \cite[Theorem 4.2]{bubeck2015convexbook} provides a general bound on the regret of OMD with $\rho$-strongly convex mirror map $\Phi$ w.r.t a norm $\norm{\,\cdot\,}$ on a convex set $\mathcal{S}$. 
The map $\Phi(x) = \frac{1}{2}\norm{x}_q^2, q \in (1, 2]$ is $\rho=q-1$ strongly convex w.r.t $\norm{\,\cdot\,}_q$ over $\mathcal{D} = \mathbb{R}^N$ a direct result from \cite[Lemma 17]{Shalev2007}, and the dual norm of $\norm{\,\cdot\,}_q$ is $\norm{\,\cdot\,}_p$ (Hölder's inequality).Take $\mathcal{S} = \mathcal{X}$.
%
% Using Theorem~\ref{theorem:bubeckomdbound}, we obtain
% \begin{equation}
% 		\Regret_T(\OMD_\Phi)  \leq 
% 		\frac{D_{\Phi}(\vec{x}_*, \vec{x}_1)}{\eta} + \frac{\eta}{2 \rho} \sum^T_{t=1} \norm{\nabla f_{r}(\vec{x}_t)}^2_p, 
% 	\label{eq:omd_general_regret}
% \end{equation}
% where $\vec{x}_*$ is the best fixed decision in hindsight.
%
%
The minimum value of $\Phi(x)$ over $\mathcal{X}$ is achieved when we spread the capacity mass $k$ over the decision variable, i.e.,  $x_i = \frac{k}{N}, i \in \mathcal{N}$. If we select $\vec{x}_1$ to be the minimizer of $\Phi(\vec{x})$, then we have $\nabla \Phi^T(\vec{x}_1) (\vec{x} - \vec{x}_1) \geq 0 , \forall \vec{x} \in \mathcal{X}$ \cite[Theorem~2.2]{Hazanoco2016}, so we obtain 
$
   D_\Phi(\vec{x}, \vec{x}_1) = \Phi(\vec{x}) - \Phi(\vec{x}_1) - \nabla {\Phi(\vec{x}_1)}^T (\vec{x} - \vec{x}_1) \leq \Phi(\vec{x}) - \Phi(\vec{x}_1)
$.
Moreover, it is easy to check that $\Phi$ is maximized at a sparse point $\x_* \in \mathcal{X} \cap \{0,1\}^N$; thus, we have $
     D_\Phi(\vec{x}, \vec{x}_1) \leq \Phi(\vec{x}_*) - \Phi(\vec{x}_1)$.
By replacing $\x_1$ and $\x_*$ with their values in the previous equation we get  $  \textstyle \Phi(\vec{x}_1) =\frac{1}{2} \left(\left(\frac{k}{N}\right)^q N\right)^{\frac{2}{q}} = \frac{1}{2} k^2 N^{-\frac{2}{p}}$, and $  \textstyle  \Phi(\vec{x}_*) =\frac{1}{2} k^{\frac{2}{q}} = \frac{1}{2} k^2 k^{-\frac{2}{p}}$. Thus, we have  $D_\Phi(\vec{x}, \vec{x}_1) \leq \frac{1}{2} k^2 \left(k^{-\frac{2}{p}} - N^{-\frac{2}{p}} \right) = D^2$.
% \begin{equation}
%      D_\Phi(\vec{x}, \vec{x}_1) \leq \frac{1}{2} k^2 \left(k^{-\frac{2}{p}} - N^{-\frac{2}{p}} \right) = D^2.
%      \label{eq:qnorm_bregman_distance}
% \end{equation}
Note that the maximum of $\norm{\vec{r}}_p$ is achieved when $\frac{R}{h}$ components are set to $h$, then the following bound holds on the gradients $\underset{r \in \mathcal{R}}{\max} \norm{ \nabla f_{\vec{r}} (\vec{x})}_p  \leq \underset{\vec{r} \in \mathcal{R}}{\max}  \norm{\vec{w}}_\infty \norm{ \vec{r}}_p  = \norm{\vec{w}}_\infty h \left( \frac{R}{h}\right)^\frac{1}{p} = L_p$.  The gradients are bounded in the dual norm $\norm{\nabla f_{r}(\vec{x}_t)}_p \leq L_p, \forall \vec{r} \in \mathcal{R}$. The final bound follows by Theorem~\ref{theorem:bubeckomdbound},  plugging the constants $D$ and $L$ in~\eqref{eq:omd_general_regret}, and selecting the learning rate that achieves the tightest bound $\eta = \sqrt{{(q-1) k^2 \left(k^{-\frac{2}{p}}  - N^{-\frac{2}{p}}\right)} /\left( {\norm{w}^2_\infty h^2 \left(\frac{R}{h}\right)^{\frac{2}{p}} T}\right)}$. \qed

% \vspace{-.75em}\subsection{Proof of Corollary~\ref{corollary:regret_qnorm_q2}}
% \label{proof:regret_qnorm_q2}

% \sloppy
% Using the regret bound given in~\eqref{eq:omd_general_regret}, and setting $q = 2$, we obtain $
%     \text{Regret}_T(\text{OGD}) \leq \norm{\vec{w}}_\infty h k \left(\frac{R}{h}\right)^{\frac{1}{2}} \sqrt{\left(k^{-1} - N^{-1}\right)T} =\norm{\vec{w}}_\infty\sqrt{ hR k \left(1 - \frac{k}{N}\right) T}$. \qed %This concludes the proof.
    
%     \fussy
\vspace{-.75em}\subsection{Proof of Corollary~\ref{corollary:regret_qnorm_q1}}
\label{proof:regret_qnorm_q1}
Taking $\alpha = \frac{k}{N}$ and $\beta = \frac{N h }{R}$, we can rewrite~\eqref{eq:qnorm_regret} to have $\text{Regret}_T(\text{OMD}_{q\text{-norm}})  \leq \norm{\vec{w}}_\infty R \beta^{\frac{1}{q}} \sqrt{\frac{\alpha^{2/q} - \alpha^2}{q-1} T}$. We take the limit $q \to 1$, to obtain the upper bound
\begin{align*}
 &\textstyle\text{Regret}_T(\text{OMD}_{1\text{-norm}})  \leq \underset{q\to 1}{\text{lim}} \norm{\vec{w}}_\infty R \beta^{\frac{1}{q}} \sqrt{\frac{\alpha^{2/q} - \alpha^2}{q-1} T}= \norm{\vec{w}}_\infty R \beta \sqrt{ \left[ \alpha ^{2/q}\right]'_{q=1} T}\\
 &=\textstyle \norm{\vec{w}}_\infty R \beta \sqrt{\left[ -2 q^{-2} \alpha^{2/q}\log(\alpha)\right]_{q=1} T}
 = \norm{\vec{w}}_\infty R \alpha \beta \sqrt{ 2 \log(\alpha^{-1}) T} = \norm{\vec{w}}_\infty h k \sqrt{ 2 \log\left(\frac{N}{k}\right) T}. \qed
\end{align*}

\vspace{-.75em}\subsection{Proof of Theorem~\ref{theorem:optimalq_q2}}
\label{proof:optimalq_q2}
We take the simplified version of the regret of the general class of $q$-norm mirror maps in Eq.~\eqref{eq:qnorm_regret}, select $\alpha = \frac{k}{N}$ and $\beta = \frac{N h }{R}$, so we get $
 \text{Regret}_T(\text{OMD}_{q\text{-norm}})  \leq  \norm{\vec{w}}_\infty R \phi(q) \sqrt{T}$, where $\phi(q) \triangleq \beta^{\frac{1}{q}} \sqrt{\frac{\alpha^{2/q} - \alpha^2}{q-1}}$.
The tightest regret bound is achieved with $q^*$ that minimizes $\phi(q)$.
We have %compute   the first derivative as 
\begin{align}
\textstyle\phi'(q) = - \alpha^2 \beta^{\frac{1}{q}} \frac{q^2 \left(\alpha^{2/q - 2} - 1\right)+ 2 (q-1) \left( \alpha^{2/q - 2}\log(\alpha)  + \left(\alpha^{2/q - 2} - 1\right) \log (\beta)\right) }{2 q^2 (q-1)^2 \sqrt{\frac{\alpha^{2/q} - \alpha^2}{q-1}}}.\label{eq:minimizer_q_norm}
\end{align}
We study the sign ($-J(q)$) of the derivative of the minimizer in Eq~\eqref{eq:minimizer_q_norm}
\begin{align}
\textstyle &J(q) = \textstyle q^2 \left(\alpha^{2/q - 2} - 1\right)+ 2 (q-1) \left( \alpha^{2/q - 2}\log(\alpha)  + \left(\alpha^{2/q - 2} - 1\right) \log (\beta)\right)\\
&\geq \textstyle 2 q (1-q) \log(\alpha) + 2 (q-1)\left(2\frac{1-q}{q}  \log(\alpha) \log(\alpha \beta) + \log(\alpha) \right) \\
&\geq  \textstyle 2 q (q-1) \left( \left(1-q\right)\log(\alpha) + 2\frac{1-q}{q} \log(\alpha) \log(\alpha \beta) \right). 
\end{align}
Note that $\left(1-q\right)\log(\alpha) \geq 0$ and $\frac{1-q}{q} \log(\alpha) \geq 0$. We take $\frac{R}{h} \leq k$, this gives $\alpha \beta \geq 1$ and $J(q) \geq 0$ implying $ \text{sign}(\phi'(q)) = - \text{sign}(J(q)) =-1$. We conclude that
$\phi(q)$ is a decreasing function of $q \in (1,2]$ when $\frac{R}{h} \leq k$; therefore, the minimum is obtained at $q = 2$ for $\frac{R}{h} \leq k$. \qed
\vspace{-.75em}\subsection{Proof of Theorem~\ref{theorem:optimalq_q1}}
\label{proof:optimalq_q1}
We have the following regret upper bound for the $q$-norm mirror map, as $q \to 1$ from Corollary~\ref{corollary:regret_qnorm_q1}: $
\text{Regret}_T(\text{OMD}_{1\text{-norm}}) \leq \norm{\vec{w}}_\infty h k \sqrt{ 2\log\left(\frac{N}{k}\right) T}$. In \cite{chesneau2020new}, it is proved that the $\log$ function satisfies $\text{log}(u + 1) \leq \frac{u}{\sqrt{u + 1}}, u \geq 0$.
We take $u = \frac{N}{k} - 1$, and  note that $N \geq k>0$, so  we get $u \geq 0$. We have the following $
\text{log}\left(\frac{N}{k}\right) \leq \frac{N - k}{\sqrt{Nk}} =  \sqrt{\frac{N}{k}} \left( 1 - \frac{k}{N}\right)$.
Thus, the upper bound in  Corollary~\ref{corollary:regret_qnorm_q1} Eq.~\eqref{eq:regret_qnorm_q1} can be loosened to obtain  $\textstyle \text{Regret}_T(\text{OMD}_{1\text{-norm}}) \leq   \norm{\vec{w}}_\infty  k h\sqrt{2 \text{log}\left(\frac{N}{k}\right) T} \leq \norm{\vec{w}}_\infty   \sqrt{2 \sqrt{ N k } h^2 k \left(1 - \frac{k}{N}\right) }$. If we take $\frac{R}{h} \geq 2 \sqrt{Nk}$, then this upper bound is tighter than the upper bound on the regret of OGD in Corollary~\ref{corollary:regret_qnorm_q2}. \qed
\vspace{-.75em}\subsection{\OMDNEG{} and \OMD{} with $q$-norm Mirror Map Correspondence}

% We first provide the following supporting Lemma.
% \begin{lemma}
% \label{lemma:supporting_equiv_qnorm}
% For every $\x \in \mathcal{X} \cap \mathbb{R}^N_{>0}$ we have $ \textstyle\lim_{q\to 1} \frac{\norm{\x}^2_q - k^2}{(q-1)}=  -2k^2 \log(k) +2 k   \sum^N_{i=1} x_i \log(x_i)$.
% \end{lemma}
% \begin{proof}
% Because $\norm{\x_1}_1$, we have \begin{align}
%     \textstyle\lim_{q\to 1} \frac{\norm{\x}^2_q -  k^2}{(q-1)} = h'(1)\label{eq:limit_equivalence},
% \end{align}
% where 
% \begin{align*}
%  \textstyle h'(q) = \deriv{\norm{\x}^2_q}{ q} = \deriv{\left(\sum^N_{i=1} x_i^q\right)^{2/q}}{q} = \deriv{\left(2/q \log\left( \sum^N_{i=1} x^q_i\right)\right)}{q}\norm{\x}^2_q  =  \left(-\frac{2}{q^2}\log\left( \sum^N_{i=1} x^q_i\right) + \frac{2\sum^N_{i=1} x_i^q \log(x_i)}{q\sum^N_{i=1} x^q_i}\right) \norm{\x}^2_q
% \end{align*}
% The value of the above equation for $q = 1$ is $
%      h'(1) =  \left(-2 \log(k) + \frac{2}{k}\sum^k_{i=1} x_i \log(x_i)\right) k^2= -2k^2 \log(k) +2 k   \sum^N_{i=1} x_i \log(x_i)$. Plugging the above equality in Eq.~\eqref{eq:limit_equivalence} concludes the proof.
% \end{proof}
% We denote the limiting algorithm of $\OMD_{q\nrm}$ for $q$ converges to $1$ as  $\OMD_{1\nrm}$. The next theorem shows the relation between  $\OMD_{1\nrm}$ and $\OMD_{\negent}$

%  The following theorem demonstrates that such limiting algorithm shares the same update rule with  \OMDNEG{} (excluding the projection step) over the simplex, and they are equivalent over the simplex.

% We define $\OMD_{1\norm}$ as the limuiting algorith

\begin{theorem}
\label{theorem:equivalent_1norm_negentropy}
The algorithm $\OMD_{1\nrm}$ defined as the limitting algorithm obtain by taking $q$ converges to 1 of  $\OMD_{q\nrm}$  with learning rate $\eta_q = \eta  (q-1) k$, intermediate states $\vec y^{(q)}_{t}$, and fractional states $\vec x^{(q)}_{t}$ for $t \geq 1$  is equivalent to \OMDNEG{} configured with learning rate $\eta \in \mathbb{R}_+$  over the simplex (capped simplex $\mathcal{X}$ with $k = 1$), when both policies are initialized with same state in $\mathbb{R}^N_{>0} \cap \mathcal{X}$. Moreover, $\OMD_{1\nrm}$ has a multiplicative update rule over the capped simplex.
\end{theorem}
\begin{proof}
Let $\vec g_t= \nabla f_{\vec r_t}(\x_t)$ be the gradient of the cost function  at time slot $t$. From lines 2--3 in   Algorithm~\ref{algo:online_mirror_descent} and Eq.~\eqref{eq:q-norm-update} we obtain the following $
    \hat{y}^{(q)}_{t,i} = \left(\nabla \Phi(\vec{x}_t)\right)_i + \eta_q g_{t,i} = \text{sign} (x_{t,i})\frac{|x_{t,i}|^{q-1}}{\norm{\vec{x}_t}^{q-2}_q} - \eta_q g_{t,i}$
for a given $q \in (1,2]$. The algorithm guarantees that $\x_t \in \mathcal{X} \cap \reals^N_{>0}$, then $x_i > 0, \forall i \in \mathcal{N}$. So, we get $
   \textstyle \hat{y}^{(q)}_{t,i} =\frac{(x_{t,i})^{q-1}}{\norm{\vec{x}_t}^{q-2}_q} - \eta_q g_{t,i}, \forall i \in \mathcal{N}$.
Note that $ - \eta g_{t,i}$ is non-negative. The numbers $p$ and  $q$ are conjugate numbers (see Sec.~\ref{subsection:q-norm-mirror-maps}) and satisfy $q-1 = \frac{1}{p-1}$.  
We use Eq.~\eqref{eq:q-norm-update2} to get the expression of $y^{(q)}_{t+1,i}$  as
\begin{align}
   &\textstyle\frac{\left(\frac{(x_{t,i})^{q-1}}{\norm{\vec{x}_t}^{q-2}_q} - \eta_q g_{t,i}\right)^{p-1}}{\left(\sum_{i \in \mathcal{N}} \left(\frac{(x_{t,i})^{q-1}}{\norm{\vec{x}_t}^{q-2}_q} - \eta_q g_{t,i}\right)^p\right)^{\frac{p-2}{p}}} = \frac{\left(\frac{(x_{t,i})^{q-1}}{\norm{\vec{x}_t}^{q-2}_q} - \eta  (q-1) k g_{t,i} \right)^{p-1}}{\left(\sum_{i \in \mathcal{N}} \left(\frac{(x_{t,i})^{q-1}}{\norm{\vec{x}_t}^{q-2}_q} - \eta  (q-1) k g_{t,i}\right)^p\right)^{\frac{p-2}{p}}} = \textstyle\frac{ x_{t,i} \norm{\vec{x}_t}^{(2-q)}_q \left(1 - \eta  (q-1) k g_{t,i} \frac{\norm{\vec{x}_t}^{q-2}_q}{(x_{t,i})^{q-1}} \right)^{p-1}}{\left(\sum_{i \in \mathcal{N}} (x_{t,i})^{{(q-1)p}} \left(1- \eta  (q-1) k g_{t,i} \frac{\norm{\vec{x}_t}^{q-2}_q}{(x_{t,i})^{q-1}} \right)^p\right)^{\frac{p-2}{p}}}.\nonumber
\end{align}
We rewrite the above expression solely in terms of $p$, and taking the limit for $q$ converges to $1$ is equivalent  to let $p$ diverges to $+\infty$, so we have for all $i \in \mathcal N$

\begin{align}
  y_{t+1,i} \triangleq \lim_{p \to +\infty} y^{\left(\frac{p}{p-1}\right)}_{t+1,i} &= \textstyle x_{t,i} k   \left( \lim_{p \to +\infty} \frac{\left(1 - \frac{\eta g_{t,i}}{p-1}\right)^{p-1}}{\left(\sum_{i \in \mathcal{N}} x_{t,i} \left(1-  \frac{\eta g_{t,i}}{p-1}  \right)^p\right)^{\frac{p-2}{p}}}\right) =  \textstyle x_{t,i} k \frac{ \exp \left(-\eta g_{t,i}\right)}{\sum_{i \in \mathcal{N}} x_{t,i} \exp \left(-\eta g_{t,i}\right) }.
\label{eq:mul_update_omd}
\end{align}
% Note that  $\norm{\vec y_{t+1}}_1 = \sum_{i \in \mathcal{N}} x_{t,i} k \frac{ \exp \left(-\eta g_{t,i}\right)}{\sum_{i \in \mathcal{N}} x_{t,i} \exp \left(-\eta g_{t,i}\right) } = k$.
The intermediate state of $\OMD_{1\nrm}$  in  Eq.~\eqref{eq:mul_update_omd} is a multiplicative update rule identical to the update rule of \OMDNEG~($y_{t+1,i} = x_{t,i}\,e^{-\eta g_{t,i}}$ in Eq.~\eqref{eq:multi}) with an additional multiplicative factor $  \frac{ k}{\sum_{i \in \mathcal{N}} x_{t,i} \exp \left(-\eta g_{t,i}\right) }$. For $k = 1$, the intermediate state of $\OMD_{1\nrm}$  in  Eq.~\eqref{eq:mul_update_omd} is feasible (i.e., $\x_{t+1} = \vec y_{t+1} $ and the projection has no effect). On the other hand, the neg-entropy projection in the case of $k=1$ is just a  normalization of the intermediate states (i.e., $x_{t+1,i} = \frac{x_{t,i}\,e^{-\eta g_{t,i}}}{\sum_{i \in \mathcal{N}} x_{t,i} \exp \left(-\eta g_{t,i}\right) }$ for $ i \in \mathcal{N}$). Thus, the states obtained by the two algorithms coincide. 
% Note that  $\norm{\vec y_{t+1}}_1 = \sum_{i \in \mathcal{N}} x_{t,i} k \frac{ \exp \left(-\eta g_{t,i}\right)}{\sum_{i \in \mathcal{N}} x_{t,i} \exp \left(-\eta g_{t,i}\right) } = k$. The intermediate state $\y_{t+1}$ is obtained through a multiplicative update rule similar to \OMDNEG{}, except we have the intermediate state $\y_{t+1}$ is scaled by $\frac{k}{\sum_{i \in \mathcal{N}} x_{t,i} \exp \left(-\eta g_{t,i}\right) }$. In the simplex (capped simplex with $k=1$) the projection has no effect, and it is easy to check  the obtained states are always feasible.
\end{proof}

\vspace{-.75em}\subsection{Proof of Theorem~\ref{theorem:regret_negentropy}}
\label{proof:regret_negentropy}

The neg-entropy mirror map is $\rho=\frac{1}{k}$-strongly convex w.r.t the norm $\norm{\,\cdot\,}_1$ over $\mathcal{X} \cap \mathcal{D}$~\cite[Example~2.5]{ShalevOnlineLearning}. The dual norm of $\norm{\,\cdot\,}_1$ is $\norm{\,\cdot\,}_\infty$. By taking $p \to \infty$ of $L_p$ in the Proof of Theorem~\ref{theorem:qnorm_regret}  we can consider as bound for the gradient in Eq.~\eqref{eq:omd_general_regret}
\begin{align}
\label{e:proof_negentropy_ub_L}
    L = \norm{\vec{w}}_\infty h.
\end{align}
The initial  state $\x_1$ with $x_{1,i} =k / N, \forall i \in \mathcal{N}$ is the  minimizer of $\Phi$, and we have $\Phi(x) \leq 0, \forall \x \in X$. Thus 
\begin{equation}
\textstyle    D_\Phi(\x, \x_1) \leq \Phi(\x) - \Phi(\x_1) \leq -\Phi(\x_1) = -\sum^N_{i=1} \frac{k}{N} \log\left(\frac{k}{N}\right) = k \log\left(\frac{N}{k}\right) = D^2.
    \label{e:proof_negentropy_ub_B}
\end{equation}
The bound follows by  Theorem~\ref{theorem:bubeckomdbound},  plugging~\eqref{e:proof_negentropy_ub_L} and~\eqref{e:proof_negentropy_ub_B} in~\eqref{eq:omd_general_regret}, and  selecting the learning rate that gives the tightest upper bound, that is $\eta = \sqrt{\frac{2 \log(\frac{N}{k})}{\norm{w}^{2}_\infty h^2 T}}$. \qed %This gives the desired upper bound of regret.

\vspace{-.75em}\subsection{Proof of Theorem~\ref{theorem:bregman_divergence}}
\label{proof:bregman_divergence}
%\begin{proof}
We adapt the Euclidean projection algorithm in \cite{wang2015projection}. 
Finding the projection $\mathbf{x} = \Pi^{\Phi}_{\mathcal{X} \cap \mathcal{D}}(\mathbf{y})$ corresponds to solving a convex problem as $D_{\Phi}(\mathbf x, \mathbf y)$ is convex in $\mathbf x$ and $\mathcal X \cap \mathcal D$ is a convex set.
Without loss of generality, we assume the components of $\mathbf{x} = \Pi^{\Phi}_{\mathcal{X} \cap \mathcal{D}}(\mathbf{y})$ to be in non-decreasing order. Let $b \in \mathcal{N}$ be the index of the largest component of $\mathbf{x}$ smaller than 1. The KKT conditions lead to conclude that 
if the components of $\mathbf{y}$ are ordered in ascending order, so are the components of $\mathbf{x}$. 
In particular, the smallest $b$ components of $\mathbf{x}$ can be obtained as $x_i = y_i e^\gamma $ and  $y_{b} e^\gamma  < 1 \leq y_{b+1} e^\gamma$, where $\gamma$ is the Lagrangian multiplier associated with the capacity constraint. If $b$ is known, then it follows from the capacity constraint that $\textstyle m_b \triangleq e^{\gamma} =\textstyle  \frac{k + b - N }{\sum_{i=1}^{b}{y_i}} =\textstyle  \frac{k+b-N}{\norm{\mathbf y}_1 - \sum_{i=b+1}^{N} {y_i}}$. 
We observe that necessarily $b \in \{ N - k +1, \dots, N\}$. In fact, we cannot have $b \leq N-k$. If $b \le N - k$, we get $\sum_{i = N-k+1}^{N} x_i \ge k$ and the capacity constraint implies that $ x_i = 0, \forall i \leq b$, but we must have  $x_i > 0$ since $\mathbf{x} \in \mathcal{X} \cap \mathcal{D}$ and $\mathcal{D} = R_{>0}^N$.
We can then find the value of $b$, but checking which number in $\{ N - k +1, \dots, N\}$ satisfies $y_{b} e^\gamma  < 1 \leq y_{b+1} e^\gamma$. Note that this operation only requires 
%, therefore, $b$ can only take the values $b \in \{N-k+1,..., N\}$. 
%As $b > N - k$, then we only require 
the largest $k$ components of $\mathbf{y}$. % to perform the projection.
%For a given $b$ let $\textstyle m_b \triangleq e^{\gamma} =\textstyle  \frac{k + b - N }{\sum_{i=1}^{b}{y_i}} =\textstyle  \frac{k+b-N}{\norm{\mathbf y}_1 - \sum_{i=b+1}^{N} {y_i}}$. 
The projection corresponds to setting the components $y_{b+1}, \dots, y_N$ to 1 and multiply the other $N-b$ components by $m_b$. In order to avoid updating all components at each step, we can simply set the components $x_i$ for $i>b$ (those that should be set equal to 1) to $\frac{1}{m_b}$. Then, at any time $t$, we can recover the value of $x_{t,i}$, multiplying the $i$-th component of the vector $\mathbf x$ by $P = \prod_{s=1}^{t}{m_{b,s}}$, where $m_{b,s}$ is the returned $m_b$ from the Bregman projection at time step $s$. 
%Then at any time step, the actual value of the decision variable is $\mathbf{x} P$. 
For general values of $R$ and $h$, the projection step takes $\BigO{k}$ steps per iteration and a partial sort is required to maintain top-$k$ components of $\mathbf{y}_t$ sorted; this can be done using partial sorting in $\BigO{N + k \log(k)}$  \cite{paredes2006optimal}.
When $R = h = 1$, Alg.~1 leads to only a single state coordinate update, and requires $\BigO{\log(k)}$ steps to maintain top-$k$ components of $\mathbf{x}_t$ sorted online. \qed
%\end{proof}

\vspace{-.75em}\subsection{Proof of Proposition~\ref{prop:free_updates}}
\label{proof:free_updates}
%\begin{proof}
Every time slot $t \in [T]$, we obtain an intermediate cache state  $\y_{t+1}$ through lines 2--4 in Algorithm~\ref{algo:online_mirror_descent} as $\y_{t+1} = (\nabla \Phi^{-1}) (\nabla \Phi(\x_t) - \eta \nabla f_{\vec r_t} (\x_t))$. Let $\{\x_t\}^T_{t=1}$ and $\{\x'_t\}^T_{t=1}$ be fractional cache states obtained by \OGD{} and \OMDNEG, respectively, and  $\{\y_t\}^T_{t=1}$ and $\{\y'_t\}^T_{t=1}$  be their intermediate fractional  cache states.
In the case of \OGD{}, or equivalently \OMD{} configured with mirror map $\Phi(\x) =\frac{1}{2} \norm{\x}^2_2$, the intermediate fractional states have the same components as the previous cache state for files are not requested $\y_{t,i} = \x_{t,i}$ for every $i \not\in \supp(\vec r_t)$. Similarly, we also have $\y'_{t,i} = \x'_{t,i}$ for \OMDNEG{} for every $i \not\in \supp(\vec r_t)$.
The Euclidean projection algorithm  onto the capped simplex~\cite{wang2015projection} can only set a component of the intermediate fractional cache state to one if it exceeds it, and the remaining components are either set to zero or reduced by a constant amount 
$
    \Delta = \frac{N - b - k + \sum^b_{j = a+1}{y_{t+1, j}}}{b-a},
$
where $a$ is the number of components set to zero and $b$ is the number of components strictly less than one,  and $\Delta \geq y_{t+1,a}$ (a KKT condition in~\cite{wang2015projection}) and in turn $\Delta \geq 0$ because $y_{t+1,i} \geq 0$ for any $ i \in \catalog$.
Therefore, all the components $i \not\in  \supp(\vec r_t)$ of the resulting state  are decreased or at most kept unchanged.
Similarly, the neg-entropy Bregman projection onto the capped simplex sets some components to one if they exceed it, and the remaining components are scaled by a constant $m_b$.  In our caching setting we have $y_{t+1, i} \geq x_{t,i}$ for $i \in \catalog$ in turn $\norm{\vec y_{t+1}}_1 \geq k$, thus the equality constraint $\norm{\vec x}_1 = k$ in the projection can be replaced by $\norm{\vec x}_1 \leq k$. From the KKT dual feasibility condition we obtain $-\gamma \geq  0$ and $m_b = e^\gamma \leq 1$. Thus, we have $x_{t+1,i} \leq x_{t,i}$ for every $ i \not \in \supp(\vec r_t)$.  We conclude that the update cost is zero for both policies, i.e., $\updatecost_{\vec r_t} (\vec x_t, \vec x_{t+1}) = \textstyle \sum_{i \notin \supp(\vec{r}_t)} w_i' \max(0, x_{t+1,i} - x_{t,i}) = 0$. \qed
%\end{proof}

\vspace{-.75em}\section{Integral Caching}
\label{app:online_rounding}
%     \subcaptionbox{}[0.4\linewidth]{\includegraphics[trim=90 0 90 0,clip,width=\linewidth]{graphical_res/Aug/scenario1.pdf}}
    %  \subcaptionbox{}[0.4\linewidth]{\includegraphics[trim=90 0 90 0,clip,width=\linewidth]{graphical_res/Aug/movement_costs.pdf}}

\subsection{Proof of Proposition~\ref{proposition:deterministic_worst_case_regret}}
\label{proof:deterministic_worst_case_regret}
%\begin{proof}
Consider equal service costs $w_i = 1$ for any $i$ in $\mathcal{N}$. A deterministic policy denoted by $\mathcal{A}$ selects an integral cache state $\x_t$ from $\mathcal{Z}$ for every time slot $t$, and the adversary can select a request batch $\vec r_t$ based on the selected state. Let $\vec r_t = [\mathds{1}_{\{x_{t,i} \neq 1\}}]_{i\in \catalog}$ be the request batch selected by the adversary at time $t$, so the cost incurred at any time slot $t$ is $f_{\vec r_t} (\vec x_t) = N-k$, and the total cost incurred by $\mathcal{A}$ for the time horizon $T$ is $\sum^T_{t=1}  f_{\vec r_t} (\vec x_t) = (N-k) T$. For a fixed integral cache state $\x \in \mathcal{Z}$, %, the following cost is incurred 
\begin{align}
  \textstyle \sum^T_{t=1} f_{\vec r_t} (\x) &= \textstyle\sum^T_{t=1} \sum^N_{i=1} (1- x_i) ( 1 - x_{t,i}) = T (N - 2k) + \sum^N_{i=1} x_i \sum^T_{t=1} x_{t,i}.
\end{align}
The best static cache state  $\x_*$ is given by 
$ \x_* = \argmin_{\x \in \mathcal{Z}} \sum^T_{t=1} f_{\vec r_t} (\x) = \argmin_{\x \in \mathcal{Z}} \sum^N_{i=1}  x_i \sum^T_{i=1}x_{t,i}$. The maximum value of $\sum^N_{i=1}  x_{*, i} \sum^T_{t=1}x_{t,i}$ is achieved when $\sum^T_{t=1}x_{t,i} =\sum^T_{t=1}x_{t,j}= T k/N$ for every $i,j \in \mathcal{N}$, and in this case $\vec x_{*}$ can be arbitrary in $\mathcal{Z}$.  Thus, the cost incurred by the static optimum is upper bounded by $    \sum^T_{t=1} f_{\vec r_t} (\x_*) = T (N - 2k) + \sum^N_{i=1} x_{*,i} \sum^T_{t=1} x_{t,i} \leq T (N-2k) +  \frac{Tk^2}{N}$. The regret of $\mathcal{A}$ over time horizon $T$ is lower bounded by 
\begin{align}
  \textstyle  \Regret_T(\mathcal{A}) &=  \textstyle  \sum^T_{t=1} f_{\vec r_t} (\x_t) - \sum^T_{t=1} f_{\vec r_t} (\x_*) \geq T(N-k) - T(N-2k) -T \frac{k^2}{N}  = k \left(1 - \frac{k}{N}\right)T.
\end{align}

We conclude that the regret of any deterministic policy $\mathcal{A}$ is $\Omega(T)$ compared to a static optimum selecting the best state in $\mathcal{Z}$; therefore, it also has $\Omega(T)$ regret compared to a static optimum selecting the best fractional state in $\mathcal{X}$, which includes $\mathcal Z$. \qed
%(stronger static optimum). 
%\end{proof}

\vspace{-.75em}\subsection{Proof of Proposition~\ref{proposition:regret_transfer}}
\label{proof:regret_transfer}
\begin{proof}
The expected service cost incurred when sampling the integral caching states $\z_t$ from $\x_t$ at each time $t$, by the linearity of $f_{\vec{r}_t}$ is  $  \mathbb{E} \left[\sum^T_{t=1} f_{\vec{r}_t} (\vec z_t)\right] =\sum^T_{t=1} \mathbb{E} \left[f_{\vec{r}_t} (\vec z_t)\right] =\sum^T_{t=1} f_{\vec{r}_t} \left(\mathbb{E} \left[\vec z_t\right]\right) = \sum_{t=1}^T f_{\vec{r}_t} (\vec{x}_t)$. The best static configuration $\vec{x}_*$ in the fractional setting can always be selected to be integral; this is because the objective and constraints are linear, so integrality follows from the fundamental theorem of linear programming. Hence, the expected regret for the service cost coincides with the regret of the fractional caching policy. 
\end{proof}
    \vspace{-.75em}\subsection{Proof of Theorem~\ref{theorem:rounding_independent}}
\label{proof:rounding_independent}
\begin{proof}
We consider the catalog $\mathcal{N} = \{1,2\}$, cache capacity $k = 1$, and equal service and update costs $w_i=w'_i= 1$ for $i \in \mathcal{N}$. A policy $\mathcal{A}$ selects the states $\{\x_t\}^T_{t=1} \in \mathcal{X}^T$. 
% We denote by by $I \subset [T]$ the set of time slots where the policy $\mathcal{A}$ selects integral states in $\mathcal{Z}$, where  $\x_t \in \mathcal{Z}$ for $t \in I$ and $\x_t \in \mathcal{X} \setminus \mathcal{Z}$ for $t\in [T] \setminus I$.
The randomized states obtained by $\Xi$ are $\{\z_t\}^T_{t=1}$; thus, we have $\z_t = [1,0]$ w.p. $x_{t,1}$, and $\z_t = [0,1]$ w.p. $ x_{t,2}$. %
An adversary selects the request batch as $\vec r_t = [\mathds{1}_{\{i = i_t\}}]_{i \in \catalog}$ aiming to greedily maximize the cost of the cache, where $\textstyle i_t \triangleq \argmin_{i \in \mathcal{N}} \{x_{t,i}\}$.
The expected service cost  at time $t$ is $
    \mathbb E \left[{f_{\vec r_t} (\z_t)}\right] = %f_{\vec r_t} (\x_t) = 
    1 - x_{t, i_t}$, and the expected update cost is  $ \mathbb{E}\left[\updatecost_{\vec r_t} (\z_t, \z_{t+1})\right] =  x_{t,i_t} (1- x_{t+1, i_t})$. An update cost is incurred when $i_t$ is requested and the state changes from $z_{t,i_t} = 1$ to $z_{t+1,i_t} = 0$; we pay a unitary cost due to fetching a single file that is not requested with probability $ \mathbb{P} \left(z_{t,i_t} = 1, z_{t+1, i_{t}} = 0 \right)$, and this gives the first equality. We use independence of the random variables $\z_t$ and $\z_{t+1}$ to obtain the second equality.
A fixed state $\x = \left[\frac{1}{2}, \frac{1}{2}\right]$ incurs a cost of $\frac{1}{2}$ for every timeslot $t \in [T]$. We define the instantaneous extended regret w.r.t. the fixed state $\x$ for every timeslot $t \in [T]$  as $
    a_t = \mathbb{E}\left[f_{\vec r_t} (\x_t) + \updatecost_{\vec r_t} (\z_t, \z_{t+1})\right] - f_{\vec r_t} (\x)  = \mathbb{E}\left[f_{\vec r_t} (\x_t) + \updatecost_{\vec r_t} (\z_t, \z_{t+1})\right] - \frac{1}{2}$. Observe here that looking for a minimizer over $\mathcal{X}$ or over $\mathcal{Z}$ is equivalent. Because $\x$ is not necessarily the minimizer of the aggregate service cost $\sum^T_{t=1} f_{\vec r_t} (\x)$, we can lower bound the extended regret~\eqref{eq:integral_extendedregret} as $ \textstyle \mathrm{E\mbox{-}Regret}_T(\policy, \Xi) \geq \sum^T_{t=1} a_t$. Without loss of generality, assume that $T$ is even so $\sum^T_{t=1} a_{t} = \textstyle\sum^{T/2}_{k=1} \left( a_{2k}  + a_{2k-1}\right) $, and we have 
{\footnotesize\begin{align}
   \textstyle \sum^T_{t=1} a_{t} &= \sum^{T/2}_{k=1} \left( 1 - x_{2k-1,i_{2k-1}} - x_{2k, i_{2k}} + x_{2k-1,i_{2k-1}} (1 - x_{2k,i_{2k-1}}) +  x_{2k,i_{2k}} (1 - x_{2k+1,i_{2k}}) \right)\nonumber\\
    &\geq \textstyle\sum^{T/2}_{k=1} \left( 1 - x_{2k-1,i_{2k-1}} - x_{2k, i_{2k}} + x_{2k-1,i_{2k-1}} (1 - x_{2k,i_{2k-1}})\nonumber \right).
\end{align}}
From the definition of $i_{2k}$  we have $x_{2k,i_{2k}} \leq (1- x_{2k, i_{2k-1}})$ for any $k \in [T/2]$. Thus, 
{\footnotesize
\begin{align}
    &\textstyle\sum^T_{t=1} a_{t} \geq\textstyle \sum^{T/2}_{k=1} \left(  1 - x_{2k-1,i_{2k-1}} - x_{2k, i_{2k}} + x_{2k-1,i_{2k-1}}  x_{2k,i_{2k}}  \right)=\sum^{T/2}_{k=1}  1 - x_{2k-1,i_{2k-1}} - x_{2k,i_{2k}}  \left( 1- x_{2k-1,i_{2k-1}}\right) \nonumber\\
   &\geq \textstyle \sum^{T/2}_{k=1} \left( 1 -  x_{2k-1,i_{2k-1}} - \frac{1}{2}\left( 1- x_{2k-1,i_{2k-1}}\right)\right) = \sum^{T/2}_{k=1} \left(\frac{1}{2} - \frac{1}{2}  x_{2k-1,i_{2k-1}} \right) \geq \frac{T}{8}. \nonumber
\end{align}
}
The second and third inequalities are obtained using $x_{t,i_t} \leq \frac{1}{2}$ for every timeslot $t \in [T]$; a direct result from the definition of $i_t$. We combine the above lower bound with $\textstyle \mathrm{E\mbox{-}Regret}_T(\policy, \Xi) \geq \sum^T_{t=1} a_t$ to obtain $\mathrm{E\mbox{-}Regret}_T(\policy, \Xi) \geq \frac{T}{8}$.
\end{proof}

\vspace{-.75em}\subsection{Family of Coupling Schemes with Sublinear Update Cost}
The following theorem provides a sufficient condition for the sublinearity of the expected total update cost of the random cache states $\{\z_t\}^T_{t=1}$ obtained through a rounding scheme~$\Xi$ from the input fractional states $\{\x_t\}^T_{t=1}$. 
\begin{theorem}
\label{theorem:rounding_coupled}
Consider an OMD Algorithm and a joint distribution of $(\vec z_t,\vec z_{t+1})$ 
that satisfy (a)~$\mathbb{E}[\vec z_t] = \vec{x}_t$ and $\mathbb{E}[\vec z_{t+1}] = \vec{x}_{t+1}$, and (b)~$\mathbb{E} \left[ \updatecost_{\vec r_t}({ \vec z_{t}, \vec z_{t+1}})\right] = \BigO{\norm{\vec{x}_{t+1} - \vec{x}_{t}}_1}$. This algorithm incurs an expected service cost equal to the service cost of the fractional sequence. 
Moreover, if $\eta = \Theta\left({\frac{1}{\sqrt{T}}}\right)$, the algorithm has also  \BigO{\sqrt{T}} expected update cost and then  \BigO{\sqrt{T}} extended regret.

\end{theorem}

\begin{proof}
Consider that the sequence $\{\vec{x}_t\}^T_{t=1}$ is generated by an OMD algorithm, configured with a $\rho$-strongly convex mirror map $\Phi$ w.r.t a norm $\norm{\,\cdot\,}$. Assume that we can find a joint distribution of $(\vec z_t,\vec z_{t+1})$ satisfying $\mathbb{E} \left[ \updatecost_{\vec r_t}({ \vec z_{t}, \vec z_{t+1}})\right] = \BigO{\norm{\vec{x}_{t+1} - \vec{x}_{t}}_1}$, where $\mathbb{E}[\vec z_t] = \vec{x}_t$ and $\mathbb{E}[\vec z_{t+1}] = \vec{x}_{t+1}$. Then there exists a constant $\gamma_1 > 0$, such that $\mathbb{E} \left[ \updatecost_{\vec{r}_t} (\vec z_{t} , \vec z_{t+1})\right] \leq \gamma_1 \norm{\vec{x}_{t+1} - \vec{x}_t}_{1}$.
Moreover, there exists $\gamma_2 > 0$, such that
$ \gamma \norm{\vec{x}_{t+1} - \vec{x}_t}_{1} \leq \gamma_1 \gamma_2  \norm{\vec{x}_{t+1} - \vec{x}_t}$, and this gives
$
\mathbb{E} \left[ \updatecost_{\vec{r}_t} (\vec z_{t} , \vec z_{t+1})\right] \leq \gamma_1 \gamma_2 \norm{\vec{x}_{t+1} - \vec{x}_t}.
$ As $\Phi$ is $\rho$-strongly convex w.r.t. the norm $\norm{\,\cdot\,}$, it holds
\begin{align}
&D_\Phi(\vec{x}_t, \vec{y}_{t+1}) =   \Phi(\vec{x}_t) - \Phi(\vec{y}_{t+1}) + {\nabla \Phi(\vec{x}_{t})}^T (\vec{y}_{t+1} - \vec{x}_t) + {\left(\nabla \Phi(\vec{x}_{t}) - \nabla \Phi(\vec{y}_{t+1})\right)}^T (\vec{x}_t - \vec{y}_{t+1} ) \nonumber \\
&\leq - \frac{\rho}{2} \norm{\vec{x}_t - \vec{y}_{t+1}}^2 + \eta g^T_t (\vec{x}_t - \vec{y}_{t+1}) \leq - \frac{\rho}{2} \norm{\vec{x}_t - \vec{y}_{t+1}}^2 + \eta \norm{\vec{x}_t - \vec{y}_{t+1}} L\leq \frac{\eta^2 L^2}{2 \rho}
\label{eq:bounding_bregman_step}
\end{align}
The above inequalities  are obtained using the strong convexity of $\Phi$ and the update rule, Cauchy-Schwarz inequality, and the inequality $a x - b x^2 \leq \max_{x} a x - b x^2 = a^2/4b $ as in the last step in the proof of \cite[Theorem 4.2]{bubeck2015convexbook}, respectively. We have 
% \mathbb{E} \left[ \updatecost_{\vec{r}_t} (\vec z_{t} , \vec z_{t+1})\right]  &\leq  M \norm{\vec{x}_{t+1} - \vec{x}_t}
$ \textstyle\norm{\vec{x}_{t+1} - \vec{x}_t}\textstyle\leq  \sqrt{\frac{2}{\rho} D_\Phi(\vec{x}_t, \vec{x}_{t+1})}  \leq   \sqrt{\frac{2}{\rho} D_\Phi(\vec{x}_t, \vec{y}_{t+1})} \stackrel{\eqref{eq:bounding_bregman_step}}{\leq }\sqrt{2 \eta^2 \frac{L^2}{2\rho^2}} \leq  \frac{L\eta}{\rho} $.
The first inequality is obtained using the strong convexity of $\Phi$, and the second using the generalized Pythagorean inequality\cite[Lemma 11.3]{Bianchi_Prediction}.
We combine $\textstyle\norm{\vec{x}_{t+1} - \vec{x}_t} \leq L\eta/\rho$ and  $\mathbb{E} \left[ \updatecost_{\vec{r}_t} (\vec z_{t} , \vec z_{t+1})\right] \leq \gamma_1 \gamma_2 \norm{\vec{x}_{t+1} - \vec{x}_t}$ to obtain $    \mathbb{E} \left[ \updatecost_{\vec{r}_t} (\vec z_{t} , \vec z_{t+1})\right] \leq \gamma_1 \gamma_2 \norm{\vec{x}_{t+1} - \vec{x}_t} \leq  \gamma_1 \gamma_2 \frac{L\eta}{\rho}$. The total update cost  is $\sum^{T-1}_{t=1} \mathbb{E} \left[ \updatecost_{\vec{r}_t} (\vec z_{t} , \vec z_{t+1})\right]  \leq \gamma_1 \gamma_2 \frac{L\eta  }{\rho} T$. When OMD  has a fixed learning rate $\eta = \Theta\left(\frac{1}{\sqrt{T}}\right)$, we obtain $\sum^{T-1}_{t=1} \mathbb{E} \left[ \updatecost_{\vec{r}_t} (\vec z_{t} , \vec z_{t+1})\right]  = \BigO{\sqrt{T}}$. The expected service cost is $\mathbb{E} \left[\sum^T_{t=1} f_t(\z_t)\right] = \sum^T_{t=1} f_t(\x_t) = \BigO{T}$; the first equality is obtained from the linearity of the expectation operator and the function $f_{\vec r_t}$, and the second equality is obtained using the bound in Eq.~\eqref{eq:omd_general_regret} with $\eta= \Theta\left(\frac{1}{\sqrt{T}}\right)$.
\end{proof}

\vspace{-.75em}\subsection{Proof of Theorem~\ref{theorem:algorithm_rounding_coupled}}
% \label{proof:elementary_movement}
\label{proof:algorithm_rounding_coupled}
Lemma~\ref{lemma:elementary_movement_preserves_marginals} and Lemma~\ref{lemma:expected_movement_lemma} guarantee that Algorithm~\ref{algo:online_rounding} used  with an OMD algorithm satisfies the hypothesis of Theorem~\ref{theorem:rounding_coupled} and, hence, provides sublinear extended regret. 

\begin{SCfigure}[40][t]
\centering
    \includegraphics[trim=80 0 75 0,clip,width=0.55\linewidth]{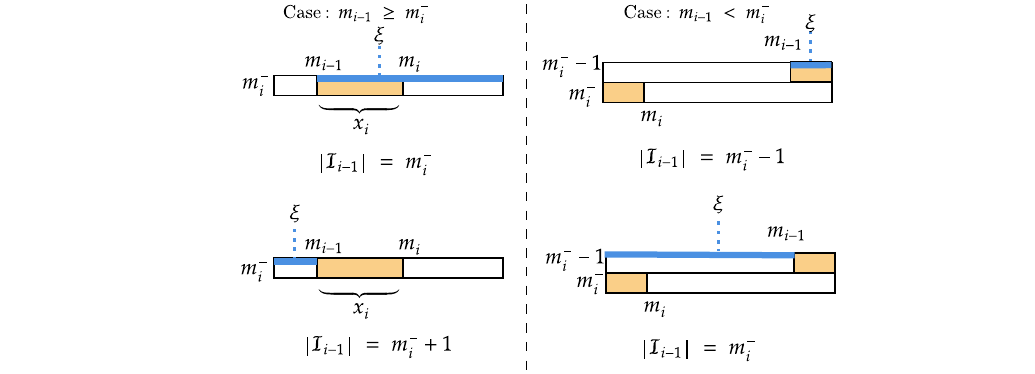}
    %  \subcaptionbox{}[0.4\linewidth]{\includegraphics[trim=90 0 90 0,clip,width=\linewidth]{graphical_res/Aug/movement_costs.pdf}}
    %   \subcaptionbox{}{}
\caption{The support and distribution of the random variable $|\mathcal I_{i-1}|$ for $m_{i-1} \geq m_i^-$ (left) and $m_{i-1} < m_i^-$ (right). The blue shaded area represents the probability that $|\mathcal I_{i-1}|$ takes the value indicated below the corresponding image. This figure can be viewed as a subset of the rows in Fig.~\ref{fig:elementary_movement}.  }
    \label{fig:visual_proof_expected_value}
\end{SCfigure}

\begin{SCfigure}[40][t]
\centering
    \includegraphics[trim=10 0 90 0,clip,width=0.55\linewidth]{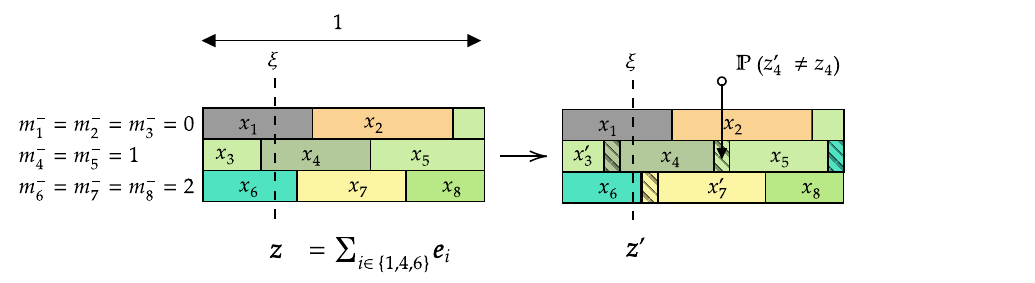}
\caption{The random integral cache configuration obtained by calling \Rounding{} given the  fractional cache state  $\x$ and $\xi$ (left). When $\xi$ is kept fixed, the  probability over the initial choice of $\xi$ the random integral cache configuration rounded from a new fractional state $\x' =\x- \delta \vec e_3 + \delta \vec e_7 $ is different is illustrated by the dashed areas (right).}
    \label{fig:elementary_movement}
\end{SCfigure}

\begin{lemma}
\label{lemma:elementary_movement_preserves_marginals}
The  random integral cache state $\vec z$ obtained by calling  Algorithm~\ref{algo:online_rounding} with fractional cache configuration input $\x \in \mathcal{X}$ satisfies $
    \vec z  \in \mathcal{Z}$ and $\mathbb{E}_\xi [\vec z] = \x$.
\end{lemma}
\begin{proof}
We employ the shorthand notation $m_i = \sum^i_{j=1} x_j$ and $m^-_i = \floor{m_i}$. The choice of $\xi$ (see~Fig.~\ref{fig:elementary_movement}) defines $k$ different thresholds $\xi, \xi+1,\dots, \xi+k-1$. For each threshold, we select the first item, whose accumulated mass exceeds the threshold. As $m_N=k$, we are guaranteed to exceeds all $k$ thresholds, and as $x_i \leq 1$, we are guaranteed to select one item for each threshold. Therefore $\vec z$ belongs to $\mathcal{Z}$. From Algorithm~\ref{algo:online_rounding} for any  $i \in \mathcal{N}$ we have $
    \mathbb{P} ( z_i = 1) = \mathbb{P} (\mathcal I_i \setminus \mathcal I_{i-1} = \{i\}) = \mathbb{P} \left( m_i \geq \xi + |\mathcal I_{i-1}|  \right)$.

If $m_{i-1}\geq m^-_{i}$, see Fig.~\ref{fig:visual_proof_expected_value} (left), then $|\mathcal I_{i-1}| \in \{m^-_i, m^-_i + 1\}$ and 
$ \mathbb{P} \left( m_i \geq \xi + |\mathcal I_{i-1}|  \right) =\frac{m_i - m_{i-1}}{(m^-_i + 1- m_{i-1})} \cdot (m^-_i + 1 - m_{i-1}) + 0 \cdot (m_{i-1} - m^-_{i})= x_i.
  $
If $m_{i-1} < m^-_i$, see Fig.~\ref{fig:visual_proof_expected_value} (right), then $|\mathcal I_{i-1}| \in \{m^-_i-1, m^-_i\}$ and 
$\mathbb{P} \left( m_i \geq \xi + |\mathcal I_{i-1}|  \right) =  1 \cdot (m^-_i - m_{i-1}) + \frac{m_i - m^-_i}{m_{i-1} - m^-_{i} + 1} \cdot  (m_{i-1} - m^-_{i} + 1)= m_i - m_{i-1} = x_i.$
\end{proof}

\sidecaptionvpos{figure}{c}

\sidecaptionvpos{figure}{c}

% The boundedness of the expected movement of two random integral cache obtained by an elementary movement states obtained by calling Algorithm~\ref{algo:online_rounding}.
% \begin{figure}[t]
%     \centering
%     \includegraphics[width=\linewidth]{graphical_res/Aug/movement_costs.pdf}
%     \caption{The random integral cache configuration obtained by calling \Rounding{} given the  fractional cache state  $\x$ and $\xi$ (top figure). When $\xi$ is kept fixed (online coupled rounding), the  probability the random integral cache configuration rounded from a new fractional state $\x' =\x- \delta \vec e_3 + \delta \vec e_7 $ is different is illustrated by the dashed areas (bottom figure). }

    % (left figure) and $\x' = \x - \delta \vec e_3 + \delta \vec e_7$ (right figure). The probability of movement of the integral cache configurations is illustrated by the dashed areas. The catalog size is taken to be $N = 8$ and the cache size $k=3$.}
    % \label{fig:movement_probability}
% \end{figure}

\begin{lemma}
\label{lemma:elementary_movement}
Consider a fractional cache configuration $\x'$ obtained by an elementary mass movement of $\delta$ from $u \in \mathcal{N}$ to $v \in \mathcal{N}$ for configuration $\x \in \mathcal{X}$, i.e., $\vec x' = \x - \delta \boldsymbol e_u + \delta \boldsymbol e_v$. Algorithm~\ref{algo:online_rounding} outputs the random integral cache configurations $\vec z$, and $\vec z'$, given the input fractional cache states $\x$ and $\x'$, respectively. The random integral configurations satisfy $\mathbb{E}_\xi\left[ \norm{\vec z' - \vec z}_{1, \vec w'}\right ] \leq 2kN\norm{\vec w'}_\infty \delta$, where $
    \norm{\x}_{1, \vec w'} \triangleq \sum_{i \in \catalog} \card{x_i} w'_i$,  and note that $\updatecost_{\vec r_t} \left(\z_{t} ,\vec z_{t+1}\right) \leq  \norm{\z_t - \z_{t+1}}_{1, \vec w'}$.
\end{lemma}
\begin{proof}
The probability that a random integral cache state $\vec z'$ is changed w.r.t $\vec z$ can be upper bounded as  $     \mathbb{P} (\vec z \neq \vec z' )  \leq \sum_{i\in \mathcal{N}} \mathbb{P} (z_i \neq z'_i)$. 
Consider w.l.g that $u < v$, then  $     \mathbb{P} (z_i \neq z'_i) = 0$ for $ i \in \mathcal{N} \setminus \{u, u+1, \dots, v\}$, 
and $\mathbb{P} (z_i \neq z'_i) = \delta$ for $i \in  \{u, u+1, \dots, v\}$. We obtain  $\mathbb{P} (\vec z \neq \vec z' )  \leq \delta (v - u + 1)$ (e.g., see Fig.~\ref{fig:elementary_movement}). More generally, for $u \neq v$ we have $
         \mathbb{P} (\vec z \neq \vec z' )  \leq \delta (|v - u|+ 1)$.
We conclude that 
\begin{align*}
    \mathbb E \left[{\norm{\vec z' - \vec z}_{1, \vec w'}}\right] &\leq   \max_{(\vec z, \vec z') \in \mathcal{Z}^2}\norm{\vec z' - \vec z}_{1, \vec w'}  \cdot \mathbb{P} (\vec z \neq \vec z' ) %\\
    %&
    \leq 2k\norm{\vec w'}_\infty   (|v - u|+ 1) \delta \leq 2kN\norm{\vec w'}_\infty \delta. \qedhere
\end{align*}
\end{proof}
% \vspace{-.75em}\subsection {Proof of Theorem~\ref{theorem:algorithm_rounding_coupled}}

% \begin{figure}
%     \centering
%     \includegraphics[width=\linewidth]{graphical_res/Aug/algo.pdf}
%     \caption{The random integral cache configuration $\vec z_t$ obtained by calling \Rounding{} on the fractional cache states $\x_t$. The catalog size is $N = 8$ and the cache size $k=3$.}
%     \label{fig:online_coupled_rounding}
% \end{figure}
\sidecaptionvpos{figure}{c}
\begin{SCfigure}[50][t]
\centering
    \includegraphics[trim=70 0 70 0,clip,width=.55\linewidth]{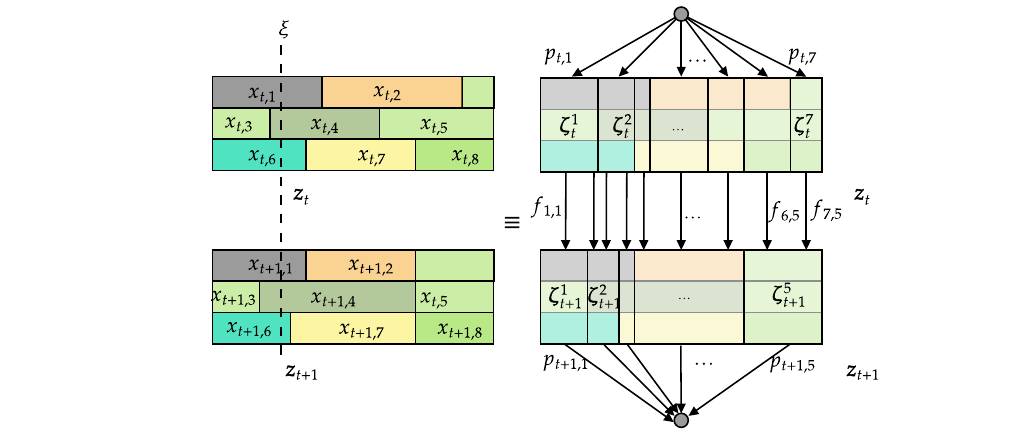}
    \caption{Coupling induced by \Rounding{}~Algorithm~\ref{algo:online_rounding} when $\xi$ is fixed. The flow $f_{i,j}$ is the joint probability $\mathbb{P}(\vec z_{t+1} = \zeta_{t+1,j}, \vec z_{t} = \zeta_{t,i})$, so that the next state is $\zeta_{t+1,j}$ and the previously selected state is $\zeta_{t,i}$.}
    \label{fig:conditioned_implicit}\vspace{-1.5em}
\end{SCfigure}

\begin{lemma}
\label{lemma:expected_movement_lemma}
% Algorithm~\ref{algo:online_rounding} outputs the random integral cache configurations $\vec z_{t}$ and $\vec z_{t+1}$  with $ \mathbb E_\xi [\vec z_t] = \x_t$ and $\mathbb E_\xi [\vec z_{t+1}] = \vec x_{t+1}$. 
The expected movement cost of the random integral cache states generate by Algorithm~\ref{algo:online_rounding} is $
  \mathbb E_\xi \left[ \updatecost_{\vec r_t} \left(\z_{t} ,\vec z_{t+1}\right)\right] =  \BigO{\norm{\vec{x}_{t} - \vec{x}_{t+1}}_{1}}$,
when $\xi$ is sampled once u.a.r. from the interval $[0,1]$ and then fixed for $t \in [T]$.
\end{lemma}
\begin{proof}

The general fractional movement caused by a policy $\mathcal{A}$ changes the cache state from fractional state $\vec{x}_t \in \mathcal{X}$ to $\vec{x}_{t+1} \in \mathcal{X}$, and we denote by $\mathcal{J} = \left\{ i \in \mathcal{N}: x_{t+1,i} - x_{t,i} > 0 \right\}$ the set of components that have a fractional increase. We have $    \vec{x}_{t+1} = \vec{x}_{t} + \sum_{j \in \mathcal{J}} \phi_j e_j - \sum_{i \in \mathcal{N} \setminus \mathcal{J} } \phi_i e_i$.
where $\phi_i, i \in \mathcal{N}$ is the absolute fractional change in component $i$ of the cache. Remark that we have $    \norm{\vec{x}_{t+1} - \vec{x}_{t}}_{1, \boldsymbol w'} = \sum_{i\in\mathcal{N}} w'_i \phi_i$.
From the capacity constraint we know that $\sum_{i \in \mathcal{N} \setminus \mathcal{J} } \phi_i = \sum_{j \in \mathcal{J}} \phi_j$. If we want to decompose this general fractional change to elementary operations, then we need to find a flow $\left[\delta_{i,j} \right]_{(i,j) \in (\mathcal{N}\setminus \mathcal{J}) \times \mathcal{J}}$ that moves $\sum_{j \in  \mathcal{J}} \phi_j$ mass from the components in $\mathcal{N} \setminus \mathcal{J}$ to to those in $\mathcal{J}$. This requires at most $N-1$ elementary operations. 
We define the map $\nu: \mathcal{N}^2 \to \mathbb{N}$ that provides an order on the sequence of elementary operations. Let $\vec z^{\nu (i,j)}$ be the random cache state that could have been sampled after the $\nu(i,j)$-th elementary operation where $\mathbb{E} \left[\z^{\nu (i,j)}\right] = \x^{\nu (i,j)}$, and the total number of operations is denoted by $|\nu| \leq N-1$. Note that by definition $\vec z^{|\nu|} = \vec z_{t+1}$, and we take $\vec z^0 = \vec z_t$.
For each of these  operations we pay in expectation at most $ 2kN\norm{\vec w'}_\infty{\delta_{i,j}}$  update cost from Lemma~\ref{lemma:elementary_movement}. Then the total expected movement cost is: %given by 
\begin{align}
\textstyle \mathbb{E}_\xi \big[\updatecost_{\vec r_t} (\z_{t} &,\vec z_{t+1})\big] \leq\textstyle \mathbb{E}_\xi [\norm{\vec z_{t+1} - \vec z_t}_{1, \vec w'}] = \mathbb{E}_\xi \left[\norm{\sum^{|\nu|-1}_{l=0} \left( \vec z^{l+1} - \vec z^l\right)}_{1, \vec w'}\right] \leq \sum^{|\nu|-1}_{l=0} \mathbb{E}_\xi \left[\norm{\vec z^{l+1} - \vec z^l}_{1, \vec w'}\right]\nonumber \\
 & \leq    \textstyle\sum_{i \in \mathcal{N} \setminus \mathcal{J} } \sum_{j \in  \mathcal{J} } 2k\norm{\vec w'}_\infty   N \delta_{i,j}  =2kN\norm{\vec w'}_\infty    \sum_{i \in \mathcal{N} \setminus \mathcal{J}} \phi_i \leq k N \norm{\vec w'}_\infty \norm{\vec x_{t+1} - \vec x_{t}}_{1}. \nonumber
\end{align}
%The second inequality follows from the triangle inequality and linearity of expectation. 
 The update cost is thus \BigO{\norm{\vec{x}_{t+1} - \vec{x}_{t}}_{1}} in expectation.
%  , and Lemma~\ref{lemma:elementary_movement_preserves_marginals} specifies that $\mathbb{E}_\xi\left[\vec z_{t+1}\right] = \x_{t}$ for every timeslot $t \in [T]$;  therefore we preserve the sublinearity of regret from Theorem \ref{theorem:rounding_coupled}. 
\end{proof}

% The deterministic integral caching policies are a particular class of fractional caching policies, where we restrict the policies to only pick integral cache states in $\mathcal{Z} \triangleq \mathcal{X} \cap \{0,1\}^N$  deterministically.

\end{appendices}
\end{document}